\documentclass[a4paper,fleqn]{cas-sc}

\usepackage[numbers]{natbib}

\def\tsc#1{\csdef{#1}{\textsc{\lowercase{#1}}\xspace}}
\tsc{WGM}
\tsc{QE}
\tsc{EP}
\tsc{PMS}
\tsc{BEC}
\tsc{DE}

\usepackage[utf8]{inputenc}             
\usepackage[T1]{fontenc}            
\usepackage{algorithm2e}                
\makeatletter
\renewcommand{\@algocf@capt@plain}{above}
\makeatother
\usepackage{graphicx}                   
\usepackage{import}                     
\usepackage{tabularx}                   
\usepackage{textcomp}                   
\usepackage{varioref}                   
\usepackage{float}
\usepackage{wrapfig}
\usepackage{xcolor}
\definecolor{LightCyan}{rgb}{0.88,1,1}
\definecolor{Gray}{gray}{0.9}
\usepackage[colorlinks]{hyperref}
\AtBeginDocument{%
  \hypersetup{
    citecolor=cyan,
    linkcolor=red,
    urlcolor=blue}}
\usepackage{cleveref}

\usepackage{verbatim}
\usepackage{caption,subcaption}         
\usepackage{babel}
\usepackage{colortbl}                   
\usepackage{csquotes}                   
\usepackage[section, below]{placeins}   
\usepackage{booktabs}
\usepackage{tikz}                       
\usepackage{pgfplots}
\pgfplotsset{compat=newest}
\usepackage{array}                      

\usepackage{listings, lstautogobble}
\usepackage{color}

\definecolor{dkgreen}{rgb}{0,0.6,0}
\definecolor{gray}{rgb}{0.5,0.5,0.5}
\definecolor{mauve}{rgb}{0.58,0,0.82}

\definecolor{tb_color_1}{RGB}{245,124,0}
\definecolor{tb_color_2}{RGB}{0,167,247}

\pgfplotscreateplotcyclelist{tb}{
semithick,tb_color_1\\%
semithick,tb_color_2\\%
}
\newcommand{\tfFromcsv}[3] {    
    \begin{tikzpicture}[#3]
        \begin{axis}[cycle list name=tb,
                     grid=both,
                     grid style={solid,gray!30!white},
                     axis lines=middle,
                     xlabel={step},
                     ylabel={#2},
                     xmin=0,
                     x label style={at={(axis description cs:0.5,-0.1)},anchor=north},
                     y label style={at={(axis description cs:-0.1,.5)},rotate=90,anchor=south},]
          \addplot table [x=Step, y=Value, col sep=comma] {#1};
        \end{axis}
    \end{tikzpicture}%
}

\usepackage{subcaption}

\usepackage{subcaption}

\begin{document}
\shorttitle{DoS and DDoS Mitigation Using Variational Autoencoders}
\shortauthors{Bårli et~al.}

\title [mode = title]{DoS and DDoS Mitigation Using Variational Autoencoders}

\author[1,2]{Eirik Molde Bårli}

\address[1]{Department of Computer Science,
    OsloMet -- Oslo Metropolitan University,
    P.O.~Box 4 St.~Olavs plass, N-0130 Oslo, Norway}

\address[2]{Department of Informatics, University of Oslo,
P.O.~Box 1080 Blindern, N-0316 Oslo, Norway}


\author[1]{Anis Yazidi}
  \author[granada]{Enrique Herrera Viedma}
\author[1]{Hårek Haugerud}

\address[granada]{Andalusian Research Institute in Data Science and Computational Intelligence,
University of Granada, Granada 18071, Spain.}

\begin{abstract}
DoS and DDoS attacks have been growing in size and number over the last decade and existing solutions to mitigate these attacks are in general inefficient. Compared to other types of malicious cyber attacks, DoS and DDoS attacks are particularly more challenging  to combat. With their ability to mask themselves as legitimate traffic, developing methods to detect these types of attacks on a packet or flow level, has proven to be a difficult task.
In this paper, we explore the potential of Variational Autoencoders to serve as a component within an intelligent security solution that differentiates between normal and malicious traffic.
Two methods based on the ability of Variational Autoencoders to learn  latent representations from network traffic flows are proposed.  The first method resorts to a  classifier based on the latent encodings obtained from Variational Autoencoders learned from traffic traces.  The second method is rather an anomaly detection method where the Variational Autoencoder is used to learn the abstract feature representations of exclusively legitimate traffic. Then anomalies are filtered out by relying on the reconstruction loss of the Variational Autoencoder.

Both of the proposed methods have been thoroughly tested on two separate datasets with a similar feature space. The results show that both methods are promising, with a slight superiority of the classifier based method over the anomaly based one.

\end{abstract}

\begin{graphicalabstract}
  \includegraphics[scale=0.7]{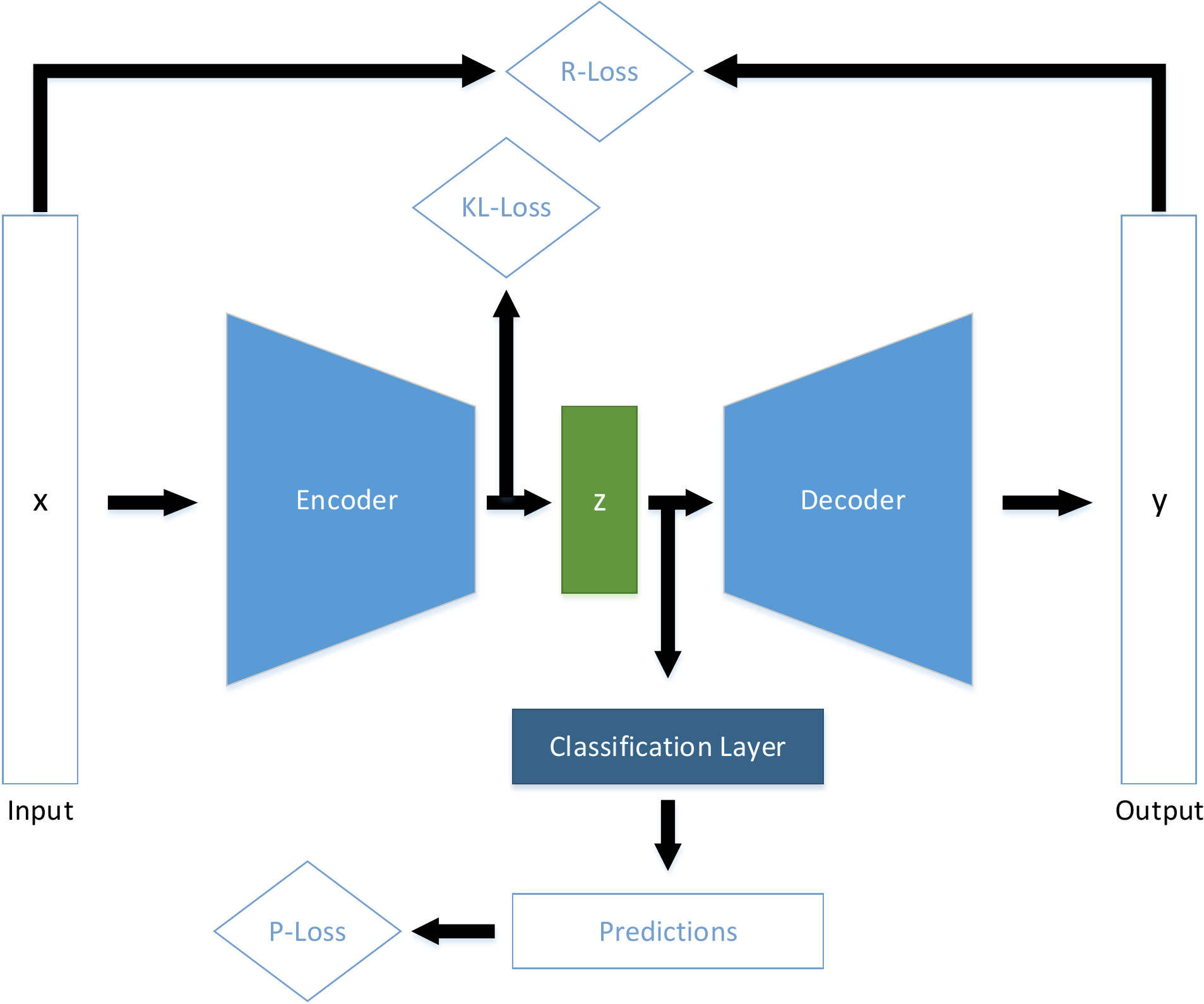}

\end{graphicalabstract}

\begin{highlights}
\item We show the potential of of Variational Autoencoders in separating between normal and malicious traffic.
\item We proposed two approaches based on Variational Autoencoders.
\item The first method resorts to a  classifier based on the latent encodings obtained from Variational Autoencoders learned from traffic traces.
\item The first method uses the reconstruction loss to detect malicious traffic as anomaly.
\item Thorough experiments demonstrate the effectiveness of the proposed methods.

\end{highlights}


\maketitle

\section{Introduction}\label{sec:1}

The Internet has been growing in size ever since its conception, allowing access to any device with networking capabilities. As new and more advanced technology is researched, created and sold, at ever decreasing prices, more and more people gain access to the Internet through one or more networking capable devices. Malicious network traffic does not originate exclusively from personal computers, but can come from multiple sources, such as devices in the Internet of things (IoT). In 2017, the amount of network devices was estimated to consist of 18 billion units, according to an ongoing initiative by Cisco to track and forecast networking trends \cite{Cisco-forecast-2019}. This number is expected to increase to about 28.5 billion devices by 2022. Embedded devices are included in these statistics as well, which do not necessarily have Internet connection, but the amount of devices with potential for launching malicious network traffic is still in the billions. Given the sheer number of units available for network attacks, it would be nearly impossible to manually create solutions for combating the problem of filtering out the malicious traffic from the harmless.

A typical IoT divice has some kind of sensor and an Internet connection and this includes devices such as printers or surveillance cameras with networking capabilities, remote health monitoring devices, domestic robots, smart light bulbs and many other similar devices. Some of these devices are less secure than others, and more vulnerable to theft in the sense that they can be used as a part of a botnet, or as a source of attack by an external party. Any device with Internet capabilities has an Internet Protocol (IP) address and an internal computing device, and is thus able to send malicious traffic. Amongst the most notable, and perhaps hardest to prevent are denial of service (DoS) and distributed DoS (DDoS) attacks. DoS and DDoS attacks have become immense threats over the last decade to any Internet connected machine. In 2015, a global survey of a number of companies conducted by Kaspersky found that 50\% of DDoS attacks caused a noticeable disruption in services, and 24\% led to a complete denial of service \cite{Kaspersky-risk-report-2015,Kaspersky-risk-report-2015-original}. As attacks continue to evolve and as the amount of available IoT devices to launch attacks from grows, these percentages could very well increase, as would the demand for working mitigation systems.

DoS and DDoS attacks are similar in intention and often similar in execution. The goal for both types of attacks is to cause a denial of service for their target, either by exhausting the victim’s bandwidth or system resources, such as CPU or memory. In general terms, causing a denial of service means to overload a victim, for example a web server, with traffic. Once a system receives more traffic than its bandwidth or system resources are able to handle, it will fail to receive or send parts of the intended data traffic. DoS means denial of service attacks that come from a single source while DDoS is denial of service attacks originating from a distributed range of machines and networks. Both types of attacks have proven effective and devastating, but the most tedious to handle is DDoS. The major reason for this is the potential significant size of a DDoS attack.

Cisco reported that the average attack size of a DDoS attack in terms of traffic load was 990 Mbps in 2017 \cite{Cisco-forecast-2019}. Because of its distributed nature, a DDoS attack can generate attacks much larger than a normal DoS attack. In late 2016, the peak size was reported to be 1.1 Tbps \cite{Scott-2016} as a result of a DDoS attack consisting of multiple compromised IoT devices. In 2018 the peak size was reported by Cisco to have reached 1.7 Tbps originating from a vulnerability in the memcached protocol \cite{Cisco-forecast-2019}, resulting in the largest DDoS attack to date.

Creating an effective mitigation system requires consideration of multiple facets of both DoS and DDoS attacks.
An effective mitigation system needs to be able to handle large amounts of traffic, processing malicious and normal traffic simultaneously. The system is required to separate harmful from harmless traffic, at such a rate that the targeted victim is able to handle the traffic load. Considering that network packets from a DDoS or DoS attack are not harmful in and of themselves, a mitigation system can let through a certain amount of malicious traffic, while prioritizing letting through as much normal traffic as possible. 
A mitigation system that fails to meet any of these requirements risks exposing the victim to the attack. In case of inability to handle incoming traffic load to a server, the mitigation system would cause a general slowdown of the server, due to how it needs to inspect each packet before forwarding or allowing the packets through. In case of  inability to separate normal traffic from malicious, the mitigation system risks blocking normal users from accessing the server, or letting too many malicious packets through. Handling large DDoS attacks can be particularly difficult, considering that the defending system has limited resources to spare for defense, while an attack has the potential to capture and use much larger amounts of devices and their resources to overpower the victim. There are commercial systems which reroutes the enormous amounts of traffic of DDoS attacks to a network of dedicated cloud services which filters out the malign traffic, so called cloud scrubbing centers\cite{moura2020into, you2020scheduling}. However, such cloud services also need an efficient method to differentiate between the attack-packets and the normal packets which are supposed to be forwarded to the destination service.    

Current mitigation systems typically focus on some form of pattern recognition or frequency based detection, or a combination of the two. It is common to have a detection system in place but often this only means that it is able to know when an attack is happening, while being unable to stop it. This puts system administrators in the precarious situation of having to combat the attack manually, which is usually only feasible after having the victim server shut down \cite{Zargar-et-al-2013}. When discussing detection systems, it generally encompasses systems that are able to recognize when an attack is happening, or has happened in the past by analyzing the network traffic logs. A mitigation system is different in the way that it is able to differentiate malicious from normal traffic on a packet or flow level, i.e.\ based on IP addresses or other packet information during the attack.

Part of what makes it difficult to differentiate between normal and denial of service traffic is because of their similar behavior. If we compare two network packets in isolation, one is from a legitimate user, and another is from a compromised computer being used as a part of an attack, in a vacuum the differences would be few to none. DoS and DDoS packets are not harmful on their own. It is the amount of traffic they send that in turn overloads a system, drowning out legitimate attempts at access. This makes it difficult to implement solutions that can keep out malicious connections.

A mitigation system needs to be able to detect what packets or IP addresses are malicious, and stop them from entering the network. Using machine learning algorithms could be a potential way for creating such a system. As machine learning can itself potentially find relations between packet information and intent.

In the last few years, we have seen a rise in the usage and success of machine learning algorithms. Machine learning, specifically deep learning, can be applied to a myriad of different problem domains including classification of different types of data and anomaly detection, using a variety of different architectures such as convolutional networks or autoencoders \cite{perera-et-al-2018}. Deep learning has proven effective at analyzing and extracting useful data patterns that manual and automatic approaches are unable to solve in other problem domains other than computer network traffic. Manual and automatic approaches rely on constant updates as well as human interaction to stay effective against DoS and DDoS attacks. A deep learning solution on the other hand, is autonomous, requiring minimal human interaction. These are among the reasons a deep learning solution could prove to be effective at combating DoS and DDoS attacks.

This article proposes two approaches for DoS and DDoS mitigation utilizing the framework of the Variational Autoencoder (VAE) by Kingma and Welling \cite{Kingma-and-Welling-2013}. The first approach is the Latent Layer Classification algorithm (LLC-VAE), which aims to identify different types of computer network traffic from the latent layer representations of the VAE. The second approach is the Loss Based Detection algorithm (LBD-VAE), which tries to identify normal and malicious traffic patterns based on the VAE reconstruction loss function. The two proposed approaches are not meant as complete solutions for mitigation systems, nor are they aimed at creating solutions that work against all types of DoS and DDoS attacks. The contribution of this article is the research on how deep learning, specifically a VAE, can detect specific types of DoS and DDoS traffic from network flows. In addition, whether this can be generalized to detect other types of DoS and DDoS attacks, and to what extent. 

\subsection{Outline}

The remainder of this article is organized as follows. Section \ref{related-work} surveys related works within the field of DoS and DDOS detection.
Section \ref{background-autoencoders} is a short review of autoencoders and  section \ref{solution} discusses the two proposed deep learning approaches. Section \ref{experimental-setup} describes the experimental setup  in which the proposed approaches have been tested as well as the datasets. Section \ref{analysis} reports  the experimental results achieved by the two proposed approaches.  Finally, in Section \ref{conclusion}, we draw final remarks and conclude the article.

\section{Related Work}
\label{related-work}

Techniques to combat DoS and DDoS attacks are many and varied. Some focus on stateless packet information, others rely on meta information from stateful packet flows. In this section, we will provide a few relevant research papers and topics in order to create an overview of different strategies and methods to both discover and mitigate DoS and DDoS attacks. We will also present research that aims to improve or adapt existing techniques relevant for our research or future research.

\subsection{Data Mining and Density-based Geographical Clustering}
IP addresses, although not always reliable to pinpoint exactly where a connection is based, provide the general geographical location such as country and city, unless the address is spoofed or hidden in layers of redirection. The authors of the article discussed here propose two novel methods for DDoS mitigation, specifically for HTTP flooding, based on hidden data structures in historical traffic \cite{kongshavn2020mitigating}. The first method is \textit{A priori}-based frequent networks (AFN), used to discover common known prefixes during training and relating it to unknown data later on. \textit{A priori} in this context is a form of association rule learning, used to investigate frequent item sets and discovering relationships between data variables. The idea is that frequent network clients will exhibit recurrent visits to the target site thus the network they originate from will be frequent as well. This means that the defense system will familiarize itself with networks seen often, and thus notice unknown data patterns originating from not commonly seen networks, such as those from DDoS.

The second method aims to mitigate DDoS based on the geographical location and anomaly detection techniques, using clustering methods. Two algorithms are used for clustering, the simple version, a density-based geographical clustering (DGC), and a variant that aims to improve on this algorithm, called reduced-density geographical clustering (RDGC). At its core, DGC estimates data patterns based on geographical locations, so that if network $z$ is close to a network $x$ that has previously successfully reached a server, then $z$ has a higher chance of reaching the same server. Since this is a clustering algorithm, the clusters must first be defined. The location of the frequent networks are chosen as core points, and all points within a given range threshold of the core, or within a given range threshold of the points belonging to the core, will be grouped together and used to define a cluster. A weakness mentioned by the author, is that DGC might be too simple, allowing too many data points to be clustered. RDGC improves the first algorithm by adding further constraints, such as how many points must be in a cluster lest it be disbanded. After three real life simulations, AFN achieved an acceptance rate of 92.83\% for legitimate traffic, and 99.89\% for illegitimate traffic respectively. RDGC achieved an 77.54\% and 99.93\% respectively. DGC achieved 57.59\% and 99.99\% respectively.

\subsection{Adaptive History-based IP Filtering}
Goldstein et al. \cite{Goldstein-et-al-2008} proposed an adaptive history-based IP filtering algorithm (AHIF) to automate the process of creating and modifying IP filtering rules for anomaly detection. The mitigation system is applied before traffic is allowed through to the target, so a potential DDoS attack is stopped by rules that are prepared before the attack takes place. Bayesian theory is applied to observed data in order to derive a risk function, based on overall loss. Their approach use this to optimize packet filtering rules by minimizing risk under a certain threshold.

A server that can handle $N$ packets in a given time frame needs to keep traffic load below $N$. This means that optimally the packets with the highest probability of being legal traffic can be let through, up to a total of $N$. In practice the AHIF algorithm uses a variable threshold to define how much traffic the server can handle at a given time. This all happens before each packet is sent through to the server, allowing for quick processing time, and stable servers. In the paper, it is argued that a certain amount of DDoS traffic can be allowed through, as long as the server runs smoothly. Therefore the packets with the highest probability of being legal traffic, even if that includes DDoS, will be allowed through. Only the necessary amount of packets to keep a stable server will be dropped. By adapting the threshold variable during runtime to fit the server load, the practical approach reaches similar results to the theoretically optimal approach.

The practical implementation of this method comes in the form of a binary tree keeping track of the IP addresses, with maximum depth of $n$ where $n$ is the most significant bit, at a maximum of 24. The last 8 bit are not used, with the reasoning that IP addresses can be easily spoofed, especially the last 8 bits, and because many different users might have many different addresses within a 24 bit network mask due to for example DHCP\footnote{Dynamic Host Configuration Protocol, an application layer protocol. Used dynamically assign IP addresses to each device on a network.}. Using 24 bits as maximum depth leads to a maximum of $2^{24}$ number of rules. Processing this amount of rules could take a great amount of time. To combat this problem, a system to prune children leaves to a common parent is implemented, effectively compressing the tree in a way that leads to no loss of rules, and a smaller $n$. The system made by Goldstein et al. \cite{Goldstein-et-al-2008} was able to handle about 100,000 firewall rules at a rate of 40,000 requests per second.

\subsection{Flow-based Stacked Autoencoder in SDN}
Niyaz et al. \cite{Niyaz-et-al-2016} presents a DDoS detection system for Software-Defined Networks (SDN), that use deep learning to detect multi-vector DDoS attacks from flow traffic. In a SDN, DDoS attacks happens on the data plane or control plane\footnote{The data plane in a SDN forwards traffic, while the control plane manages what route the traffic will take.}, and their system is thus focused on detecting DDoS on these two planes. Their detection system consists of three modules which they call "Traffic Collector and Flow Installer", "Feature Extractor" and "Traffic Classifier". These modules operate by extracting multiple different headers from TCP, UDP and ICMP packets, and generate a flow to be fed into the DDoS detection system. Each packet belonging to the same flow has the same source and destination IP, the same source and destination ports, and the same protocol type.

Multiple sparse autoencoders (SAE) are used together to form a deep learning network by stacking them after each other \cite{Niyaz-et-al-2016}. There are two ways to do this, one is to encode and decode as usual on the first autoencoder, and feed decoded output to the second autoencoder, and so on. Another method is to encode and decode as usual for training the network, but instead of using the decoded outputs, the latent layer outputs are fed to the next autoencoder in the line. So if we have the raw input $x$ feeding into the first SAE, it will be encoded into the latent layer values $g$ and decoded to $\hat{x}$. The values $g$ are used as input for the second SAE, which encodes to latent layer values $h$. After this, the authors apply a softmax classifier on the outputs of $h$ \cite{Niyaz-et-al-2016}. The final stacked autoencoder consists of two models with a classifier at the end.

For training and testing, their flow generation system extracts and creates a total of 68 features \cite{Niyaz-et-al-2016}, although separated into three protocols, TCP, UDP and ICMP. Attacks were simulated using Hping3, and Niyaz et al. claim to identify individual DDoS classes with an accuracy of about 95\%. The accuracy for differing between normal and attack traffic is claimed to be 99.82\%.

\subsection{Anomaly Detection on the CICIDS2017 Dataset Using LSTM}
\label{related-lstm-2017}

In networking, traffic is sent back and forth between machines, and some packets can be said to have a relation to each other, such as the packets transferred during a three-way-handshake. If multiple machines cooperate to send a DDoS attack, they are considered to be the parts of the same attack, even though there are multiple sources. Using machine learning to detect the relationship between packets or flows could potentially allow mitigation methods to prevent malicious traffic based on learned patterns in IP addresses and traffic frequency. Pektas and Acarman \cite{Pektacs-and-Acarman-2018}, and Radford et al. \cite{Radford-et-al-2018}, proposed two different detection methods in their respective papers using the CICIDS2017 dataset \cite{CICIDS2017} amongst others, which are used in our article as well. This dataset, discussed further in \ref{methodology-2017}, is a modern dataset containing network traffic data, both normal and malicious. It is meant for research, as a benchmark for developing detection and mitigation methods.

A method to group together network flows into two-dimensional arrays was proposed by Pektas and Acarman \cite{Pektacs-and-Acarman-2018}. The proposed system aims to detect malicious network traffic using the combination of a convolutional neural network (CNN) and a long short-term memory (LSTM)\footnote{LSTM is based on a recurrent neural network architecture.} network, to learn spatial-temporal features. Each network flow is grouped based on source and destination IP, destination port, protocol, flow direction, and label. Then the group is fed through the model as a 2D array, where each row represents one flow, sorted by their timestamps. If a group consists of a number of flows less than a given threshold value, the group is omitted. This could be considered as a form of frequency based detection added on top of the detection algorithm using a hyper parameter since malicious traffic, especially DoS and DDoS attacks, rely on sending large amounts of packets or flows in a given time frame. To configure the model network, a Tree-structured Parzen Estimator (TPE) is used, to tune the model automatically based on hyper parameter search. The authors claim the model is able to detect attack traffic at an accuracy of 99.09\%.

Radford et al. \cite{Radford-et-al-2018} proposed a method for anomaly detection using sequence modeling, utilizing a recurrent neural network (RNN) architecture with a LSTM model\footnote{A RNN is a neural network with the ability to remember previous data inputs. LSTMs is an improvement on this concept, increasing the memory capabilities.}. Five different sequence aggregation rules based on the flows provided in the CICIDS2017 dataset are evaluated using a LSTM model. The research in their paper is partially based on their previous work \cite{Radford-et-al-2018-r7-1}. Once a sequence of computer network traffic flows of length 10 has been generated, it is fed through the LSTM model. There the sequence is given a prediction of whether it is normal or malicious traffic, based on an outliers score \cite{Radford-et-al-2018} and measured using mean area under the curve (AUC). For baseline comparison, a simple frequency based method for outliers detection was used. The sequence aggregation with an LSTM model proved slightly better in a few cases, but was mostly on-par with the frequency based model, or worse. Whether this was because of the LSTM model or the sequence aggregation methods was deemed uncertain.

\subsection{Malicious traffic detection using entropy-based techniques}
Entropy, from information theory, is a measurement of uncertainty or disorder in a system, often called Shannon entropy \cite{Shannon-2001}. It is a value representing the average rate of information being drawn from a stochastic source of data. A source of data producing an entropy value closer to 1 when normalized is considered hard to predict, and a value closer to 0 is considered easier as there is less uncertainty. Behal et al. proposed the usage of generalized information entropy (GE) and generalized information distance (GID) for separating so called low rate DDoS (LW-DDoS) and high rate DDoS (HR-DDoS) from normal traffic and flash events (FE) \cite{Behal-and-Kumar-2017}. The idea is to group network traffic into sets, where packets were grouped together in 10 second time frames. The entropy and information distance is then measured for each set. It was discovered that DDoS traffic flows had more similar traffic, as their IP addresses are more closely grouped in relation to time. This leads to lower entropy values within a set containing more DDoS traffic, and higher information distances between normal and DDoS traffic.

To better understand the relationships between different features in network traffic and how they can be used for anomaly detection, Nychis et al. published an empirical evaluation on the subject, using entropy \cite{Nychis-et-al-2008}. Features were gathered from bi-directional flow data.

The relationships between the features source IP, destination IP, source port, destination port, in-degree, out-degree and flow size distribution (FSD) were measured using entropy, and given correlation scores. Note that FSD is the packet per flow measurement. The study found high correlation between certain features, perhaps most notably between ports and IP addresses. However it was also found that the correlation between ports and addresses had limited usability for anomaly detection, and argued that they are ineffective for both scanning and flood type attacks. Interestingly, the FSD and degree distribution scores had some success in detecting anomalies, and the entropy scores between normal traffic and malicious traffic had a noticeable difference \cite{Nychis-et-al-2008}.

\subsection{Flow-based DoS attack Detection with techniques based on Computer Vision}
Autoencoders from machine learning can be used for anomaly detection, by separating malicious from normal traffic using pattern recognition. If a data input fed into the model is not recognized, it would be considered an anomaly. Tan et al. proposed the use of computer vision techniques for anomaly detection in network traffic, specifically for DoS attack traffic \cite{Tan-et-al-2015}.

Features from inbound network traffic is fed through the detection system, and stored as one-dimensional feature vectors called records. In computer vision, earth mover’s distance (EMD) can be used to detect dissimilarities between two images. To apply this idea to their system, Tan et al. transforms inbound records into two-dimensional matrices, similar to images. Profiles for normal network traffic is generated based on multivariate correlation analysis (MCA) from previous work \cite{Tan-et-al-2014-r13-1}. MCA works to find correlation between features and generating normal records from the inbound records coming from the datasets. Then a reformulation of EMD \cite{Ling-and-Okada-r13-2} is applied to the generated record matrices, measuring dissimilarities between each record. Any unmatched records will be determined as attacks. Evaluation of the detection system on the KDD’99 \cite{KDD99} dataset was reported to have achieved 99.95\% accuracy, and on the ISCXIDS2012 \cite{Shiravi-et-al-2012} dataset to be 90.12\%.

\subsection{Anomaly Detecting With Hidden Semi-Markov Model}
Hidden Markov models (HMM) have a variety of different applications, such as research on time series data or for sequence recognition. Xie and Yu proposed a solution for detecting application layer DDoS attacks as anomalies by learning from user behavior on web pages with a hidden semi-Markov model (HsMM) \cite{Xie-and-Yu-2009}. HsMM is an extension of HMM that adds an explicit state duration, and is made for live training. The HsMM is a behavior model that learns from looking at normal user behavior in regards to how they behave when browsing a given web page, using the address bar, hyper links and reading web page content. From this, a normal user is defined with a mean entropy value which will be used for comparison with the filter. Requests from an outside source reach the victim web page, and are stored over time as a request series, or as it is called in the paper, a HTTP request sequence, similar to how network packets would be handled by a RNN. The average entropy of this sequence is calculated at the detection system filter, and used for comparison with the entropy from the generated user behavior characteristics, made by the HsMM. The research and experiments in the paper were only tested on application layer DDoS attacks, but showed promising results, with as much as 98\% detection rate.

Not many solutions for detecting or mitigating DoS and DDoS attacks focus on learning from user behavior. It could be a potential avenue for further research.

\section{Autoencoders}
\label{background-autoencoders}

An autoencoder is a deep learning framework that utilize an NN framework to perform a variety of different tasks, primarily for unsupervised learning, where the backpropagation target values are set to be equal to the input. Variations of an autoencoder allows for classification, anomaly detection, and generative tasks, amongst other uses. Autoencoders are feedforward networks, meaning that the representative ANN is a directed acyclic graph, and that they use back-propagation for training. The composition of an autoencoder always has at least two parts, an encoder and a decoder. The encoder encodes the input $x$ to a hidden layer $h$, selecting which dimensions to learn from with the function $h = f(x)$. The decoder tries to make a reconstruction, $r$, of the input from the hidden layer, with the function $r = g(h)$. The concept of autoencoders have been around for over a decade, with one example dating back to 1988 by Bourland and Kamp\cite{Bourlard-and-Kamp-1988}. Historically the hidden layer mapping has been deterministic, but more modern solutions use stochastic mapping, with the functions $p_{encoder}(h | x)$ and $p_{decoder}(x | h)$\cite[p.~499]{Goodfellow-et-al-2016}.

The reconstruction of the hidden layer is not a perfect replication, nor should it be, and is recipient to noise. For an autoencoder model to be useful, it needs to generalize over training data, lest we end up with a model that performs poorly on foreign data, as we will see later in \cref{analysis}, \nameref{analysis}. One of the major pros to using this kind of model is its capability to determine which parts of the input are important, by forcing it to learn the useful properties of the data it is given during training. In a way, when using an autoencoder we are often more interested in the encodings of the data and the latent layer representations, than we are of the actual reconstruction of the decoder. A good autoencoder is one that is able to properly select which dimensions from the input to use in the hidden layer and to what degree. In such a way that the decoder is able to produce a good approximation on as few dimensions as possible. There are mainly two methods for dimension selection used in autoencoders, dimensionality reduction and regularization. Dimensionality reductions is when each hidden layer in the model contains fewer nodes than the preceding layer. This forces the model to select the most important features from the previous layer, and throw away the least important ones. Regularization selects the nodes with the greatest positive impact on the model's result, and lessens the impact of the other nodes. There are multiple, different regularization techniques, and some of them will be discussed in the following sections.

The imperfect reconstruction of the input data can be both an advantage and a disadvantage in machine learning. Autoencoders, like other machine learning algorithms, must be applied to problems they are fit to solve. Reconstructing the input is only possible if the model has seen similar data during training. If an autoencoder is only given images of cats during training, it will not be able recognize images of e.g.\ birds. For this article, it means that the model will not be able to classify a DDoS or DoS attack if we only train it on normal data. However, because of this exact property, an autoencoder could be used as an anomaly detector, essentially differing the two classes by only recognizing normal data, and unable to recognize anomalies.

\begin{algorithm}[htbp]
\SetAlgoLined
    \For {each input $x$}{
        Feedforward $x$ and compute activations for each layer\;
        Sample the hidden layer $z$\;
        Obtain output $y$\;
        Measure deviation of $y$ from $x$\;
        Backpropagate to update weights and node values\;
    }
    \caption{A simple autoencoder}
\end{algorithm}

\subsection{Sparse Autoencoder}
\label{sparse-autoencoder}
A sparse autoencoder (SAE) is a regularized variation of an autoencoder with potentially more nodes in the hidden layer than in the input layer. This means that to extract useful features to learn from at the input layer, the SAE appends a regularizing function to the normal loss function of an autoencoder\cite{ng-et-al-2011}\cite[p.~502]{Goodfellow-et-al-2016}. Because of this, the hidden layer only has a few select nodes active at a time, forcing the SAE to learn the most useful properties of the input. Furthermore, each type of input activates different nodes in the hidden layer. There is normally overlap between properties of for example network packets, hence they would often activate some of the same nodes, but the point of a SAE is to only activate the relevant nodes in the hidden layer, customized for those types of input. This is called the sparsity constraint. In theory, this means that the total amount of active nodes in the hidden layer could be as large as the different properties of the input, leading to some SAEs to have larger hidden layers than input.

The loss function of an autoencoder can be described as $L( x, g( f(x) ) )$, where $f$ is an encoder and $g$ is a decoder. The goal is to minimize the difference between the input and the output. This is done by penalizing $g( f(x) )$ for being dissimilar from the input x. A sparse autoencoder adds a sparsity penalty to the loss function, which is commonly done in one of two ways. One way is to use L1 regularization, also called Lasso Regression\footnote{From statistics. Lasso Regression shrinks the coefficient of less important features, reducing their impact.}, which ends up looking like this, $L( x, g( f(x) ) ) + \Omega( f(x) )$ where $\Omega( f(x) )$ is the regularization term. Note that regularized networks typically regularize the weights that connects the nodes. SAEs however, apply regularization on the activations of the nodes. The term $\Omega( f(x) )$ where $f(x)$ is the hidden layer $h$, can be simplified as $$\Omega(h) = \lambda \displaystyle\sum_{i} |h_i|$$

Here $\lambda$ is a hyperparameter, and the following formula is the absolute sum of all activations of the nodes $i$ in the hidden layer.

Another way to apply a sparsity penalty is by using Kullback-Leibler divergence, or KL-Divergence for short. KL-divergence is a measure of the divergence between two probability distributions, used to measure their similarities, or dissimilarities. Given the probability distributions $p$ and $q$, KL-divergence is a measure of how well $q$ approximates $p$, by calculating the cross-entropy $H(p, q)$ minus the entropy $H(p)$, to get the KL-term $D_{KL}(p || q) = H(p, q) - H(p)$. The KL-divergence is a central part of the Variational Autoencoder, which will be explained in detail in \cref{background-vae}. Niyaz et al. proposed using a stacked sparse autoencoder in order to detect DDoS attacks in software defined networking (SDN)\cite{Niyaz-et-al-2016}. In their paper they use KL-divergence for their sparse autoencoders to put a constraint on the hidden layer to maintain low average activation values. They also present a method for layering the sparse autoencoders to use as a classifier. The sparsity penalty term can be written as $$\beta\displaystyle\sum_{j=1}^{N} KL(p||\hat{p}_{j})$$

Where $\beta$ is a hyperparameter to adjust the sparsity penalty term. $\hat{p}_{j}$ is the average activation value of a hidden node $j$ over all the training inputs, and $p$ is a Bernoulli random variable\footnote{From statistics, a Bernoulli distribution is the discrete probability distribution of a random variable with boolean values.} that represents the ideal value distribution. The KL-loss is at a minimum when $p=\hat{p}_{j}$.

\subsection{Denoising Autoencoder}

The principle of a denoising autoencoder (DAE) is simple. As was explained earlier, an autoencoder aims to optimize the loss function by minimizing the difference between the input and the reconstructed output. A DAE is regularized and can be overcomplete, meaning it uses regularization to extract useful features. Differing from a SAE, a DAE does not apply a penalty to the loss function, but instead changes the reconstruction error term. The loss function is changed from the vanilla version $L( x, g( f(x) ) )$ to $L( x, g( f(\tilde{x}) ) )$, trying to optimize on $f(\tilde{x})$ instead of $f(x)$, where $\tilde{x}$ is a corruption of the input. The output is then compared to the uncorrupted input. DAEs differ only in a minor way from vanilla autoencoders. By adding noise to the input data, a DAE is forced to learn the most prominent features to be able to reconstruct the original input, essentially learning to remove the noise.

A DAE is generally used for creating outputs free of noise. If applied to images, it is possible to reconstruct missing parts, for example if there is a lens-flare covering part of the image, the DAE could provide a copy of the image without the flare. A DAE could also be used to restore missing or hard to read letters and words in a text, or unclear sound could be repaired to sound cleaner. Vincent et al.\cite{vincent-et-al-2010} presented a simple stacked DAE and tested it on a variety of different datasets, including the MNIST image dataset\cite{MNIST}. What they showed us is that by using a denoising criterion, we can learn useful higher level representations of the input data. In the paper by Vincent et al.\cite{vincent-et-al-2010}, the input is corrupted with simple generic corruption processes, and they mainly perform tests on image and audio samples. A denoising criterion could be useful to help the learning process of an autoencoder, to perform DDoS and DoS classification based on the output loss function.

\subsection{Variational Autoencoder}
\label{background-vae}

The Variational autoencoder (VAE) introduced by Kingma and Welling\cite{Kingma-and-Welling-2013} is a generative model and use the same encoding as a normal autoencoder, the difference being in how the latent variables are handled. It is based on variational Bayes\footnote{Bayes is here referring to Bayesian inference. Variational Bayes methods are used for approximating intractable integrals arising from Bayesian inference.} which is an analytical approximation to the intractable posterior distribution of the latent variables. It is used to derive a lower bound for the marginal likelihood of the observed data. The VAE present a change to variational Bayes, by reparameterization of the variational lower bound, which is called the Stochastic Gradient Variational Bayes (SGVB) estimator. Since the VAE is a generative model, its primary strength lies with how well it can create new outputs based on learned features from training. It is in addition to this also possible to extend on a VAE to use it for data classification. An example of this can be seen in the VAE of Y. Pu et. al.\cite{Pu-et-al-2016} using convolutional layers of the encoder and decoder to perform semi-supervised learning on image datasets.

The main difference between a traditional autoencoder and a VAE is how they use the layer between the encoder and the decoder, commonly referred to as the latent layer. An autoencoder use the latent variables directly and decode them to use for comparison between the input and output. A VAE will instead encode into two vectors of size $n$, the vector of means $\mu$, and the vector standard deviations $\sigma$. A sampled vector is created from a collection of elements $z_{i}$ that is assumed to follow a Gaussian distribution\footnote{A Gaussian distribution, also called a normal distribution, is a function that represents the distribution of a group of random variables as a symmetrical bell-shaped graph with the mean value at the center.}, where each element $i$ comes from the $i$-th element in $\mu$ and $\sigma$. Thus we can write each element in the sampled vector as $z_{i}\sim{\mathcal{N}(\mu_{i}, \sigma_{i}^{2})}$\footnote{From statistics. It reads: "z drawn from a normal distribution with mean $\mu$ and standard deviation $\sigma$"}.

\begin{figure}[htbp]
    \centering
    \includegraphics[scale=0.45]{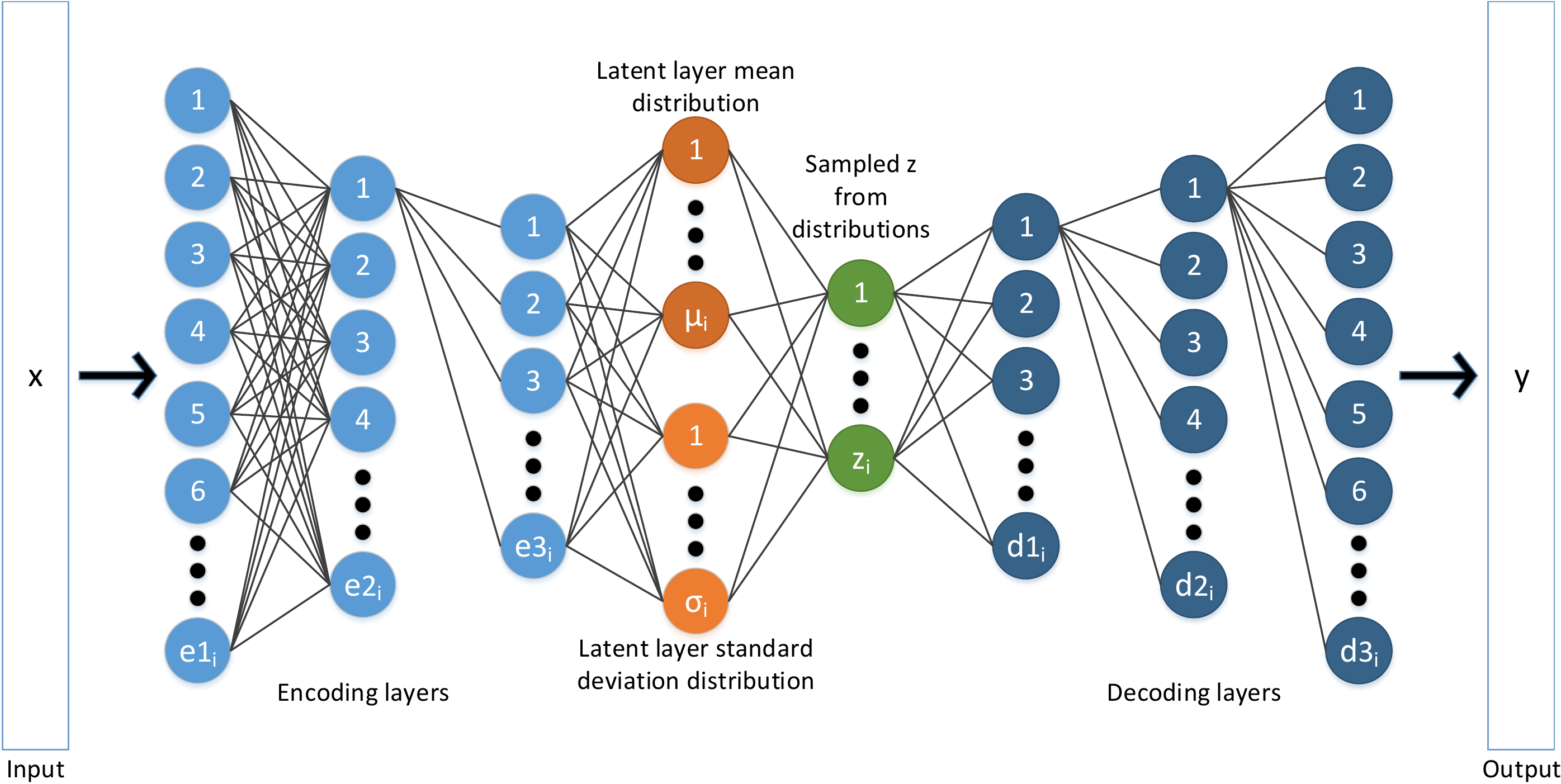}
    \caption[Variational Autoencoder With Fully Connected Layers]{A VAE with fully connected layers. Each vertical line of nodes represent one layer. The latent mean and latent variance layers are two separate layers, which the layer z sample from. All nodes in a layer have weights connected to the nodes in the adjacent layer, like seen in the first two encoding layers. (For simplification, only some of the weights are drawn in this figure).}
    \label{fig:background-fcvae}
\end{figure}

\subsubsection{VAE in Detail}
\label{background-vae-detail}

Let $z$ be a latent representation of the unobserved variables, and $g(z)$ a differentiable generator network. $x$ is sampled from a distribution $p(x; g(z))$, which can be written as $p(x|z)$. Here $p(x|z)$ represents a probabilistic decoder presenting a distribution over the possible values of $x$ given $z$. When using the probabilistic decoder, we get an observation $x$ from the hidden variable $z$. However, what we want is to infer the characteristics of $z$, thus we need $p(z|x)$ and the integral marginal likelihood $p(x)$. The problem is that $p(x)$ is intractable, meaning that we cannot evaluate or differentiate the marginal likelihood\footnote{For further details about this problem, see the original paper by Kingma and Welling\cite{Kingma-and-Welling-2013}.}. The solution to this is to create an approximation of the true posterior with another distribution $q(z|x)$, which will be the recognition model, a probabilistic encoder. We can use KL-divergence to measure the difference between these two probability distributions, as discussed earlier in \cref{sparse-autoencoder}. The goal is to minimize the difference in order for the two distributions to be as similar as possible, we then get $\min{KL}(q (z|x) || p (z|x) )$. This can be used to maximize the lower bound $\mathcal{L}(q)$ of the marginal likelihood of the observed data, so that we get $\mathcal{L}(q) = E_{z\sim{q(z|x)}} \: log \: p(x|z) - D_{KL} (q (z|x) ) \, || \, p (z) )$. The first term $E_{z\sim{q(z|x)}} \: log \: p(x|z)$ represents the reconstruction term, and $D_{KL} (q (z|x) ) \, || \, p (z) )$ represents the Kullback-Leibler (KL) term. It ensures that the approximate posterior q is similar to the true posterior p.

\subsubsection{Reparameterization Trick}

We have now seen the basic explanation of how a VAE works, and the math behind it. To fix the problem with the integral marginal likelihood $p(x)$, the "reparameterization trick" is introduced.

\begin{figure}[htbp]
    \centering
    \includegraphics[scale=0.8]{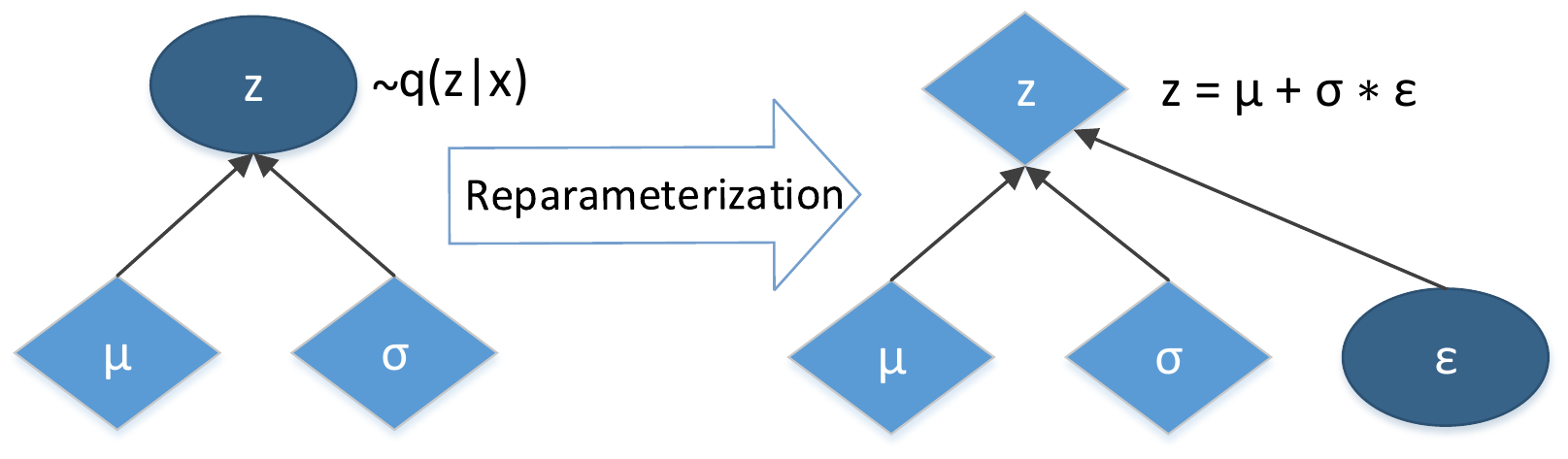}
    \caption[Reparameterization Trick]{Diamond shapes represent deterministic dependencies, and oval shapes represent random variables.}
    \label{fig:background-repa}
\end{figure}

Since an autoencoder relies on a NN to forward data and backpropagate for training, we should not have a latent variable $z$ be a random variable sampled from $q(z|x)$. A NN generally has poor performance when performing backpropagation on random variables. This would lead to the decoded output being too different from the input. We know that the probabilistic encoder $q(z|x)$ is Gaussian, because it produces a distribution over the possible values of $z$ from a data point $x$. In other words, $q(z|x) = {\mathcal{N}(\mu, \sigma^{2})}$. Now let $\varepsilon$ be an auxiliary noise variable $\varepsilon \, \sim{\mathcal{N}(0, 1)}$. We can reparameterize the encoder $q(z|x)$, so we get $z = \mu + \sigma \cdot \varepsilon$ as seen on the right hand side of figure \ref{fig:background-repa}.

\section{The proposed Deep Learning Algorithms for Attack Mitigation using Variational Autoencoders}
\label{solution}
DoS and DDoS mitigation has been researched for many years, and several, different approaches have been created. Popular methods for mitigation include pattern recognition, similar to how one would detect viruses, or detection of malicious sources based on network traffic frequency. Pattern recognition mitigation systems have proven effective in certain scenarios, such as when a victim is the target of a known attack, but is also known to have several drawbacks. These kinds of systems are prone to human error, and require constant maintenance to operate, needing code updates every time a new attack surface, or whenever a known one is altered. Traffic frequency based mitigation systems work by blocking network traffic based on high traffic frequency, or allotting a certain amount of bandwidth to each connected IP. Monitoring traffic frequency can be effective if there is a lot of traffic from one source, allowing a mitigation system to block attackers that take up too much of a system's resources. A problem with this approach is that sometimes normal users might be regarded as attackers. For example if they try to refresh a website many times because of slow loading. Checking for frequency alone could also let certain types of attacks through, such as DDoS attacks with a large network of machines with different addresses.

With deep learning, it is possible to let the mitigation system filter normal from malicious traffic autonomously. Network traffic can be fed through a deep learning algorithm, which filters individual packets or flows based on learned features. For this article, incoming network packets will be transformed into traffic flows, before being fed to the deep learning algorithms. We propose two separate deep learning algorithms to filter network traffic flows, Latent Layer Classification on a Variational Autoencoder (LLC-VAE), and Loss Based Detection on a Variational Autoencoder (LBD-VAE).
These two deep learning algorithms learn patterns by themselves, instead of relying on older techniques where the attack patterns must be inserted  manually.

The contribution of this article is to research how, and how well the proposed deep learning approaches, the LLC-VAE and the LBD-VAE, can filter malicious from normal traffic. The LLC-VAE and LBD-VAE will be used to learn from a few types of DoS and DDoS attacks. While there exist many more types of malicious computer network traffic, the samples in these datasets will be used to analyze whether the two proposed approaches can reliably be used as DoS and DDoS mitigation systems. For the remainder of this section, we will present how the two proposed approaches are designed, as well as discuss different options for designing them.

\subsection{Motivation}
As we have mentioned earlier, network packets moving between a client and a server can vary greatly in shape and form even though they follow the same protocols. Likewise, many network packets can be very similar, with only small differences separating them. The same goes for packets belonging to DDoS and DoS attacks, they can be very similar to other attack packets, and normal packets. One of the goals we aim at achieving in this article, is to be able to efficiently separate DDoS and DoS attacks from normal traffic. An autoencoder, particularly a VAE, could be a well-suited tool for this. It is a challenging problem to separate malicious and normal traffic, sometimes with no more than a minute difference in time separating them, hence we need a tool that is able to find small details and differences, as well as finding features that are important for the seperation. Standard autoencoders, and other implementations of autoencoders, learn features from input data in a discrete fashion when encoded to the latent layer. A VAE on the other hand encodes features as probability distributions with the use of variational inference\cite{Kingma-and-Welling-2013,blei-et-al-2017}, which for  our case causes similar packets to be encoded and decoded in a similar manner. We sample from this distribution to get the latent attributes.

A common problem with machine learning in general is gathering reliable data to train and test on. What is more, deep learning algorithms require large amounts of data to generalize and train robust and deep features. While a VAE needs large amounts of data just like many other deep learning methods, one of its strengths relies  in its ability to generalize over similar features, and ignore noise. With a VAE, one can create a model capable of learning smooth latent state representations of the input. There are two parts to the learning process of a VAE, the reconstruction loss and the KL-divergence loss. Using purely reconstruction loss causes it to behave like a normal autoencoder, simply reconstructing the input to the output, with potentially large gaps in between different classes. Using purely KL-divergence loss we end up with outputs which all use the same unit Gaussian, causing different classes of data to be grouped and mixed between each other. Other autoencoders which do not rely on variational inference have been used for both DDoS and DoS detection, and was discussed in \cref{related-work}, \nameref{related-work}. Because the VAE combines these two loss terms into one, a potential use for it is to group different classes of data with similar features close to each other. This is one of the reasons why a VAE is a generative model, since it can extract and generate new data based on data points with likeness to each other. For detecting and classifying DDoS and DoS, the generative part can be ignored, instead of using the VAE to remove noise, and generalizing over similar features in a manner that allows understanding of data not included during training of the model.

Generating new samples using a VAE is straightforward. One simply removes the encoder part after training, leaving $z$ to sample from $\mathcal{N}(0, 1)$\cite{Doersch-2016}. We do not need to generate new data from the learned features of DDoS, DoS and normal data since we are using it for detection and mitigation. However, it should be noted that a general problem with the VAE architecture is that the normal and generated outputs come out blurry, or noisy\cite{Kingma-and-Welling-2013,Doersch-2016}. The model will sometimes ignore less prevalent features of the input, potentially causing this blurriness. This weakness is most obvious when looking at images as outputs. How much this affects datasets consisting of network packets, and how important it is for training and testing is difficult to tell.

\subsection{First Proposed Approach: Latent Layer Classification on a Variational Autoencoder (LLC-VAE) }
\label{methodology-llc-vae}

The first proposed approach, Latent Layer Classification on a Variational Autoencoder (LLC-VAE), utilize the strength of the variational autoencoder \cite{Kingma-and-Welling-2013} as the underlying architecture for a latent layer classification network. Based on the ability of the variational autoencoder to learn latent representations of various classes of a dataset, the LLC-VAE approach aims to classify different types of computer network traffic, and separate normal traffic from malicious. The performance will be documented in \ref{analysis}, \nameref{analysis}, together with how the model performs on different settings.

A representation of the LLC-VAE deep learning model can be seen below in figure \ref{fig:methodology-llc}. First, one flow from the dataset is loaded into memory and transformed to readable format by the model. One flow is represented as one feature vector, and together they are grouped into mini-batches before being fed into the encoder. The encoder performs dimensionality reduction on the mini-batch over multiple layers, further transforming the feature vectors, until they have been encoded to the latent layers\footnote{The hidden layers between the VAE encoder and decoder are called latent layers in this article.} of means and standard deviations. Based on this the latent layer $z$ is sampled, and the KL-Loss value is produced. This output, now represented as a vector of nodes with reduced dimensions, is sent to a fully connected layer that outputs the unscaled class predictions. The softmax function, short for softargmax \cite{Goodfellow-et-al-2016}, is applied on the class predictions, so that the nodes are normalized to a legal probability density function (PDF), where each node represents a single class. To optimize the predictions, we use cross entropy over the softmax predictions, to generate a numerical loss value which is called \texttt{prediction loss}, or P-Loss for short. The latent layer $z$ also feeds its vector of nodes to the decoder. The decoder aims to accomplish the opposite of the encoder, increasing the dimensionality through multiple layers to generate a reconstruction of the original feature vector. The reconstructed feature vector is an approximation of the input. These two vectors are compared to generate a reconstruction loss, or R-Loss for short, by using mean squared error: 
        $$L(x) =  \frac{1}{n} \sum_{i=1}^n (y_{i} - x_{i})^{2}$$
Here $x$ represents the model input, $y$ the output, and $n$ the number of features. The values generated from each of the loss functions, R-Loss, KL-Loss, and P-Loss, are combined and optimized over. We use the Adam optimizer by Kingma and Ba \cite{Kingma-and-Ba-2014}, to perform stochastic gradient-based optimization of the model. The optimizer will backpropagate through the network, updating the weights between each layer based on the total loss.

\begin{figure*}[htbp]
    \centering
    \includegraphics[scale=0.7]{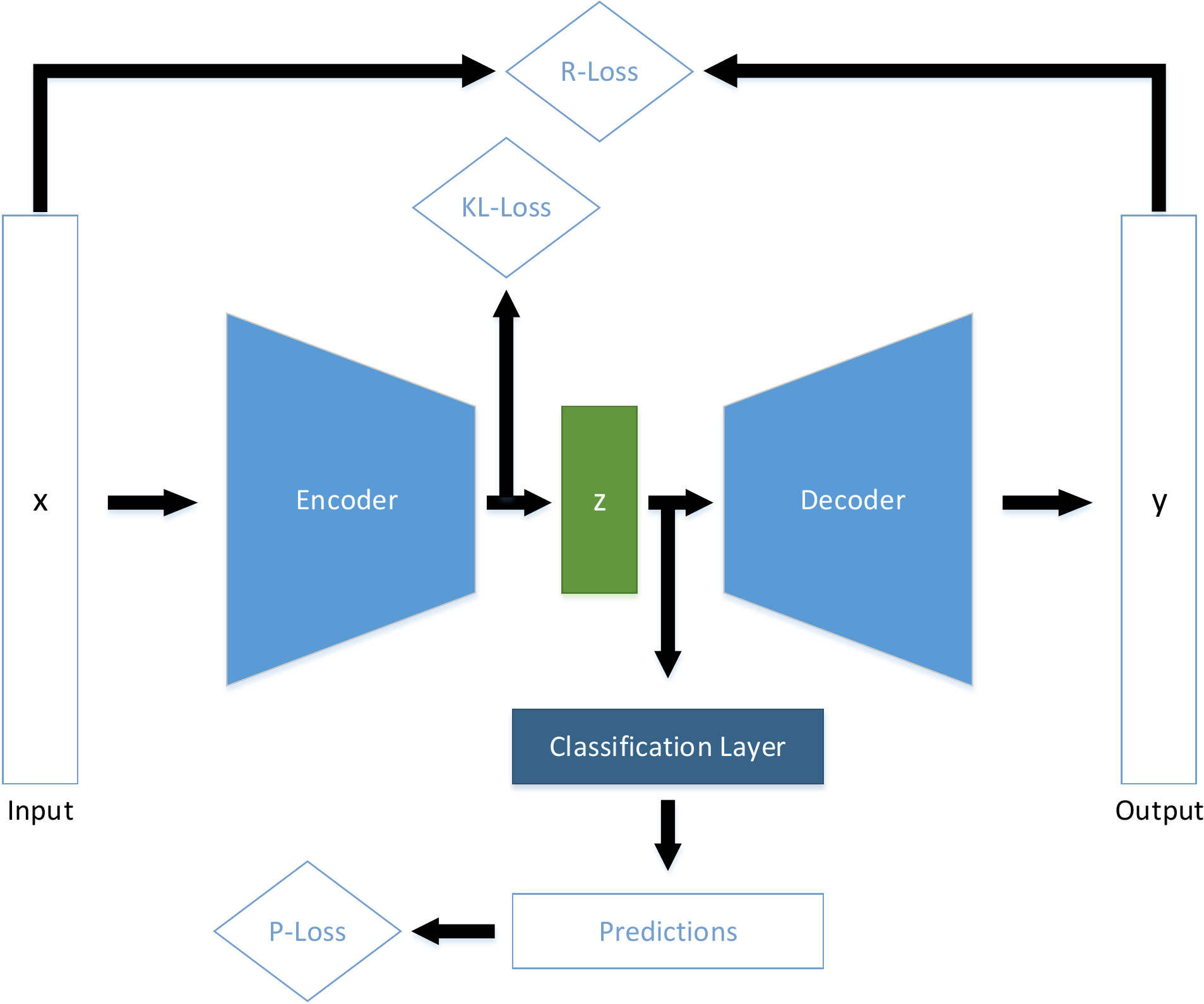}
    \caption[LLC-VAE Model]{Latent layer classification on a VAE. Rectangles represent node layers. Diamond shapes represent the loss values. The trapezoids represents the shrinking nature of the encoder, and the expanding nature of the decoder. The classification layer is a fully connected layer with a softmax activation function.}
    \label{fig:methodology-llc}
\end{figure*}

\paragraph{Deciding What Type of Layers to Use}

A variational autoencoder is inherently a deep learning algorithm, with multiple hidden layers. There is a minimum of three hidden layers, excluding the input and output layers, with potential for adding more. The encoder and decoder has a minimum of one hidden layer each, while there is one hidden layer $z$ as seen in figure \ref{fig:methodology-llc}, also called the latent layer. There are various types of layers available to use for the encoder and decoder in a VAE, some of which will be explored here. Deciding on how many hidden layers that should be present in the encoder and decoder will be decided through tuning in \ref{analysis}, \nameref{analysis}.

\paragraph{Fully Connected Layers}

A fully connected layer is the layer typically associated with classification problems in multilayer perceptrons, but is also available to use with other neural networks, such as a VAE. The implementation is fairly simple, all nodes in a layer is connected to all nodes of each adjacent layer, where each node stores the node values, and the connections store the weight values. These values are used to predict the outcome of a given problem during the training process. The outputs of this layer type are calculated with the linear function $\text{outputs} = \text{activation}(\text{inputs} \cdot \text{kernel} + \text{bias})$, where kernel is a weight matrix created by the layer. Using multiple fully connected layers will allow for the classification of nonlinear problems \cite{Irie-and-Miyake-1988}. 

\paragraph{Recurrent Layers}

Recurrence is primarily used for sequence modeling, that is to say a recurrent layer is able to remember what has been seen previously. Intuitively, implementing memory in a neural network to observe relationships between different traffic flows seems like a good idea, allowing a deep learning model to remember previously seen, as well as ongoing, attacks and their sources. This method has shown some success, as discussed in \ref{related-work}, \nameref{related-work}.

\paragraph{Convolutional Layers}
\label{methodology-conv-layers}

Convolutional layers can help greatly with classifying network traffic flows, with the ability to create complex feature abstractions from simpler ones in order to understand complex feature relationships. Convolutional layers calculate the outputs linearly in a sliding window manner, using the convolution operation often denoted as $(x * w)(t)$, where $x$ is the input function at a given time $t$ and $w$ is the weighting function \cite{Smith-1997} \cite{Goodfellow-et-al-2016}. Convolutional layers have the ability to capture spatial and temporal dependencies, and are therefore excellent for usage in object detection \cite{Sermanet-et-al-2013}, and image recognition \cite{He-et-al-2016}. Convolutional layers have proven effective for a variety of different problems, including multiple classification problems, and is a strong candidate to use with a VAE. While a fully connected layer learns a representation of an input based on each feature, a convolutional layer selects important features with sliding window detectors, making it better at ignoring redundant information and learning useful representations through multiple abstract feature layers.

\begin{figure}[htbp]
    \centering
    \includegraphics[scale=0.4]{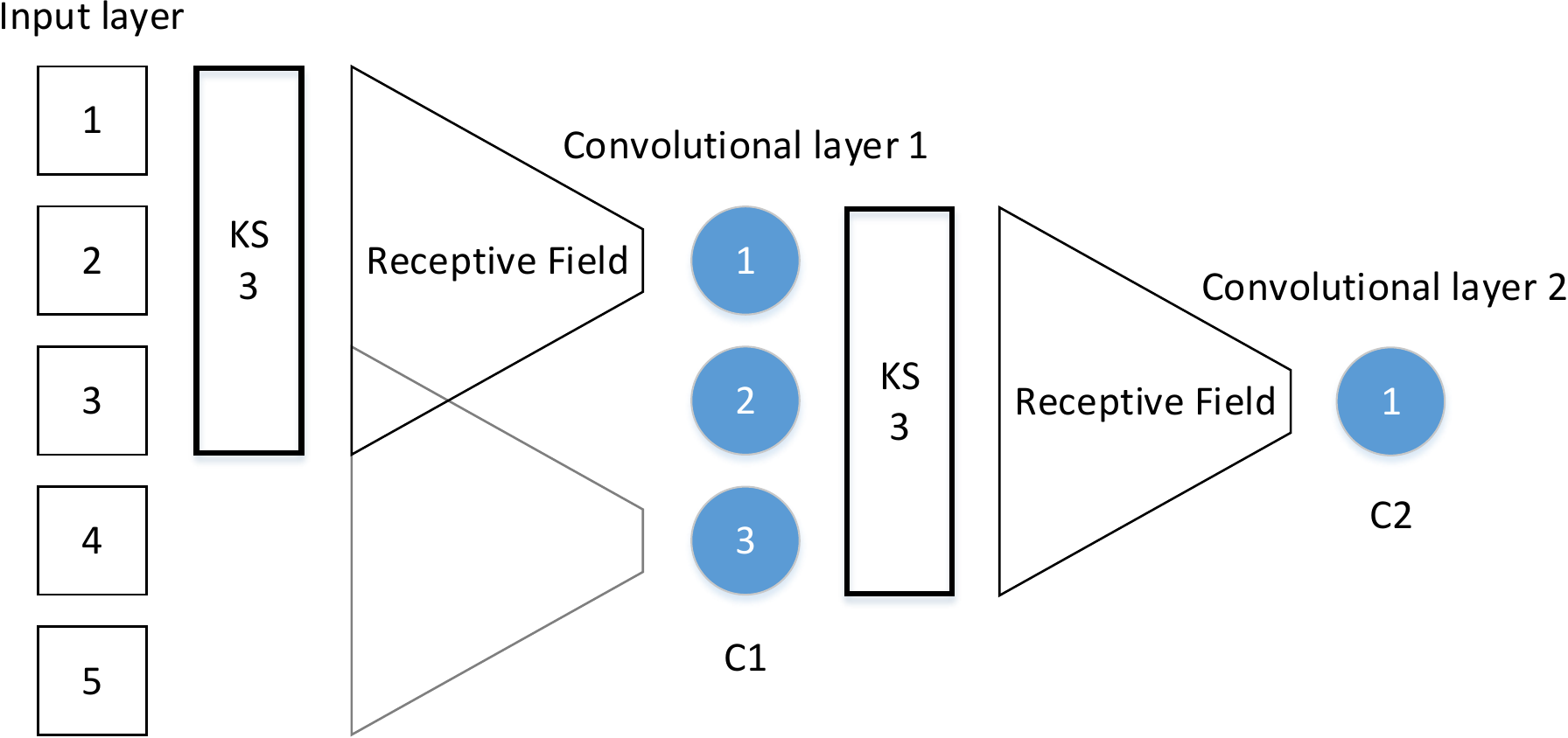}
    \caption[Convolutional Layers in an Encoder]{This figure is a simplification of how the encoder part of the VAE works with convolutional layers. The decoder is similar, but performs the same operations in reverse. The rectangles represent the kernels of size 3 nodes across the height and width dimensions respectively. The squares from the input layer represents 5 features. The trapezoids represents the concept of \texttt{Receptive Fields} in ML. The circles represent the abstract feature nodes of the convolutional layers.}
    \label{fig:methodology-convolutions}
\end{figure}

A convolutional layer learns abstract features from the input layer with a technique called \texttt{sliding window}. A matrix, called a kernel, performs mathematical operations on the node values in a layer, beginning with the feature nodes from the input layer. From the example in figure \ref{fig:methodology-convolutions}, there can be seen a total of 5 features from the input layer. The traffic flows used in this article are one-dimensional, hence when defining the kernel size we only need to define the size of a single dimension. From the example, each node in the first convolutional layer learns from nodes 1, 2, and 3 in the input layer. The second node in the first convolutional layer learns from input nodes 2, 3, and 4, while the last node learns from input nodes 3, 4, and 5. Put in another way, each node from the first convolutional layer has a receptive field of size 3. Likewise, the second layer also has a local receptive field of 3. When using multiple convolutional layers, each node in each layer has a local receptive field of a given size, but the effective receptive field size increases every time a layer is added. For example, the first layer in figure \ref{fig:methodology-convolutions} has a receptive field size of 3. The following layer will learn from the feature abstractions of the first layer, causing it to have a local size of 3, but an effective size of 5, in the input layer. This means that the node in the second layer has learned an abstraction from the relations between all the nodes in the input layer.

\paragraph{Regularization}

The term regularization has a wide range of uses. Regularizing a deep learning model means preventing overfitting, avoiding exploding and vanishing gradients, and generally keeping the training phase stable and improving various issues. An autoencoder inherently performs a form of regularization. A VAE specifically, performs dimensionality reduction, and as a result, forces the model to choose the most important features through node selection.

Weight regularization is the addition of a penalty term, to prevent exploding gradients. It can be applied to the hidden layers of the encoder and decoder when using convolutional or fully connected layers. Exploding gradients is a problem for many different neural network architectures, where the layer weights grow out of control causing various issues with the model loss. This could for example be when the loss does not gain traction and does not improve, or the loss could end up with a NaN value, due to floating point overflow. We use the weight decay regularization technique, L2, typically called ridge regression, on the kernel values of the convolutional layers. This will encourage the layer weights to grow towards smaller values around 0. Weight regularization might not be necessary, but does not harm the model performance, and so there is no reason not to implement it.

A regularizing layer can be added after each layer in the encoder and decoder. Two different regularizing layers can be used for the approaches in this article, dropout layers \cite{Srivastava-et-al-2014} or batch normalization layers \cite{Ioffe-et-al-2015}. Adding either of these to an encoder and a decoder is meant to ensure stable learning, prevent overfitting, and to improve the exploding and vanishing gradients problem. Dropout is a technique that gives a keep probability to each node in a given model network that updates during training. If a node has a keep probability that has sunken below a given threshold, the weights will be multiplied with 0 during the forward pass, causing the gradients to become 0 during backpropagation. As a result of this, a number of nodes in the model network will effectively be removed, forcing the model to learn more robust features and teaching each layer to rely on different nodes from the previous layer. Batch normalization can be used instead of dropout to prevent overfitting, as well as to prevent vanishing and exploding gradients.

\paragraph{Latent Layer Classification}

The encoder and decoder of the VAE can be modified to use different types of layers. To represent the vectors $\mu$, $\sigma$, and $z$, we are using fully connected layers. The output from $z$, as seen in figure \ref{fig:methodology-llc}, is sent to another fully connected layer which attempts to predict the classes that the different flows belong to. A softmax cross entropy loss function is used to improve the classification by performing cross-entropy between the predicted classes, and the labeled true classes. In the dataset CSECICIDS2018 \cite{CSECICIDS2018}, there are a total of 8 classes, each used for training and validation by the LLC-VAE. Either of the cross-entropy functions, softmax or sigmoid, could be utilized as loss functions for the LLC. Each flow can only belong to one class, and the higher the probability of a flow belonging to one specific class, the less the probability of that flow belonging to another class. That is to say the class predictions are dependent on each other. For the LLC-VAE we will be using the softmax activation function.

The dataset CICIDS2017 \cite{CICIDS2017} have two fewer DDoS classes, for a total of 6. This dataset will be used for testing. Having fewer classes for the test network, compared to the training network, is not a problem, as these 6 classes are present in both datasets, and comparison can be done on these classes alone. However, the test results will give less comprehensive results compared to the validation results. Classifying specific DoS and DDoS attacks could pose a different problem. Although the LLC-VAE is able to classify specific attacks and normal computer network traffic, doing so could lead to less accuracy when differing between normal and malicious traffic. When the model needs to do more than anomaly detection, it is required of it to train more precisely tuned weights for the different flow features. This could mean that the model achieves more precise predictions for specific attacks at the cost overall accuracy. For this reason, both a multi class variant, and an anomaly detection variant will be examined in \ref{analysis}, \nameref{analysis}.

\paragraph{Latent Layer z}

At the core of the VAE is the latent layer $z$ that samples from the vector of means, and vector of standard deviations. The two vectors of means and standard deviations are implemented using fully connected layers that are initialized with random values. Previously we discussed how in a VAE, the term $- D_{KL} (q (z|x) ) \, || \, p (z) )$ is used to approximate the true posterior. This is the KL-Loss term, and can be written as: $$\dfrac{1}{2}\sum_{j=1}^{J} = (1 + log((\sigma_{j})^{2}) - (\mu_{j})^{2} - (\sigma_{j})^{2})$$ The KL-Loss term is used as a measurement of the divergence between two probability distributions, the vector $z$ that samples from $\mathcal{N}(\mu_{i}, \sigma_{i}^{2})$, and the standard normal distribution. The sampled vector $z$  is implemented as $z = \mu + \sigma \cdot \varepsilon$ where $\varepsilon$ is an auxiliary noise variable drawn from the standard normal distribution, $\varepsilon \sim \mathcal{N}(0, 1)$. When minimizing the KL-Loss term, the vector of means $\mu$ and vector of standard deviations $\sigma$ will be optimized to resemble the target distribution. This means we can initialize the two fully connected layers with random values, and let them learn the target distribution during model training. 

The latent layer $z$ not only learns to describe the input flows $x$, but also representations of $x$ that have similar features. The mean and standard deviation vectors which $z$ sample from give a distribution to each flow. Each flow is represented as a data point, and each point have a probability distribution. Other points within the probability distribution have a higher chance of belonging to the same class, as opposed to a normal autoencoder where each point have a direct encoding, only decoding specific encodings in the latent space. Intuitively, this allows a VAE to not only learn latent representations of seen input features, but generalizes in a way that allows for an interpretation of unseen flows and flow features with slight variations.

\subsection{Second Proposed Approach:  Loss Based Detection on a Variational Autoencoder (LBD-VAE)}
\label{methodology-lbd-vae}

The second proposed approach, Loss Based Detection on a Variational Autoencoder (LBD-VAE), is a variant of the first approach. Instead of classifying the different traffic types at the latent layer, the LBD-VAE performs anomaly detection based on the reconstruction loss of the VAE after it has been trained. The LBD is based on the notion that an autoencoder is only able to reconstruct data that has been previously shown to it. Both of the proposed approaches uses the underlying framework of the VAE. This means the decisions about the VAE structure, what layers to use, when to use regularization, how the latent layer operates, etc. will remain mostly the same. The LBD-VAE could end up running on different tuning settings compared to the LLC-VAE, but the parameters available for tuning will remain mostly the same.

\begin{figure*}[htbp]
    \centering
    \includegraphics[scale=0.6]{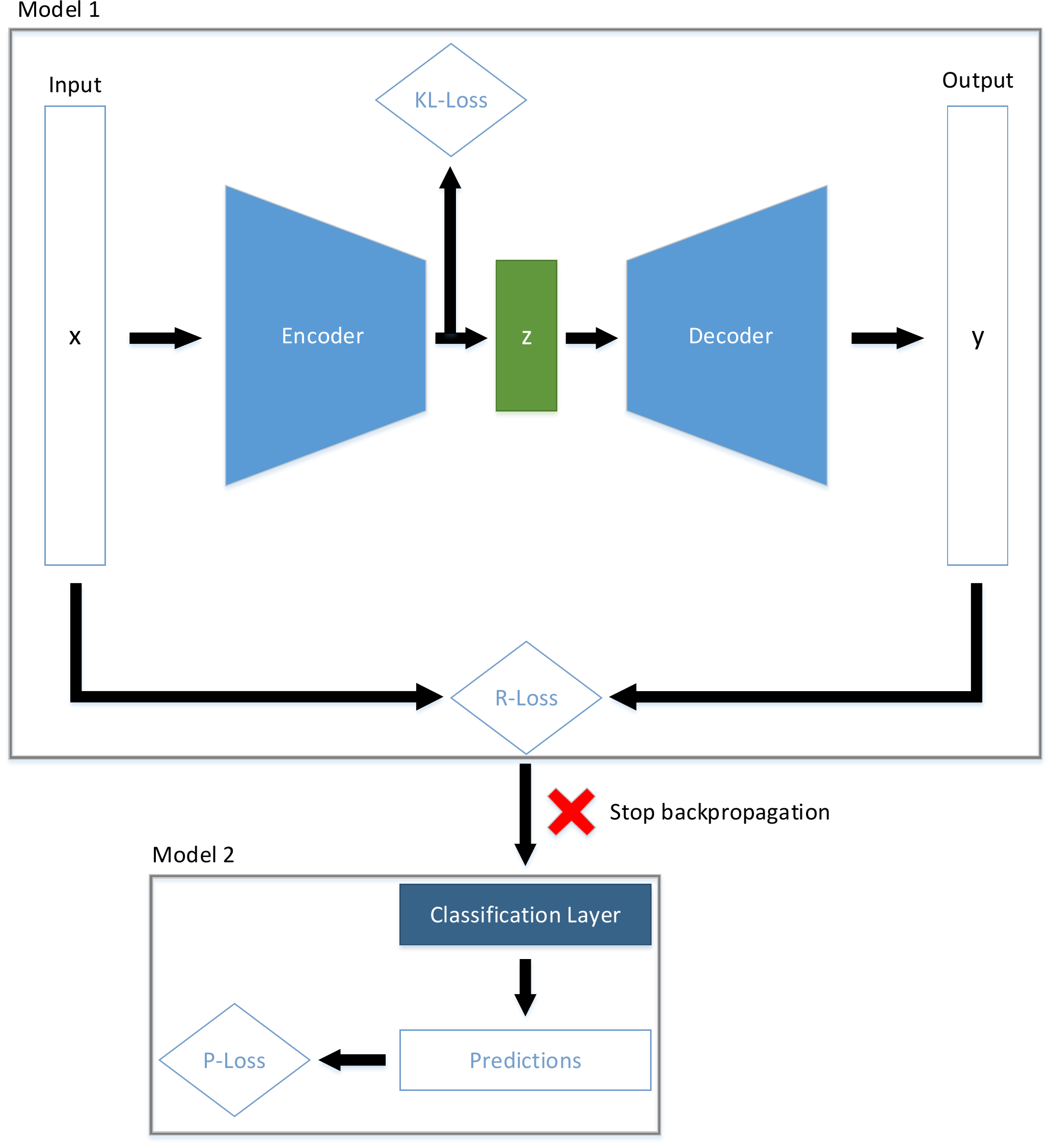}
    \caption[LBD-VAE Model]{Loss based detection on a VAE. Rectangles represent node layers. Diamond shapes represent the loss values. The trapezoids represents the shrinking nature of the encoder, and the expanding nature of the decoder. The classification layer is a fully connected layer with a sigmoid activation function. The surrounding frames depict two stages of model training. First model 1 is trained to completion, then model 2 is trained, based on the reconstruction loss of model 1.}
    \label{fig:methodology-lbd}
\end{figure*}

An overview of the LBD-VAE can be seen in figure \ref{fig:methodology-lbd}. The figure is separated into two parts, "model 1" and "model 2". These are both part of the LBD-VAE model, but will be separated for easier understanding of how the second approach works. The first part, model 1, is simply an implementation of the VAE by Kingma and Welling \cite{Kingma-and-Welling-2013}, and will be trained separately from the second part, model 2. To begin training the LBD-VAE, first we will create a dataset containing exclusively benign computer network traffic from the CSECICIDS2018 \cite{CSECICIDS2018} dataset. The idea is that by only training model 1 on benign data, the VAE will only learn the patterns of those types of flows, and as a side effect, be unable to recognize malicious data. Training only on benign data also means that the LBD-VAE have the potential of being a robust mitigation system against any type of attack.

\paragraph{Loss Based Detection}

When training the LBD-VAE, first we need to train the first part, model 1, to completion. After having trained model 1 on exclusively benign data, the idea is that the reconstruction loss will be low for benign data, and higher for malicious data. After that the second part, model 2, is trained based on the reconstruction loss from model 1. During training of model 2, we will be using a mixed dataset with both benign and malicious computer network traffic flows. The input data is fed through the whole model, starting from input $x$ as seen in figure \ref{fig:methodology-lbd}. Each flow will generate a KL-Loss and a R-Loss value, but only the R-Loss will be used to train model 2. The R-Loss will be the only feature used in training a fully connected layer that outputs the unscaled class predictions. To generate a prediction loss value (P-Loss), we perform cross-entropy on the unscaled class predictions with either a softmax or sigmoid activation function
\footnote{Both functions can be used here, and can be considered equivalent for a binary classification problem:\\$\text{Sigmoid} = \dfrac{1}{1 + e^{-x}}  = \dfrac{e^{x}}{e^{0} + e^{x}} = \text{Softmax}$}. 
When performing backpropagation based on the P-Loss to update the weights and node values of model 2, it is important to prevent backpropagation through model 1. In Tensorflow, this can be done by treating the reconstruction loss as a constant, with the use of the function \texttt{tf.stop\_gradient()}. This ensures that model 2 will be trained separately from model 1, and prevent the P-Loss from interfering with the weights and node values of model 1. Failing to prevent backpropagation through model 1 from model 2 would cause the first part to learn from malicious data, which would defeat the purpose of the LBD-VAE approach.

\section{Experimental Setup}
\label{experimental-setup}

This section will be used to present the proposed algorithms from the previous section. We will explain the specifications of the system used to run the programs, the computer network setup that the datasets came from, and what tools that were used to create the proposed algorithms. Before moving on to the analysis section, we will explain how the models can be tuned, and suggest some presets.

\subsection{System Specifications}

All the experiments were performed on a single system, using the two datasets CSECICIDS2018 \cite{CSECICIDS2018} and CICIDS2017 \cite{CICIDS2017}. The computer system runs on Windows 10 64-bit version. For deep learning, the most important hardware components are the memory, CPU and GPU. The system has 16GB of RAM, uses a Intel Core i7-5930K CPU with 6 cores at 3.50GHz (stock frequency), and a Nvidia GeForce GTX 1080 graphics card with 8GB of dedicated GPU-memory.

When running the experiments, it is possible to adjust the batch size that is fed to the deep learning models as a hyperparameter. The number of flows in one batch should usually be as high as possible to ensure optimal learning and speed up model convergence time \cite{Smith-et-al-2017, Keskar-et-al-2016}. Because the models mainly run on the GPU, the only limitation on batch size is the dedicated GPU-memory. After running a few experiments, we found a batch size of about 1000 flows per batch to be the optimal amount to ensure stable training. The deep learning models mainly run using the GPU, so the CPU is at about 15\% usage during runtime when the GPU is at max memory load. RAM usage by the model is about 6.5GB during runtime, but it should be noted that this is because we load the whole training dataset into memory to reduce training times. It is possible to rely more on the disk to reduce memory usage, but this would result in slower training times.

\subsection{Datasets}
\label{experimental-computer-networks}

For this article, the Intrusion Detection System (IDS) dataset created through a collaboration between the Communications Security Establishment (CSE) and the Canadian Institute for Cybersecurity (CIC), CSECICIDS2018 \cite{CSECICIDS2018}, will be used for training the deep learning based models. Furthermore, we will use the CICIDS2017 dataset \cite{CICIDS2017}, as a test set for the deep learning models\footnote{Both datasets, CSECICIDS2018 and CICIDS2017, are from the Canadian Institute for Cybersecurity. This is also the case for the datasets ISCX2012 and NSL-KDD.}. For an analysis of the two IDS datasets and their features, refer to the paper by  Sharafaldin et al. \cite{Sharafaldin-et-al-2018}. A similar dataset, the ISCXIDS 2012 by Shiravi et al. \cite{Shiravi-et-al-2012}, containing many of the same types of attacks have also been considered for use in this article. Due to the similarities, only one or two of these datasets were needed, and the newer 2017 and 2018 versions has been chosen for their more recent contents. For a list of the features used in both datasets, see table \ref{tab:methodology-features}. The features were extracted using CICFlowmeter\footnote{CICFlowmeter is a tool made in the Java programming language, used to generate traffic flow data from network packets.} \cite{Lashkari-et-al-2017}, turning the corresponding raw PCAP data into network flows in a CSV file format. The flows are bidirectional, where one flow represents both the forward and backward direction of a group of packets. TCP flows were terminated by a FIN packet, while UDP flows were terminated by a flow timeout. The CSV files are available for public download at \cite{CICIDS2017,CSECICIDS2018}.

\begin{table}[htbp]
    \centering
    \arrayrulecolor{black}
    \resizebox{\columnwidth}{!}{
    \begin{tabular}{llll}
        \arrayrulecolor{black}\hline
        Column 1         & Column 2       & Column 3         & Column 4                                \\
        \arrayrulecolor{black}\hline
        Src IP           & Flow IAT Min   & Pkt Len Std      & Subflow Fwd Pkts                        \\
        Dst IP           & Fwd IAT Tot    & Pkt Len Var      & Subflow Fwd Byts                        \\
        Dst Port         & Fwd IAT Mean   & FIN Flag Cnt     & Subflow Bwd Pkts                        \\
        Protocol         & Fwd IAT Std    & SYN Flag Cnt     & Subflow Bwd Byts                        \\
        Flow Duration    & Fwd IAT Max    & RST Flag Cnt     & Init Fwd Win Byts                       \\
        Tot Fwd Pkts     & Fwd IAT Min    & PSH Flag Cnt     & Init Bwd Win Byts                       \\
        Tot Bwd Pkts     & Bwd IAT Tot    & ACK Flag Cnt     & Fwd Act Data Pkts                       \\
        TotLen Fwd Pkts  & Bwd IAT Mean   & URG Flag Cnt     & Fwd Seg Size Min                        \\
        TotLen Bwd Pkts  & Bwd IAT Std    & CWE Flag Count   & Active Mean                             \\
        Fwd Pkt Len Max  & Bwd IAT Max    & ECE Flag Cnt     & Active Std                              \\
        Fwd Pkt Len Min  & Bwd IAT Min    & Down/Up Ratio    & Active Max                              \\
        Fwd Pkt Len Mean & Fwd PSH Flags  & Pkt Size Avg     & Active Min                              \\
        Fwd Pkt Len Std  & Bwd PSH Flags  & Fwd Seg Size Avg & Idle Mean                               \\
        Bwd Pkt Len Max  & Fwd URG Flags  & Bwd Seg Size Avg & Idle Std                                \\
        Bwd Pkt Len Min  & Bwd URG Flags  & Fwd Byts/b Avg   & Idle Max                                \\
        Bwd Pkt Len Mean & Fwd Header Len & Fwd Pkts/b Avg   & Idle Min                                \\
        Bwd Pkt Len Std  & Bwd Header Len & Fwd Blk Rate Avg & Label                                   \\
        Flow IAT Mean    & Pkt Len Min    & Bwd Byts/b Avg   &                                         \\
        Flow IAT Std     & Pkt Len Max    & Bwd Pkts/b Avg   &                                         \\
        Flow IAT Max     & Pkt Len Mean   & Bwd Blk Rate Avg &
    \end{tabular}
    } 
    \caption[Default Features Table]{All features used from the CICIDS2017 and CSECICIDS2018 datasets. The features are in order, starting from top to bottom, and from left to right. For an explanation of the features and what they represent, see the dataset information page \cite{CSECICIDS2018}.}
    \label{tab:methodology-features}
\end{table}

\paragraph{CICIDS2017}
\label{methodology-2017}

This dataset, CICIDS2017 \cite{CICIDS2017}, contains a variety of both malicious and benign computer network traffic records. Amongst them, the majority are flows of normal traffic, labeled \texttt{Benign}. The DoS and DDoS attacks are labeled \texttt{DoS Slowloris}, \texttt{DoS Slowhttptest}, \texttt{DoS Hulk}, \texttt{DoS Goldeneye}, and \texttt{DDoS}. The DDoS attack was simulated using a tool called Low Orbit Ion Cannon (LOIC). LOIC is a tool that can be used for HTTP/TCP/UDP flooding of a server. A single LOIC does not generate enough traffic to cause a denial of service, as such it is typically used with many computers simultaneously, composing a DDoS attack. In total there are over 2.2 million flows of benign traffic, and over 380 thousand flows of DoS and DDoS traffic. For a complete enumeration of the data, their classes, and the attack vectors, see table \ref{tab:methodology-2017}. The amount of normal traffic heavily outweighs the amount of malicious traffic. This unevenness of the distributions will be taken into account during training and testing. Deep learning algorithms needs large amounts of data to be able to learn robust, general, and accurate deep features. Whether this dataset and the CSECICIDS2018 dataset provides a sufficient amount of flows for the two proposed approaches will be discussed in \ref{analysis}, \nameref{analysis}. Although there is an unevenness to the distributions between normal and malicious traffic, there is still a considerable amount of malicious traffic that can be used, and previous research using CICIDS2017 has shown that this dataset could be used to achieve adequate results, see \ref{related-lstm-2017} from the related work section. Using this dataset for training can pose a challenge when it comes to the Slowloris, Slowhttptest, and Goldeneye attacks, as there are very few examples of those classes compared to the other types of traffic.

\begin{table}[htbp]
    \centering
    \arrayrulecolor{black}
    \begin{tabular}{lrl}
        \hline
        Traffic Type     & Number of Flows                     & Attack Vector  \\
        \arrayrulecolor{black}\hline
        Benign           & 2,273,098                           & None                                                     \\
        DoS Slowloris    & 5,796                               & HTTP/TCP-SYN                                             \\
        DoS Slowhttptest & 5,499                               & HTTP                                                     \\
        DoS Hulk         & 231,073                             & HTTP                                                     \\
        DoS Goldeneye    & 10,293                              & HTTP/TCP                                                 \\
        DDoS LOIC-HTTP   & 128,027                             & HTTP
    \end{tabular}
    \arrayrulecolor{black}
    \caption[CICIDS2017 Overview]{Overview of traffic flow data in the CICIDS2017 \cite{CICIDS2017} dataset.}
    \label{tab:methodology-2017}
\end{table}

\paragraph{CSECICIDS2018}
\label{methodology-2018}

Similar to the 2017 dataset \cite{CICIDS2017}, this version consists of a variety of malicious traffic types, in addition to benign traffic \cite{CSECICIDS2018}. The DoS and DDoS attacks are labelled \texttt{DoS Slowloris}, \texttt{DoS Slowhttptest}, \texttt{DoS Hulk}, \texttt{DoS Goldeneye}, \texttt{DDoS LOIC-HTTP}, \texttt{DDoS LOIC-UDP}, and \texttt{DDoS HOIC-HTTP}. The DDoS attacks in the 2018 dataset are similar to the attacks in the 2017 version, with the exception of a LOIC-UDP, and a High Orbit Ion Cannon (HIOC) attack class. HIOC is an attack tool used to generate a flood attack by overflowing a victim with HTTP GET and POST requests. It was created as an improvement over the previously discussed LOIC, as well as to fix some of its shortcomings. From this dataset we have more than 7.3 million flows of benign traffic available, and over 1.9 million flows of malicious traffic, with various types of DoS and DDoS attacks. Although similar to the 2017 version, most of the attack types have a considerable amount of flows each, the \texttt{Slowloris}, \texttt{Goldeneye}, and \texttt{LOIC-UDP} attacks have notably fewer flows than the other types.

The low amount of training data might negatively impact the accuracy of the proposed mitigation methods on these types of attacks, as the two proposed approaches might have too little data to properly generalize. In \ref{analysis}, \nameref{analysis}, we will discuss how much this impacts the learning process of the two approaches, if at all.

\begin{table}[htbp]
    \centering
    \arrayrulecolor{black}
    \begin{tabular}{lrl}
        \hline
        Traffic Type     & Number of Flows                     & Attack Vector  \\
        \arrayrulecolor{black}\cline{1-2}\arrayrulecolor{black}\cline{3-3}
        Benign           & 7,372,557                           & None                                                     \\
        DoS Slowloris    & 10,990                              & HTTP/TCP-SYN                                             \\
        DoS Slowhttptest & 139,890                             & HTTP                                                     \\
        DoS Hulk         & 461,912                             & HTTP                                                     \\
        DoS Goldeneye    & 41,508                              & HTTP/TCP                                                 \\
        DDoS LOIC-HTTP   & 576,191                             & HTTP                                                     \\
        DDoS LOIC-UDP    & 1,730                               & UDP                                                      \\
        DDoS HOIC-HTTP   & 686,012                             & HTTP
    \end{tabular}
    \arrayrulecolor{black}
    \caption[CSECICIDS2018 Overview]{Overview of traffic flow data in the CSECICIDS2018 \cite{CSECICIDS2018} dataset.}
    \label{tab:methodology-2018}
\end{table}

\paragraph{Dataset files}

\begin{figure*}[htbp]
    \centering
    \includegraphics[scale=0.5]{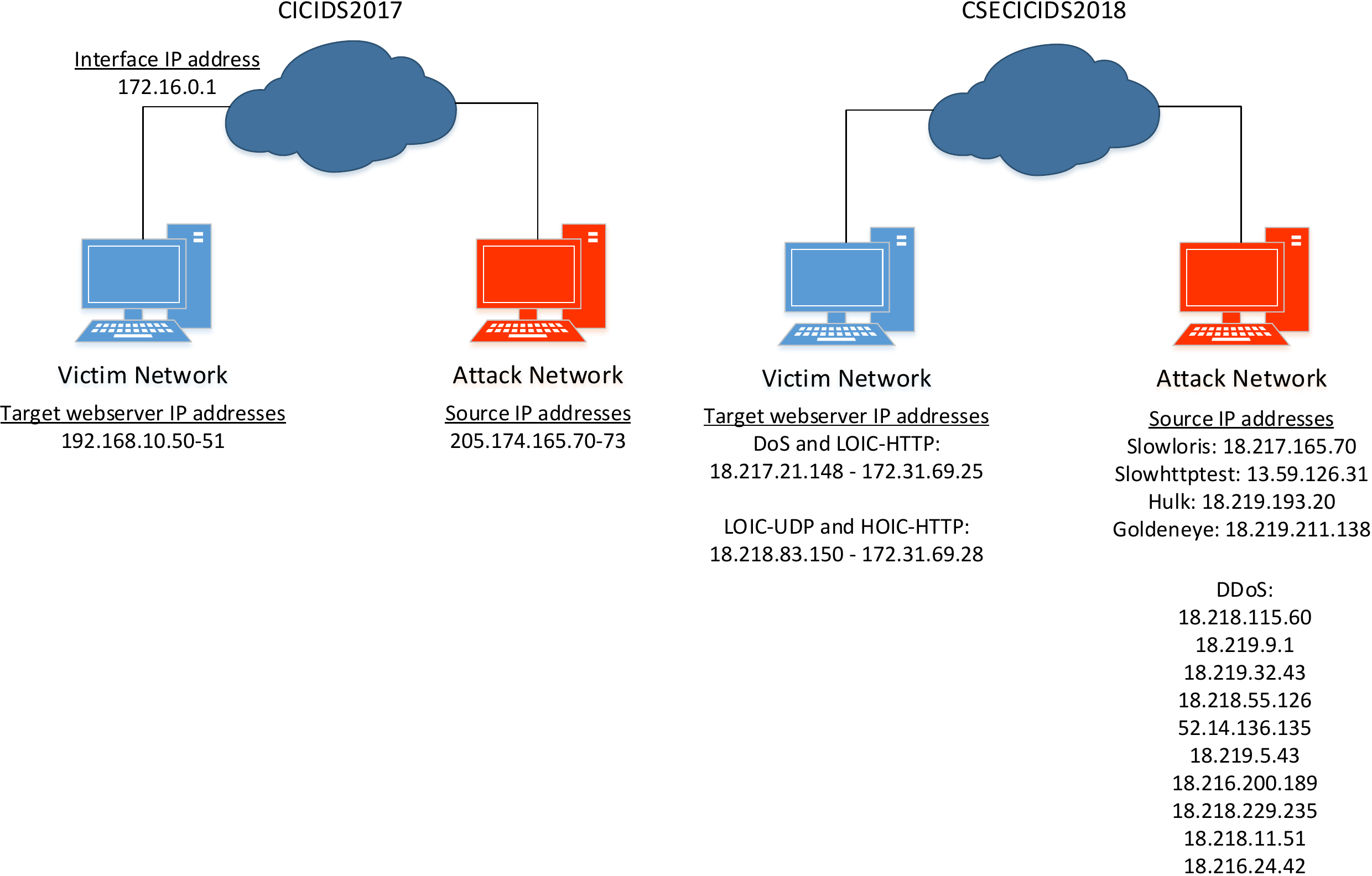}
    \caption[Simple Computer Networks of Used Datasets]{Simple example of the computer network setups used to generate the CICIDS2017 dataset (left), and the CSECICIDS2018 dataset (right). For the full network setup of the CICIDS2017 dataset, see  \cite{Sharafaldin-et-al-2018}. For the full network setup of the CSECICIDS2018 dataset, see  \cite{CSECICIDS2018}.}
    \label{fig:experimental-dataset-network}
\end{figure*}

For both datasets, we use the pre-generated CSV files publicly available for download \cite{CICIDS2017, CSECICIDS2018}. The PCAP files are available as well, but due to time constraints and the  overall good quality of the CSV files, we will not generate new datasets. The current datasets will however need transformations to streamline the feature names and the number of features, to be consistent between the two datasets used, as well as to fix weaknesses in the datasets. Among the most notable of the weaknesses in CSECICIDS2018 is the lack of source IP, destination IP, and source port on a subset of the flow traffic. To solve this problem, we looked up the attack source and destination IP for each type of attack reported in the computer network, as seen in figure \ref{fig:experimental-dataset-network}, and added them to each flow missing this data. The source ports were not feasible to recover from the original PCAP files in a reasonable time, and will thus be dropped. Another potential issue is the fact that both datasets uses simulated benign data, and not recorded real world traffic. There is no real way to mitigate this shortcoming without using other datasets. Some labels in the pre-generated CSV files might be mislabeled, which could lead to unwanted, or erroneous results \cite{Pektacs-and-Acarman-2018}. After inspection of the datasets, this problem seems to be the exception rather than the rule. As long as it is an anomaly in the data, being numerically clearly in the minority, the already noisy and random techniques of machine learning makes their impact negligible.

For the two proposed approaches presented in this article, we use the CSECICIDS2018 dataset \cite{CSECICIDS2018} for training and validation, and the CICIDS2017 dataset \cite{CICIDS2017} for testing. The terms validation set and test set, are often used interchangeably to describe the same thing, a dataset for improving a machine learning model. For the remainder of this article, when we discuss validation sets, it means flows from the same computer network as the training set. The CSECICIDS2018 dataset will be split into two subsets, one is the training set, and the other the validation set. When discussing test sets, it means flows from a different computer network than the ones featured in the training and validation sets, but that follows the same probability distribution. The CICIDS2017 dataset will be used for this. Although the test dataset, CICIDS2017, has fewer malicious flow types than the training and validation set from CSECICIDS2018, all of the flow types in CICIDS2017 are present in CSECICIDS2018, and will thus not pose a problem for using CICIDS2017 as a test set.

A simplified version of the computer networks used to simulate the flows can be seen in figure \ref{fig:experimental-dataset-network}. The CICIDS2017 network is to the left, while the CSECICIDS2018 network is to the right. In both computer networks, the benign data had multiple, different sources and destinations. The attacks always targeted the victim network, and came from a separate attack network. The IP addresses of the attack flows can be seen in figure \ref{fig:experimental-dataset-network}, below their respective networks.

\subsection{Tuning Presets}
\label{es-parameters}

An important part of any deep learning model is to tune its settings so it is able to produce a useful result. To properly capture the nuances of the two proposed approaches, and as an attempt at exploring which settings work well, we experiment with multiple, promising presets. These will be used for comparison, to show on what settings the models perform well, and on what settings they come up short. Six presets will be presented initially, with more later, each with a reference name for easy lookup. The LBD-VAE will have its own presets, detailed in \ref{analysis-lbd-results}.

The hyperparameters that will be used for tuning are:

\begin{itemize}
    \item Optimizer learning rate (LR)
    \item KL-Loss multiplier (KLM)
    \item Number of training batches iterated (Steps)
\end{itemize}

The dataset transformations used for tuning are:

\begin{itemize}
    \item Scaling technique (ST)
    \begin{itemize}
        \item No scaling (None)
        \item Min-Max normalization (N-18) or (N-18/17)\footnote{Normalization uses sampled min and max values from the used datasets. A notation of (N-18) means the values are sampled from CSECICIDS2018, while a notation of (N-18/17) means the values are sampled from both CSECICIDS2018 and CICIDS2017.}
        \item Logarithmic scaling (Log)
    \end{itemize}
\end{itemize}

Other settings that will be used for tuning:

\begin{itemize}
    \item Type of layers used for encoding and decoding (LT)
    \begin{itemize}
        \item Convolutional layers (Conv)
        \item Fully connected layers (Dense)
    \end{itemize}
    \item Regularizing layer type (RLT)
    \begin{itemize}
        \item Batch normalization (Batch)
    \end{itemize}
\end{itemize}

Unless noted otherwise, all training done with convolutional layers will use a kernel size of 5 for each layer, a stride size of 2 for the first layer, and a stride size of 1 for the other two layers. This means that the 76 input features will be reduced to 38 abstract features at the first convolutional layer of the encoder, 34 features at the second layer, and 30 features for the third layer and the latent layer.

\subsubsection{Presets}

\begin{table*}[htbp]
    \begin{subtable}[b]{0.3\textwidth}
        \caption{Preset 1}
        \label{tab:es-p1}
        \centering
        \arrayrulecolor{black}
        \begin{tabular}{l|r}
            \hline
            Parameter & Value                                             \\
            \arrayrulecolor{black}\hline
            LR        & $1 \cdot 10^{-2}$                                 \\
            KLM       & 1                                                 \\
            Steps     & 15,000                                            \\
            ST        & None                                              \\
            LT        & Conv                                              \\
            RLT       & Batch                                             \\
        \end{tabular}
    \end{subtable}
    \hfill
    \begin{subtable}[b]{0.3\textwidth}
        \caption{Preset 2}
        \label{tab:es-p2}
        \centering
        \arrayrulecolor{black}
        \begin{tabular}{l|r}
            \hline
            Parameter & Value                                             \\
            \arrayrulecolor{black}\hline
            LR        & $1 \cdot 10^{-2}$                                 \\
            \rowcolor{Gray}
            KLM       & $1 \cdot 10^{-2}$                                 \\
            Steps     & 15,000                                            \\
            \rowcolor{Gray}
            ST        & N-18                                              \\
            LT        & Conv                                              \\
            RLT       & Batch                                             \\
        \end{tabular}
    \end{subtable}
    \hfill
    \begin{subtable}[b]{0.3\textwidth}
        \caption{Preset 3}
        \label{tab:es-p3}
        \centering
        \arrayrulecolor{black}
        \begin{tabular}{l|r}
            \hline
            Parameter & Value                                             \\
            \arrayrulecolor{black}\hline
            \rowcolor{Gray}
            LR        & $1 \cdot 10^{-4}$                                \\
            KLM       & $1 \cdot 10^{-2}$                                 \\
            \rowcolor{Gray}
            Steps     & 30,000                                            \\
            \rowcolor{Gray}
            ST        & \scalebox{0.8}{N-18/17}                                              \\
            LT        & Conv                                              \\
            RLT       & Batch                                             \\
        \end{tabular}
    \end{subtable}
    \hfill
    \begin{subtable}[b]{0.3\textwidth}
        \caption{Preset 4}
        \label{tab:es-p4}
        \centering
        \arrayrulecolor{black}
        \begin{tabular}{l|r}
            \hline
            Parameter & Value                                             \\
            \arrayrulecolor{black}\hline
            LR        & $1 \cdot 10^{-4}$                                 \\
            \rowcolor{Gray}
            KLM       & $1 \cdot 10^{-4}$                                 \\
            Steps     & 30,000                                            \\
            \rowcolor{Gray}
            ST        & Log                                               \\
            LT        & Conv                                              \\
            RLT       & Batch                                             \\
        \end{tabular}
    \end{subtable}
    \hfill
    \begin{subtable}[b]{0.3\textwidth}
        \caption{Preset 5}
        \label{tab:es-p5}
        \centering
        \arrayrulecolor{black}
        \begin{tabular}{l|r}
            \hline
            Parameter & Value                                             \\
            \arrayrulecolor{black}\hline
            LR        & $1 \cdot 10^{-4}$                                 \\
            KLM       & $1 \cdot 10^{-4}$                                 \\
            Steps     & 30,000                                            \\
            ST        & Log                                               \\
            \rowcolor{Gray}
            LT        & Dense                                             \\
            RLT       & Batch                                             \\
        \end{tabular}
    \end{subtable}
    \hfill
    \begin{subtable}[b]{0.3\textwidth}
        \caption{\textbf{Preset 6}}
        \label{tab:es-p6}
        \centering
        \arrayrulecolor{black}
        \begin{tabular}{l|r}
            \hline
            Parameter & Value                                             \\
            \arrayrulecolor{black}\hline
            LR        & $1 \cdot 10^{-4}$                                 \\
            \rowcolor{Gray}
            KLM       & None                                              \\
            Steps     & 30,000                                            \\
            ST        & Log                                               \\
            \rowcolor{Gray}
            LT        & Conv                                              \\
            RLT       & Batch                                             \\
        \end{tabular}
    \end{subtable}
    \caption[Tuning Presets]{All presets parameters. The parameters that change between presets have been highlighted in gray. Preset 6 (in bold) only uses P-Loss to improve, which means that it is not a VAE, but just a simple convolutional neural network that performs dimensionality reduction.}
    \label{tab:es-presets}
\end{table*}

All the above presets were tested in \cite{baarli2019ddos}, and among them preset 4 and preset 6 were retained as the most promising ones.

\section{Analysis}
\label{analysis}

The performance of the two proposed approaches are measured by logging key features of the models, using Tensorboard plots, and confusion matrices. From Tensorboard, the graphs are plotted every 50 steps, where one step represents one batch (1,024) of flows being fed through a model. Graphs are plotted for the training dataset, the validation set, and the test set. The plotted graphs are of the prediction accuracy, P-Loss, KL-Loss, R-Loss\footnote{As discussed in \ref{methodology-llc-vae}}, and total loss mean values, as they develop over time during training, validation, and testing. The prediction accuracy mean values represents the mean accuracy of each batch. In our dataset, there is an overweight of benign traffic in the used datasets, which will influence the overall accuracy score by giving benign accuracy more weight. With this in mind, we will only use the mean accuracy score for development and performance improvement, and not as a metric to show model results. As an alternative, we can also use less benign flows, or only draw a certain percentage from each traffic type, from the datasets. To determine how well each model performs on different tuning settings, we will use confusion matrices to show the overall benign versus malicious flow traffic accuracy. As well as the accuracy within individual attack classes, separated and unaffected by the other flow accuracy scores.

\subsection{LLC-VAE Results}

The first approach that will be analyzed is the LLC-VAE. We have trained the model on the settings defined by the presets from \ref{experimental-setup}, \nameref{experimental-setup}, defined in table \ref{tab:es-presets} as reported in \cite{baarli2019ddos}. For all training runs, we will be using the CSECICIDS2018 dataset, with a 60-40 split between the training set and validation set respectively, unless noted otherwise. The training set has been modified to contain an equal amount of benign and malicious flows, to prevent learning bias. For the test set, we will be using the whole CICIDS2017 dataset.

\subsubsection{Further Adjustments}

In the following subsections, we will present further adjustments to the two most promising presets, preset 4 \ref{tab:es-p4} and preset 6 \ref{tab:es-p6}. Compared with the other presets, preset 4 and preset 6 had the overall best test results, and achieved much higher test accuracy, as reported in  \cite{baarli2019ddos}. We use the test accuracy as the deciding factor for choosing to build on these presets, because it is the best metric to show how well the LLC-VAE generalizes. The further adjustments will be used to discuss how the LLC-VAE can be improved further, and analyze whether it can be considered as a method for creating a mitigation system.

\paragraph{LIME}
\label{analysis-lime}

\begin{figure*}[htbp]
    \centering
    \includegraphics[scale=0.8]{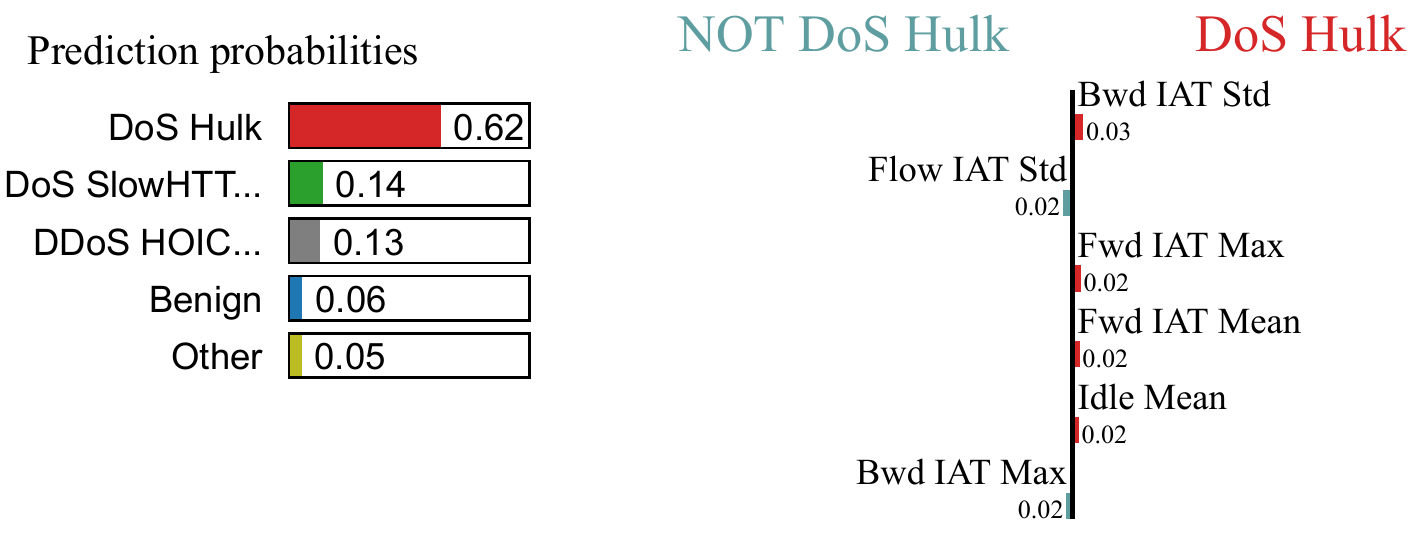}
    \caption[LIME Example For a Single Flow]{LIME example figure for a single flow. Result achieved by running LIME on the LLC-VAE using preset 4 \ref{tab:es-p4}.}
    \label{fig:lime_example}
\end{figure*}

In addition to the feature selection done by the convolutional layers, and the selection through dimensionality reduction by the encoder of the LLC-VAE model, we can manually select what input features the model should learn from. Using LIME \cite{Ribeiro-et-al-2016}, we can analyze how the presets weighed each feature, to find out which ones had the largest, and which ones had the smallest impact on the learning process. LIME shows how much impact each input feature from a flow, had on a prediction. An example of how LIME visualize the prediction probabilities for a single flow can be seen in figure \ref{fig:lime_example}. The flow from the example figure has been predicted to be a DoS Hulk attack, with a probability of 62\%. On the right hand side, we can see the top six most impactful input features, in order from most to least. Features on the side named "Not DoS Hulk", weighs against this particular flow being predicted to be a DoS Hulk attack, and vice versa for the opposite side named "DoS Hulk". The number next to each feature name represents how much they are weighted by the model, rounded to the closest, second decimal point. We can use LIME to understand how the LLC-VAE classifies flows, as well as for manual feature removal, to ensure better generalization.

To find out which input features were the most influential on the LLC-VAE training process, we ran LIME on the model after it was trained using preset 4, with the same training and validation datasets that were used previously. Running LIME is a time consuming process, so we have shortened the process by using a sample size of 50,000 flows. This produced a list of the most impactful features, in sorted order.

\begin{table}
\centering
\arrayrulecolor{black}
\resizebox{\columnwidth}{!}{
\begin{tabular}{llll}
\hline
Column 1          & Column 2         & Column 3         & Column 4                                                    \\
\arrayrulecolor{black}\hline
Bwd IAT Std       & Bwd Pkt Len Mean & Fwd Header Len   & Fwd IAT Tot                                                 \\
Pkt Len Mean      & Bwd Pkt Len Max  & Pkt Len Std      & Bwd Pkt Len Min                                             \\
Fwd IAT Max       & Active Min       & Fwd Pkt Len Max  & Bwd IAT Tot                                                 \\
Bwd IAT Min       & Idle Max         & TotLen Fwd Pkts  & Flow IAT Max                                                \\
Dst Port          & Flow Duration    & Subflow Fwd Byts & Bwd Header Len                                              \\
Init Bwd Win Byts & Bwd Pkt Len Std  & Flow IAT Min     & Subflow Fwd Pkts                                            \\
Pkt Size Avg      & Fwd IAT Std      & Fwd IAT Mean     & Active Std                                                  \\
Pkt Len Max       & Bwd IAT Mean     & Flow IAT Mean    & TotLen Bwd Pkts                                             \\
Idle Mean         & Pkt Len Var      & Fwd Seg Size Avg & Idle Min                                                    \\
Init Fwd Win Byts & Bwd IAT Max      & Fwd IAT Min      & Pkt Len Min
\end{tabular}
} 
\caption[Top Features After LIME]{Top 40 features remaining after using LIME on the LLC-VAE with preset 4 \ref{tab:es-p4}. The features are ordered from most to least impactful, from top to bottom and left to right.}
\label{tab:lime_top}
\end{table}

\paragraph{Results After Feature Selection With LIME}

To test whether manual feature selection is constructive for the LLC-VAE model performance, we tried to remove the least impactful features, only keeping the top 40, as seen in table \ref{tab:lime_top}. By removing the least impactful features, we allow the model to emphasize more on the remaining features. This could help prevent overfitting, since the model no longer tries to learn from less important data. Another method could be to select individual features, if they are deemed to be potential causes for overfitting. Doing so would be much more time consuming, and will not be done for this article.

\begin{figure}[htbp]
    \centering
    \includegraphics[scale=0.4]{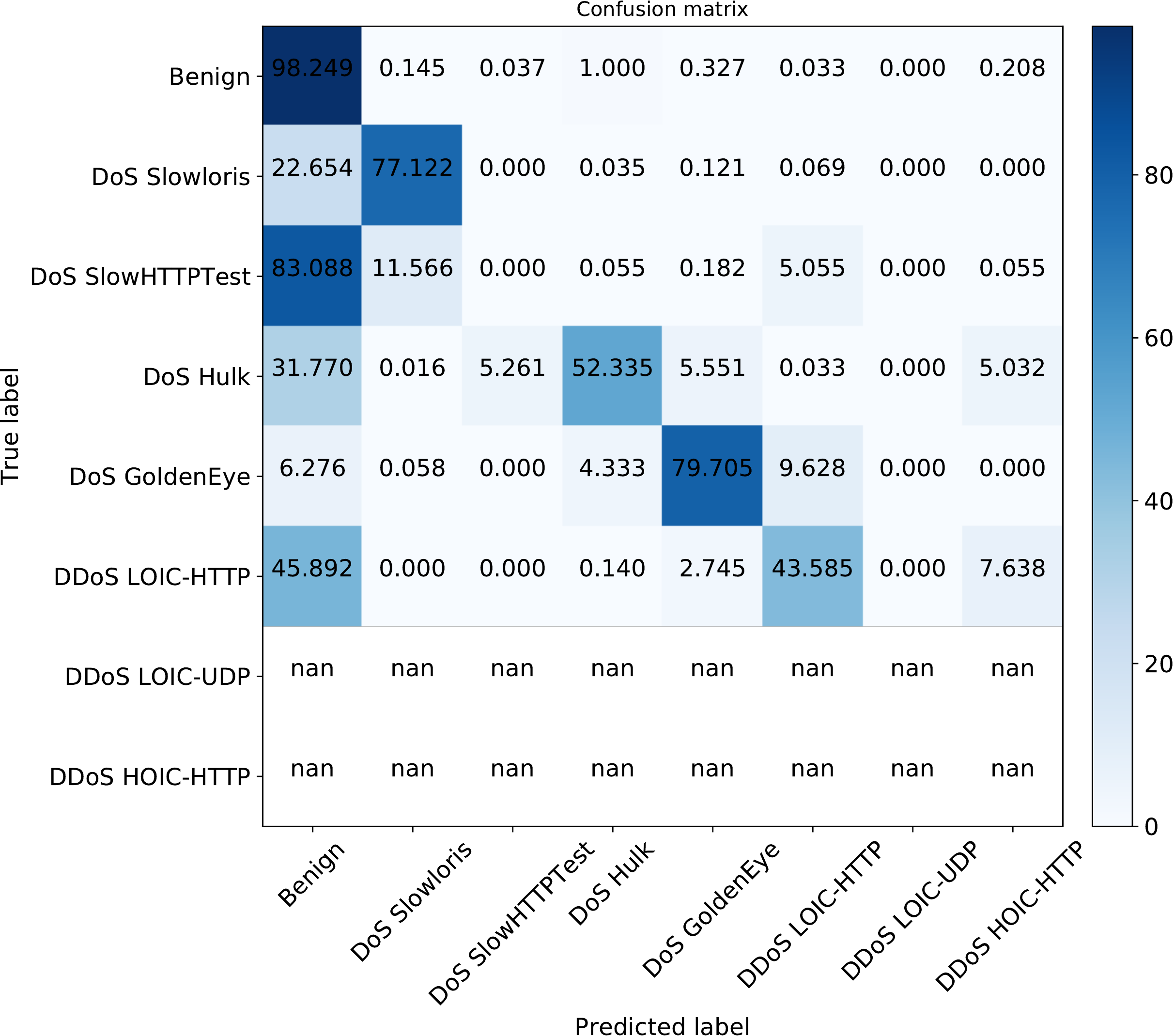}
    \caption[Preset 4 Confusion Matrix of Test Dataset]{Confusion matrix of test dataset for preset 4.}
    \label{fig:p4-te-cm}
\end{figure}

\begin{figure}[htbp]
    \centering
    \includegraphics[scale=0.4]{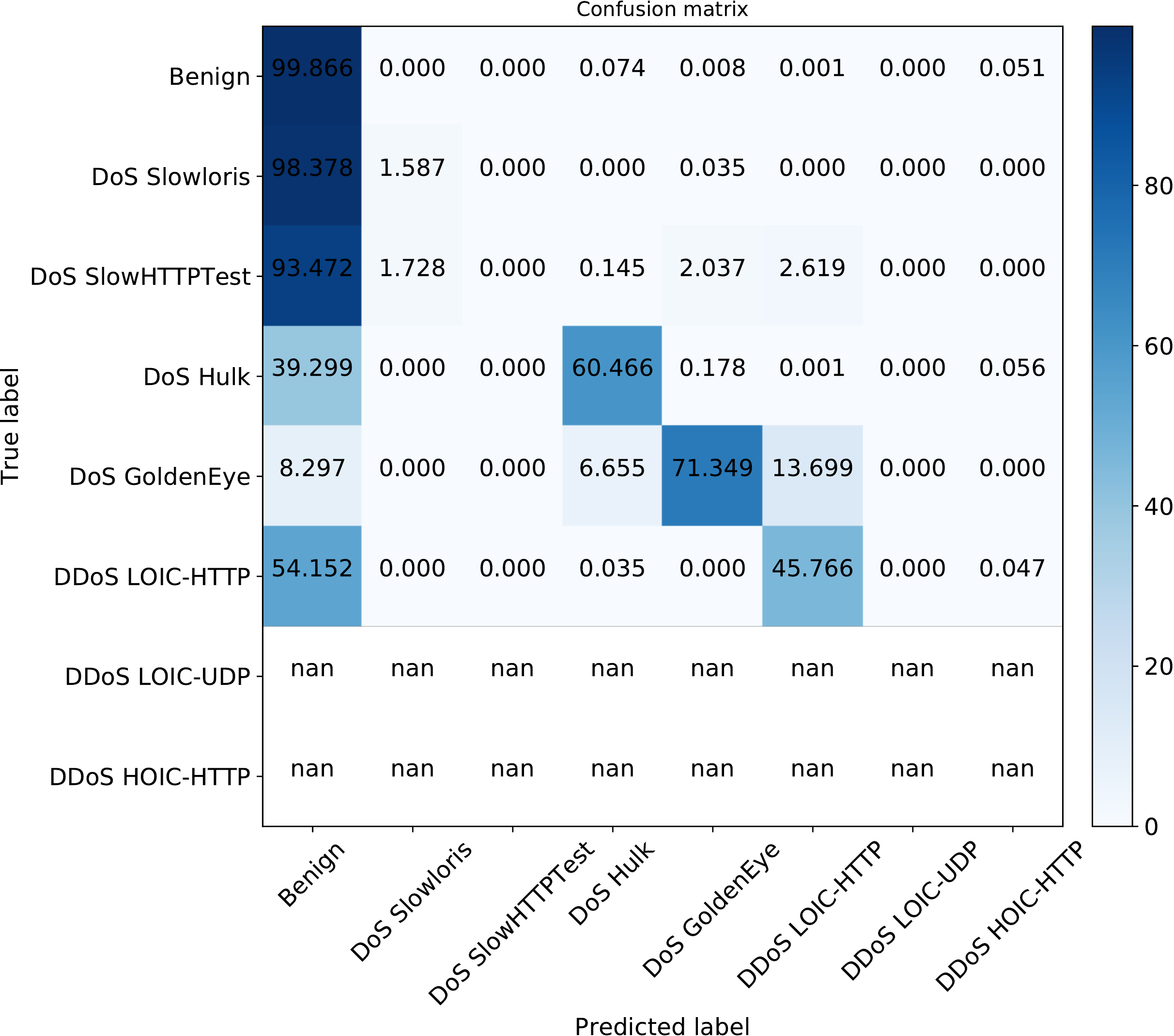}
    \caption[Preset 6 Confusion Matrix of Test Dataset]{Confusion matrix of test dataset for preset 6.}
    \label{fig:p6-te-cm}
\end{figure}

The confusion matrices from figure \ref{fig:p4-te-cm}, and figure \ref{fig:p6-te-cm}, shows the test results of of preset 4 and preset 6 respectively, before performing manual feature selection with LIME\footnote{Note that at the bottom of the confusion matrices, there are no LOIC-UDP or HOIC-HTTP attacks. This is because the test dataset contains no samples for those attack types.}. Preset 6 showed slightly better benign accuracy than preset 4, but both models had trouble classifying the malicious traffic flows. Preset 4 were particularly bad at classifying the attack \texttt{DoS SlowHTTPTest}, while preset 6 were mostly unable to classify the attacks \texttt{DoS Slowloris} and \texttt{DoS SlowHTTPTest}. Recall back to table \ref{tab:methodology-2018}, showing the number of flows for each flow type from the training dataset, CSECICIDS2018. \texttt{DoS Slowloris} had relatively few samples compared with the other attack types, which could explain why the convolutional network, using preset 6, were unable to generalize well enough to correctly classify this attack. The attack \texttt{DoS SlowHTTPTest} had relatively many flow samples in the training set, but the LLC-VAE and the convolutional network was still unable to generalize well, using preset 4 and  respectively 6. This conceptualizes the idea that the attack \texttt{DoS SlowHTTPTest} is too similar to to benign traffic flows for the LLC-VAE and the convolutional network, using preset 4 and 6, to classify it correctly.

\begin{figure}[htbp]
    \centering
    \begin{subfigure}[b]{0.4\textwidth}
        \includegraphics[scale=0.4]{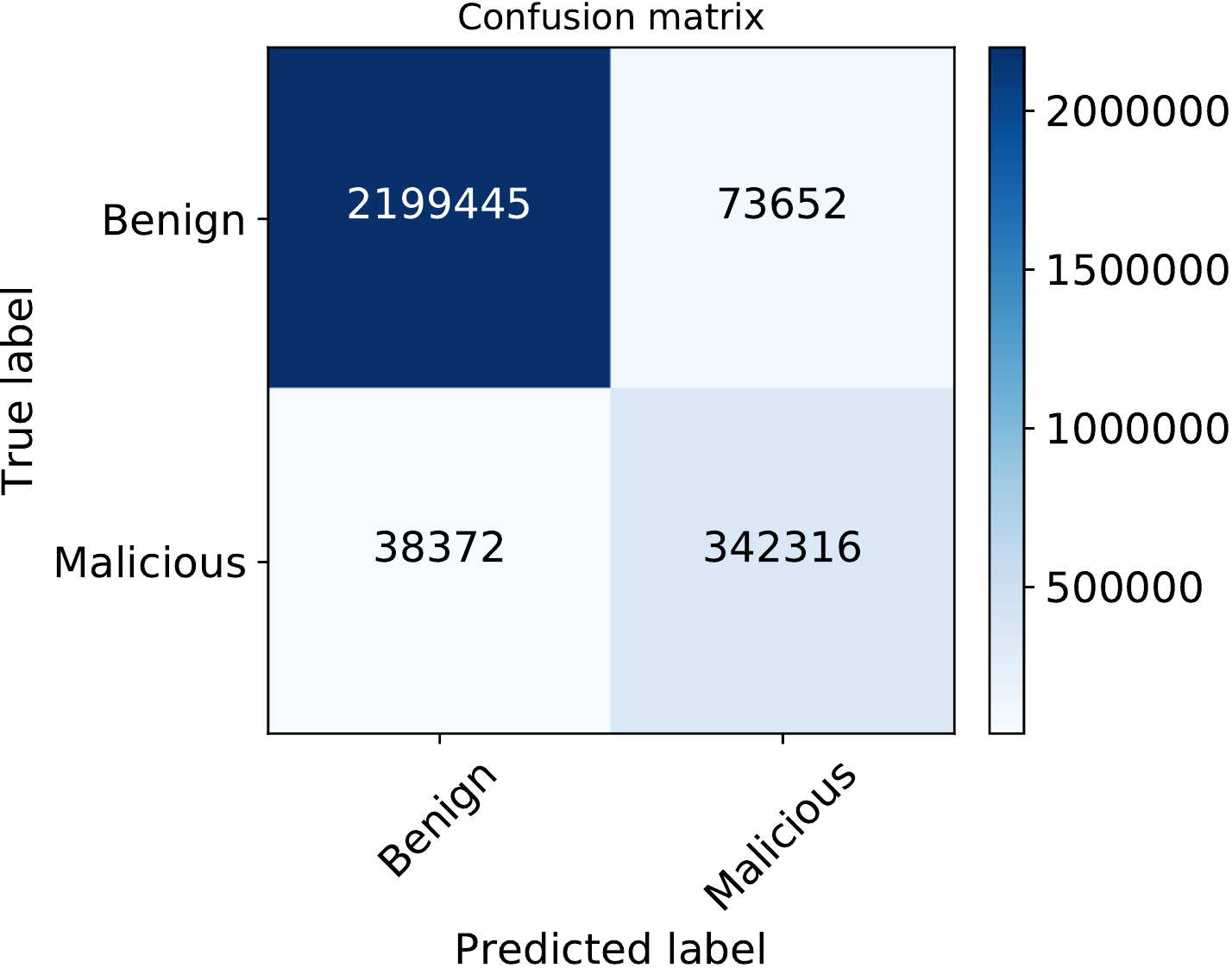}
        \caption{Amount of flows for each classification.}
    \end{subfigure}
    \hfill
    \begin{subfigure}[b]{0.4\textwidth}
        \includegraphics[scale=0.4]{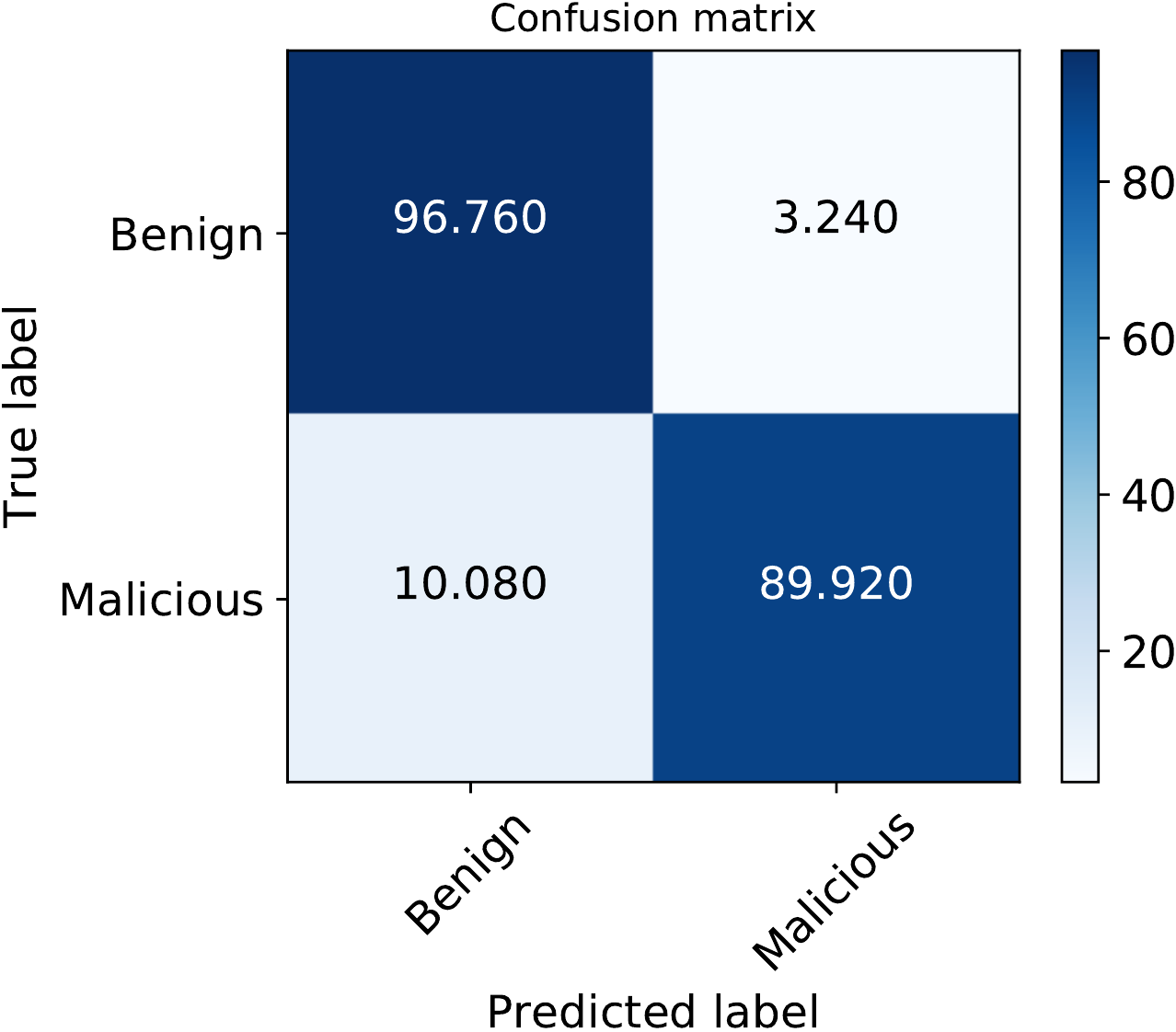}
        \caption{Accuracy values for each classification.}
    \end{subfigure}
    \caption[Confusion Matrix of Test Dataset on Preset 4a]{Confusion matrix of the test results on preset 4a.}
    \label{fig:p4-improved-LIME}
\end{figure}

\begin{figure}[htbp]
    \centering
    \begin{subfigure}[b]{0.4\textwidth}
        \includegraphics[scale=0.4]{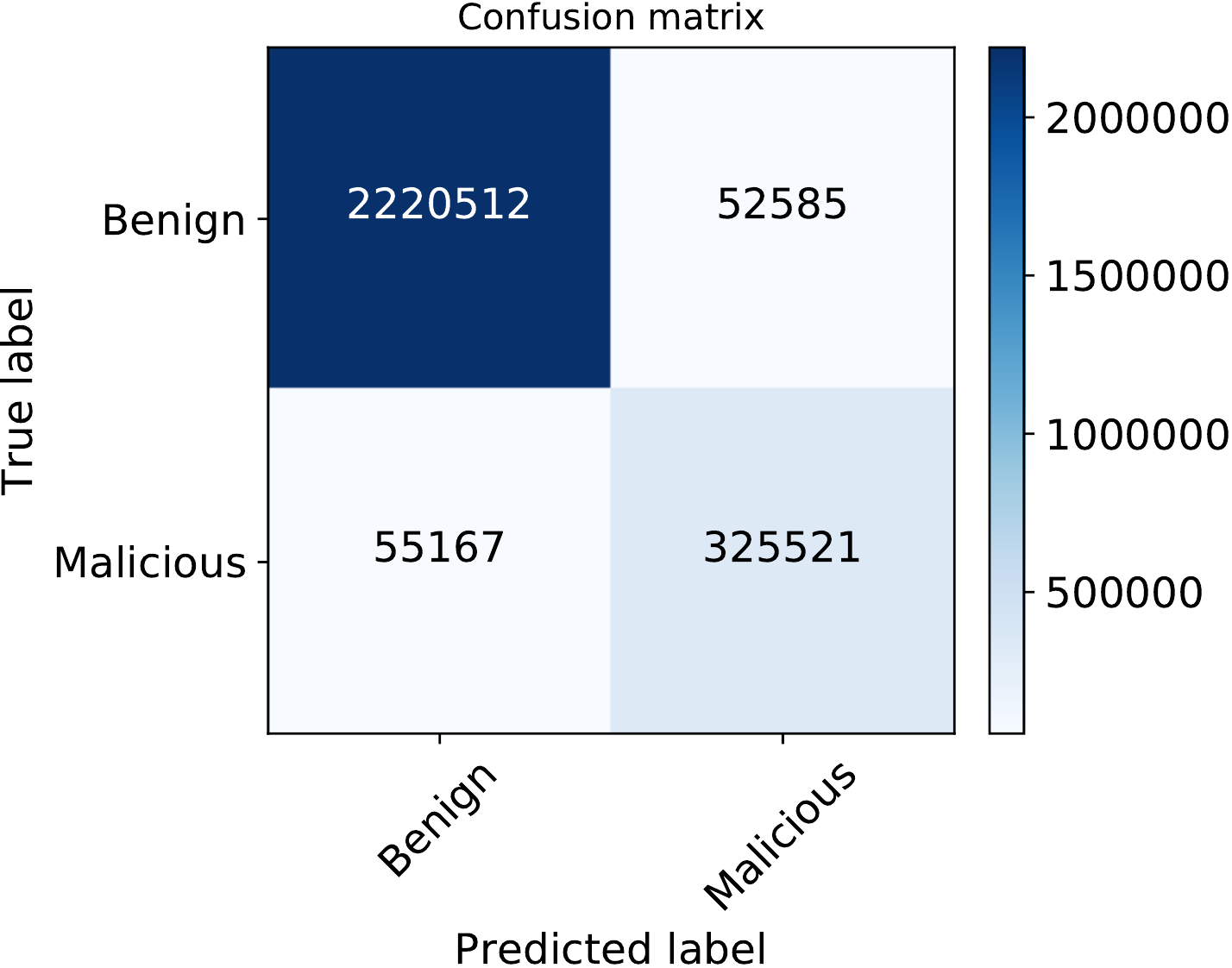}
        \caption{Amount of flows for each classification.}
    \end{subfigure}
    \hfill
    \begin{subfigure}[b]{0.4\textwidth}
        \includegraphics[scale=0.4]{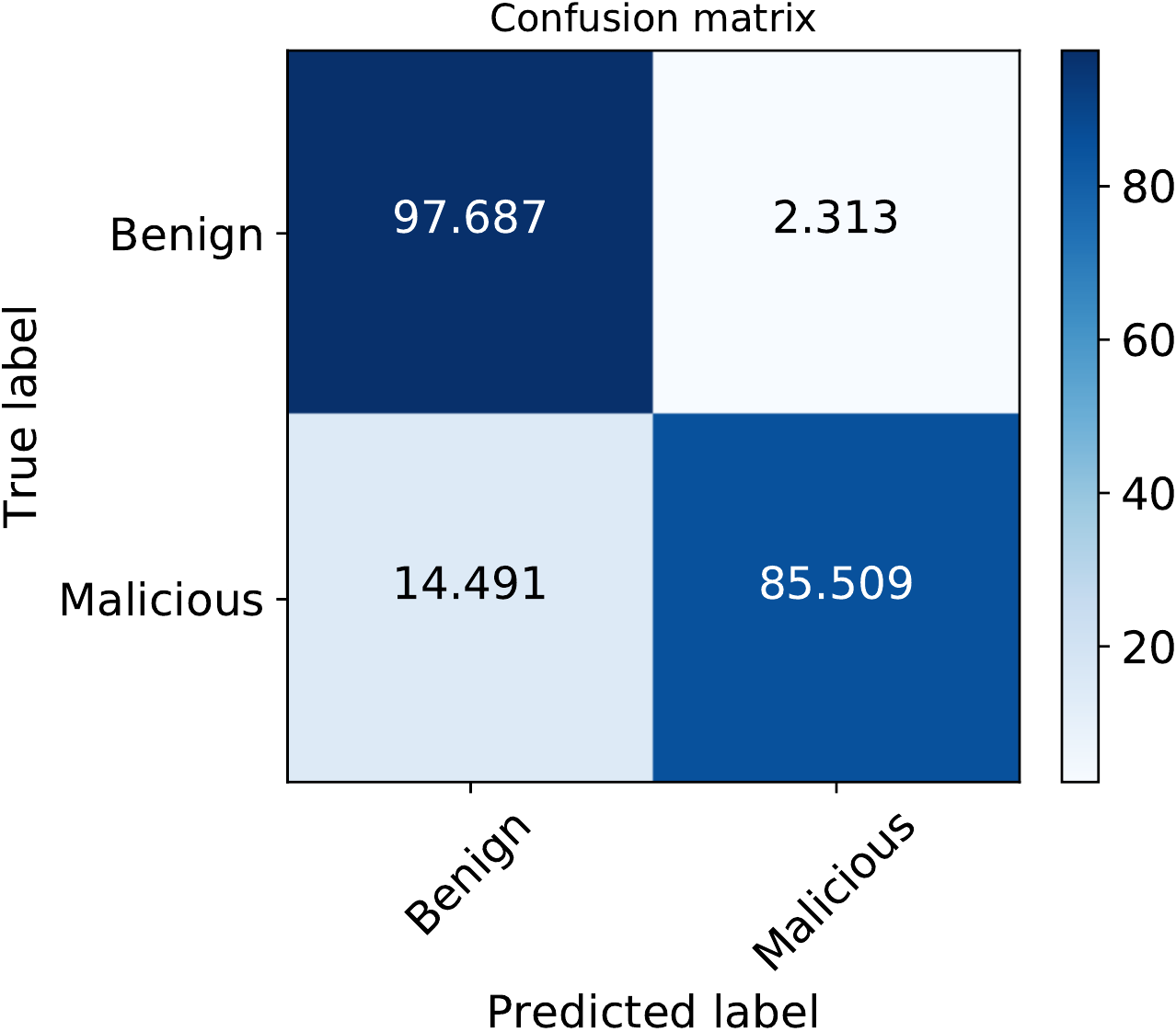}
        \caption{Accuracy values for each classification.}
    \end{subfigure}
    \caption[Confusion Matrix of Test Dataset on Preset 6a]{Confusion matrix of the test results on preset 6a.}
    \label{fig:p6-improved-LIME}
\end{figure}

After having used LIME to find the least impactful features, we kept the top 40 most impactful features, as seen in table \ref{tab:lime_top}, and ran more tests on variations of preset 4 and preset 6, as seen in table \ref{tab:presets-a}. Since differentiating between individual attack classes seemed difficult for the LLC-VAE to handle, we decided to compress them into one class named \texttt{malicious}, and instead perform binary classification. In addition to this, we have also changed the latent layer size, by changing the convolutional layers. For preset 4 and preset 6, the kernel size was 5 for each layer, and the first layer had a stride size of 2, as described in \ref{es-parameters}. The kernel size remains the same for these tests, at a size of 5, but now each of the layers have a stride size of 1. Hence the top 40 input features, \ref{tab:lime_top}, will be reduced to 28 abstract features in the latent layer, from dimensionality reduction. The confusion matrix for the test results on preset \textbf{4a} \ref{tab:p4a} can be seen in figure \ref{fig:p4-improved-LIME}, and the test results on preset \textbf{6a} \ref{tab:p6a} can be seen in figure \ref{fig:p6-improved-LIME}. From both figures, we can see that the benign accuracy has decreased by about 2\%, but the malicious accuracy has increased dramatically. As discussed previously, preset 6, and inherently 6a, is a simple, reducing convolutional network, and is used for baseline comparison with the LLC-VAE performance. The LLC-VAE performed less than 1\% worse in terms of benign accuracy, but almost 5\% better in terms of malicious accuracy. Overall the LLC-VAE performed better for this test, but could still improve with some fine tuning.

\begin{table*}[htbp]
    \begin{subtable}[b]{0.4\textwidth}
        \caption{Preset 4a}
        \label{tab:p4a}
        \centering
        \arrayrulecolor{black}
        \begin{tabular}{l|r}
            \hline
            Parameter & Value                                             \\
            \arrayrulecolor{black}\hline
            Preset    & 4                                 \\
            CT        & Binary                                                \\
            Steps     & 30,000                                            \\
            Features  & 40                                              \\
            KS        & 5                                              \\
            Stride    & 1-1-1                                            \\
        \end{tabular}
    \end{subtable}
    \hfill
    \begin{subtable}[b]{0.4\textwidth}
        \caption{Preset 6a}
        \label{tab:p6a}
        \centering
        \arrayrulecolor{black}
        \begin{tabular}{l|r}
            \hline
            Parameter & Value                                             \\
            \arrayrulecolor{black}\hline
            Preset    & 6                                 \\
            CT        & Binary                                                \\
            Steps     & 30,000                                            \\
            Features  & 40                                              \\
            KS        & 5                                              \\
            Stride    & 1-1-1                                            \\
        \end{tabular}
    \end{subtable}
    \caption[Preset 4a and Preset 6a]{Modified preset 4 \ref{tab:es-p4} and preset 6 \ref{tab:es-p5}. The parameters are as follows; \texttt{Preset} is the base preset, \texttt{CT} is short for classification type, \texttt{Features} specifies the number of input features, \texttt{KS} is short for kernel size. \texttt{Stride} specifies the stride size in the format $i_{1}-i_{2}-i_{3}$, where $i$ is the size of the first, second, and third layer respectively.}
    \label{tab:presets-a}
\end{table*}

\FloatBarrier

\begin{figure}[htbp]
    \centering
    \includegraphics[scale=0.4]{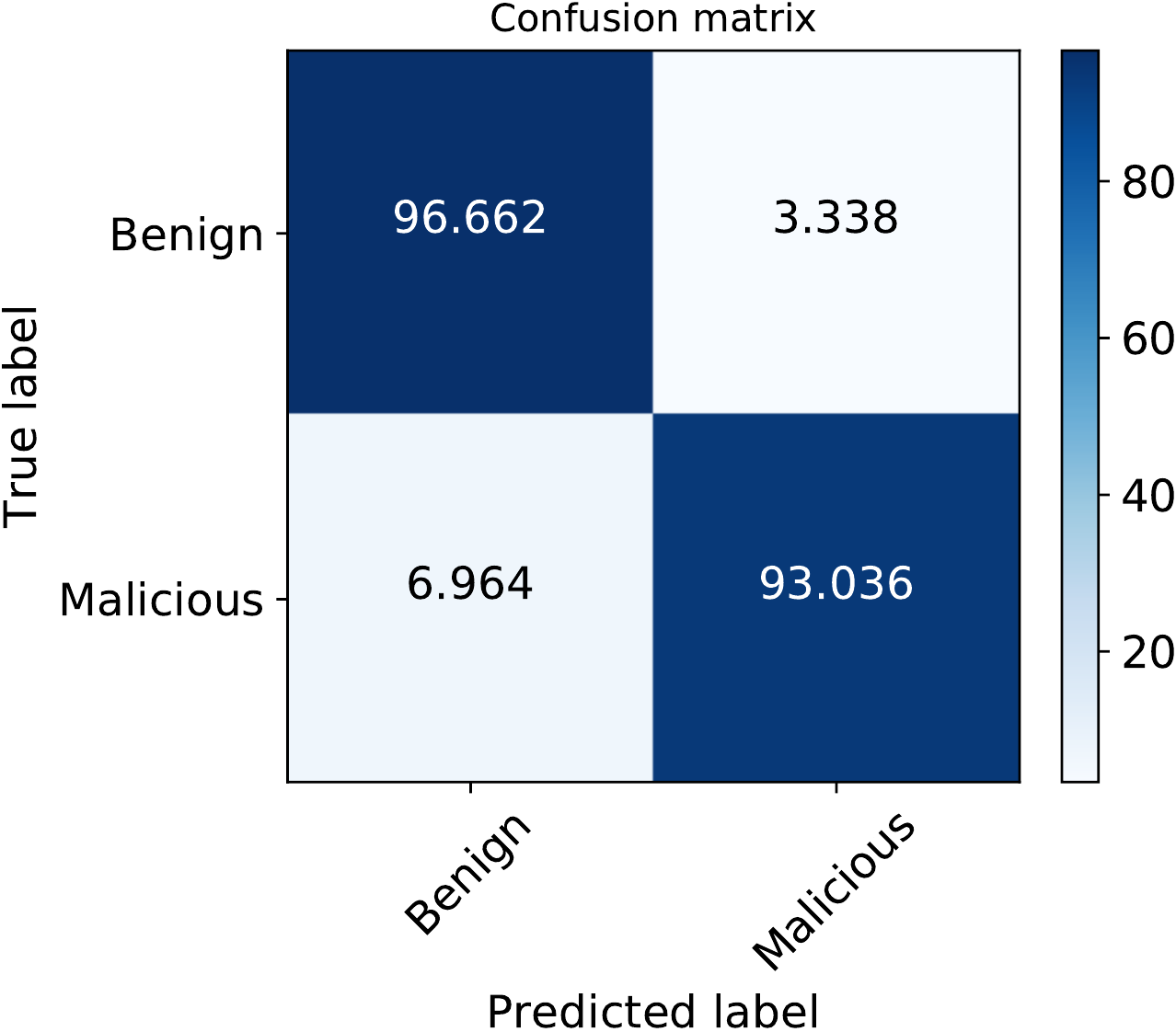}
    \caption[Confusion Matrix of Test Dataset on Preset 4a With Reduced KL-Loss Multiplier]{Confusion matrix of the test results on preset 4b, after adjusting the KL-Loss multiplier from $1 \cdot 10^{-4}$ to $1 \cdot 10^{-6}$.}
    \label{fig:p4-improved-kl}
\end{figure}

When we trained the six presets \cite{baarli2019ddos}, reducing the KL-Loss multiplier showed some improvement on overall model performance. We tried to reduce it even further with preset 4b \ref{tab:p4b}, changing the multiplier to $1 \cdot 10^{-6}$ and number of steps to 50,000. The result of doing so can be seen in figure \ref{fig:p4-improved-kl}. The benign accuracy remained mostly the same as when using a KL-Loss multiplier of $1 \cdot 10^{-4}$, but the malicious accuracy increased by more than 3\%, compared with the results of preset 4a \ref{fig:p4-improved-LIME}. Further reducing the KL-Loss multiplier caused the LLC-VAE model performance to degrade, pointing to a multiplier of $1 \cdot 10^{-6}$ to provide the best results.

\paragraph{Adjusting Receptive Field}

\begin{figure}[htbp]
    \centering
    \includegraphics[scale=0.5]{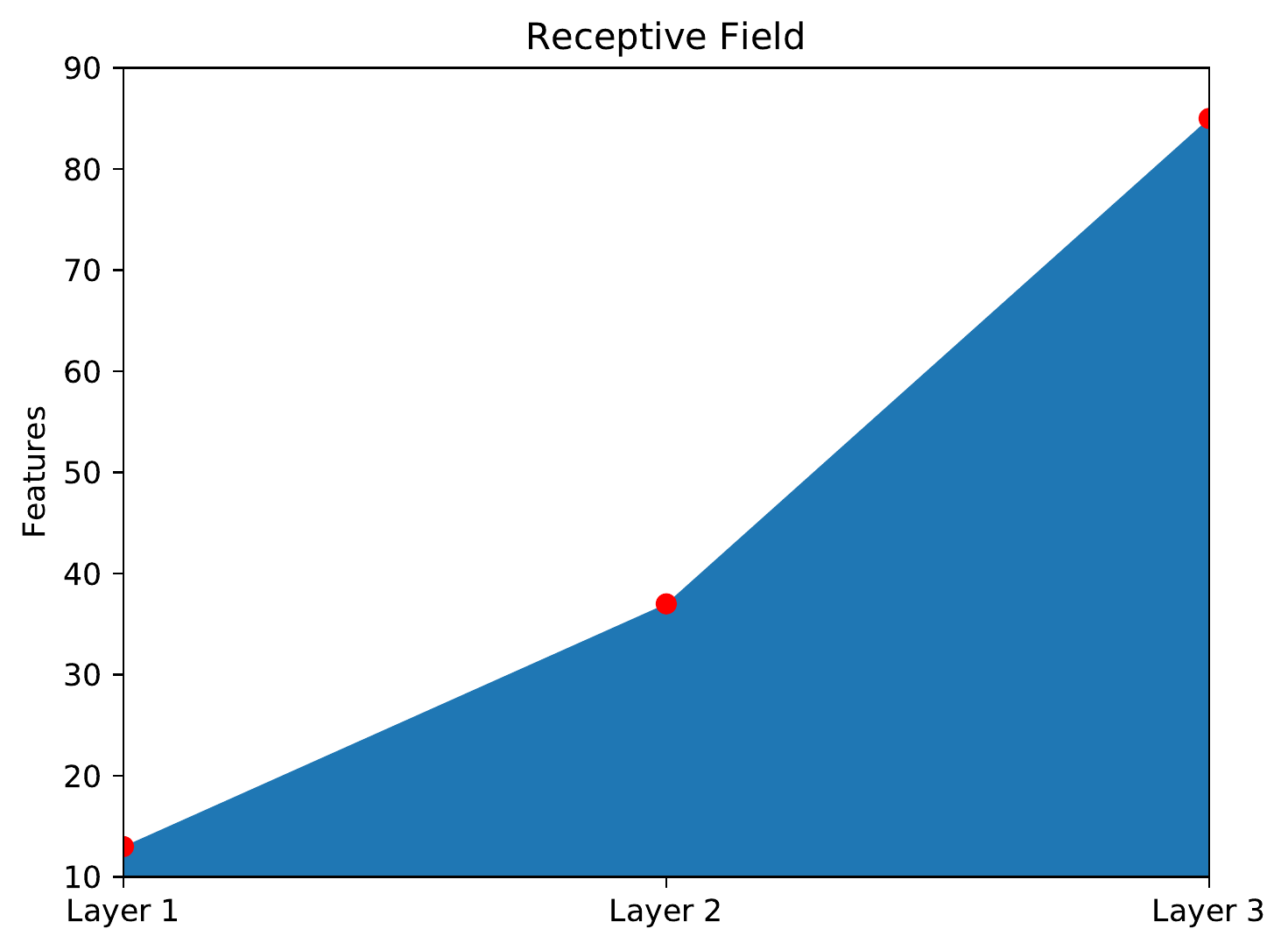}
    \caption[Receptive Field Example]{Example figure of the receptive field size for a model using three layers, a kernel size of 13, and the first two layers having a stride size of 2.}
    \label{fig:receptive-field}
\end{figure}

The tuning presets that were used to train the LLC-VAE worked as a way to try different methods for improving the model, as well as to see which settings worked, and which settings did not. Both preset 4a \ref{tab:p4a} and preset 6a \ref{tab:p6a} used convolutional layers for the encoder and decoder. Within convolutional layers, it is possible to adjust the kernel size and the stride size of the sliding window, as discussed in \ref{methodology-conv-layers}. By doing so, we can adjust the effective size of the receptive field of the latent layer, to make it learn an abstraction of the relations between a larger or smaller number of features in the input layer.

All of the original six presets \ref{tab:es-presets} uses a kernel size of 5, where the first layer has a stride size of 2, which leads the latent layer to have an effective receptive field size of 21. This means that each latent layer node is able to learn relations between 21 adjacent input features. In order for each node in the latent layer to learn from the relations between all the original 76 input features, as seen in table \ref{tab:methodology-features}, the kernel size of each layer needs to be set to 13, and the first two layers need a stride size of 2. A figure showing how the theoretical receptive field increases in this case, can be seen in \ref{fig:receptive-field}.

\begin{table*}[htbp]
    \begin{subtable}[b]{0.4\textwidth}
        \caption{Preset 4b}
        \label{tab:p4b}
        \centering
        \arrayrulecolor{black}
        \begin{tabular}{l|r}
            \hline
            Parameter & Value                                             \\
            \arrayrulecolor{black}\hline
            Preset    & 4a                                 \\
            KLM       & $1 \cdot 10^{-6}$                                                \\
            Steps     & 50,000                                            \\
            Features  & 40                                              \\
            KS        & 5                                              \\
            Stride    & 1-1-1                                            \\
        \end{tabular}
    \end{subtable}
    \hfill
    \begin{subtable}[b]{0.4\textwidth}
        \caption{Preset 4c}
        \label{tab:p4c}
        \centering
        \arrayrulecolor{black}
        \begin{tabular}{l|r}
            \hline
            Parameter & Value                                             \\
            \arrayrulecolor{black}\hline
            Preset    & 4a                                 \\
            KLM       & $1 \cdot 10^{-6}$                                                 \\
            Steps     & 140,000                                            \\
            Features  & 40                                              \\
            KS        & 7                                              \\
            Stride    & 2-2-1                                            \\
        \end{tabular}
    \end{subtable}
    \caption[Preset 4b and Preset 4c]{Two modifications on preset 4a \ref{tab:p4a}. As from earlier, \texttt{KLM} is short for KL-Loss multiplier.}
    \label{tab:presets-bc}
\end{table*}

\begin{figure*}[htbp]
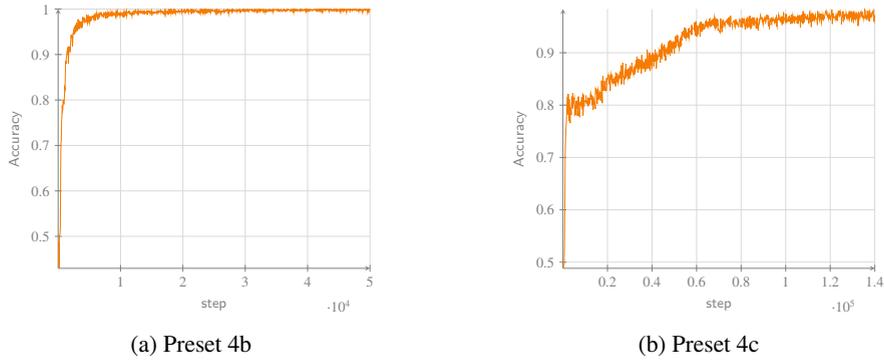

    \centering
    \begin{subfigure}[b]{0.4\textwidth}
        \centering
        \tfFromcsv{img/tensorboard/p4_modified_26_va_acc.csv}{Accuracy}{scale=0.6}
        \caption{Preset 4b}
        \label{fig:p4-latent-left}
    \end{subfigure}
    \begin{subfigure}[b]{0.4\textwidth}
        \centering
        \tfFromcsv{img/tensorboard/p4_modified_35_va_acc.csv}{Accuracy}{scale=0.6}
        \caption{Preset 4c}
        \label{fig:p4-latent-right}
    \end{subfigure}
    \caption[Validation Accuracy When Changing Receptive Field Size]{Comparison of the validation accuracy when using two different receptive field sizes with the LLC-VAE on preset 4b and 4c.}
    \label{fig:p4-modified-latent-26-35}
\end{figure*}

The best results has thus far been achieved after training the model on preset \textbf{4b} \ref{tab:p4b} with a KL-Loss multiplier of $1 \cdot 10^{-6}$, as seen in figure \ref{fig:p4-improved-kl}, with the reduced input feature set from table \ref{tab:lime_top}. To achieve an effective receptive field size that covers all the 40 features from the reduced input feature set, we adjusted the kernel size of the convolutional layers to 7, and the first and second layer to have a stride size of 2, as seen in preset \textbf{4c} \ref{tab:p4c}. This gives the latent layer nodes a theoretical receptive field size that covers 43 features. From figure \ref{fig:p4-latent-left}, we see the accuracy plot of the LLC-VAE performance using preset 4b on the validation set. From figure \ref{fig:p4-latent-right}, we see the accuracy plot of the LLC-VAE performance using preset 4c on the same set. We can see from figure \ref{fig:p4-modified-latent-26-35} that increasing the receptive field size did not improve the validation accuracy, but rather degraded it. The model converged to a solution at about step 130,000.

\FloatBarrier

\paragraph{Other Changes}

To better understand what works well and what does not, for the LLC-VAE, we have tried further fine tuning of the model based on preset 4, as well as the improvements done with LIME and the modifications on the convolutional layers. We have tried to adjust the KL-Loss multiplier, and ended up with a multiplier of $1 \cdot 10^{-6}$, as seen in preset 4b \ref{tab:p4b}, to show the best performance. Adjusting the P-Loss or R-Loss multipliers showed no improvements. Further adjustments of the convolutional layers showed that the optimal settings for the kernel sizes were 5 with a stride size of 1 for all layers, leading to a latent layer size of 28. Both larger and smaller latent layer sizes showed worse end results, even though larger latent layer sizes caused the model to converge faster.

Earlier, when discussing the datasets, we could see that some of the malicious traffic flows from the training set CSECICIDS2018 had few samples, see \ref{methodology-2018}. The attacks \texttt{DoS Slowloris}, \texttt{DoS Goldeneye}, and \texttt{DDoS LOIC-UDP} had relatively few samples compared to the other attack classes. To test whether this negatively impacted the training process, we removed them for a training run. Contrary to our initial beliefs, removing these attacks from the training set decreased model accuracy. The training and validation accuracy remained the same as before, but the test accuracy degraded. This could point to the model generalizing better when subjected to a variety of different attack types with few samples.

\begin{figure*}[htbp]
    \centering
    \begin{subfigure}[b]{0.4\textwidth}
        \includegraphics[scale=0.4]{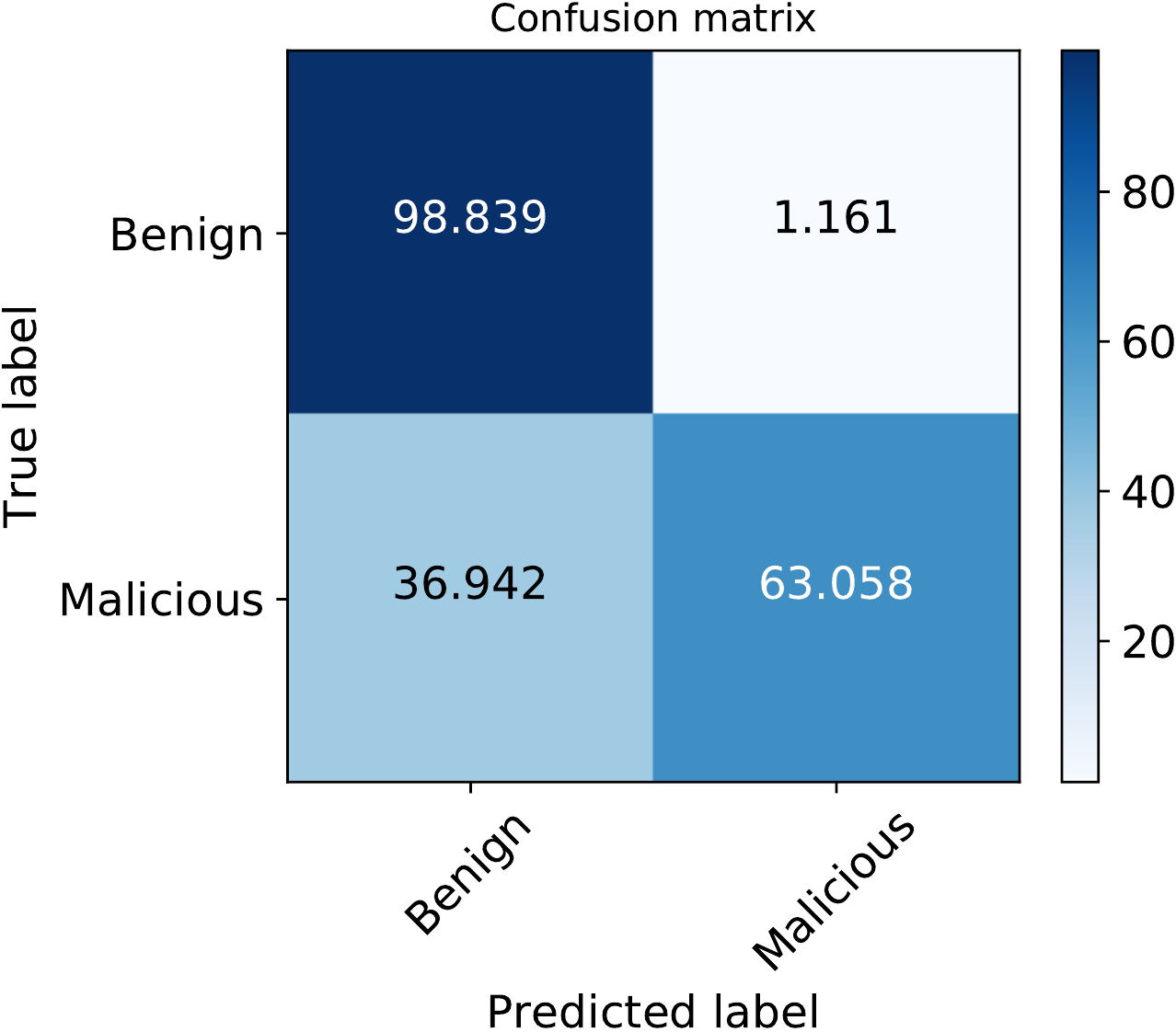}
        \caption{With IP addresses.}
        \label{fig:p4-ip}
    \end{subfigure}
    \hfill
    \begin{subfigure}[b]{0.4\textwidth}
        \includegraphics[scale=0.4]{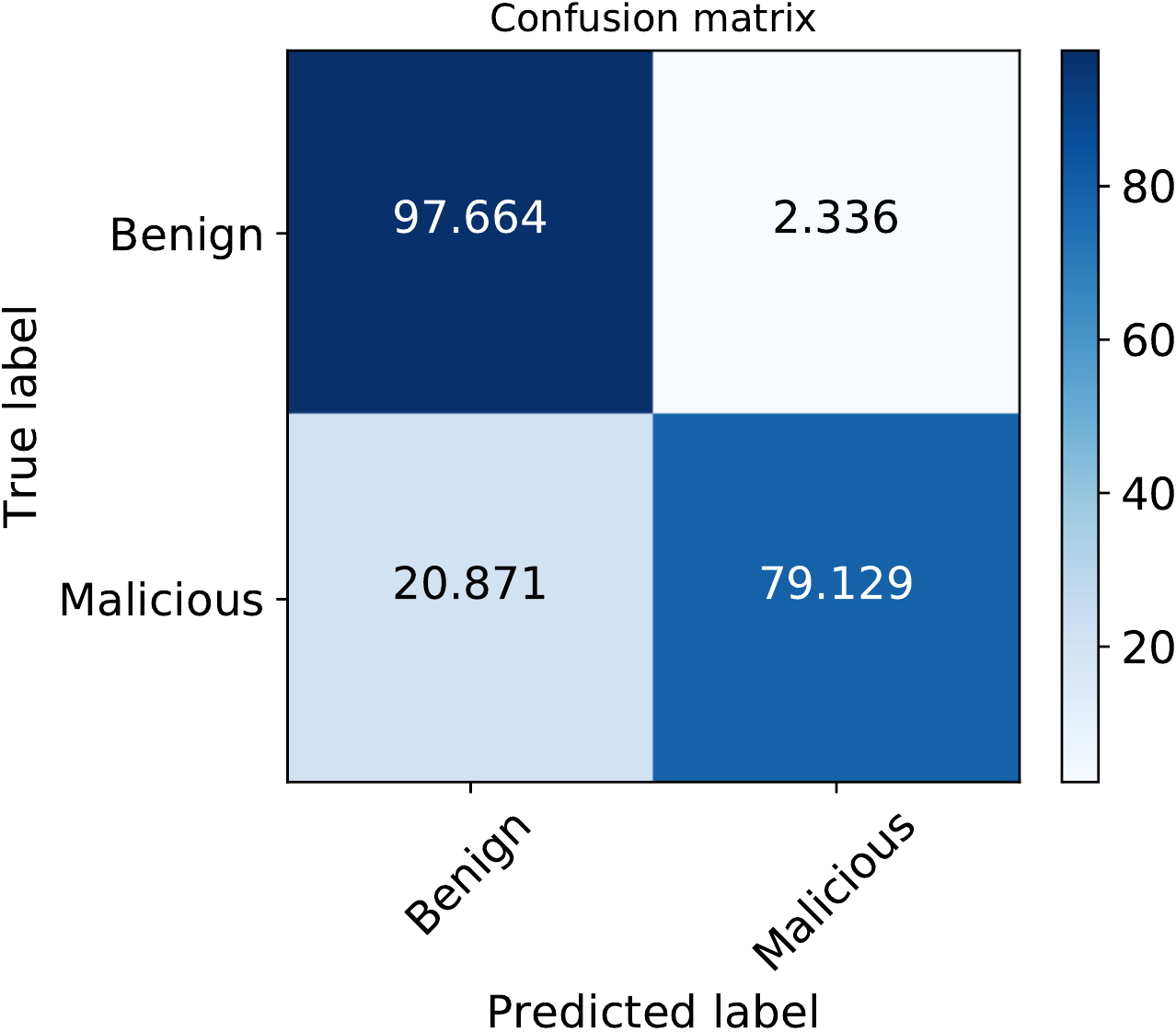}
        \caption{Without IP addresses.}
        \label{fig:p4-noip}
    \end{subfigure}
    \caption[Preset 4 Test Results Comparison, With and Without IP Addresses]{Comparison of test results of preset 4, with and without IP addresses.}
    \label{fig:p4-ip-and-noip}
\end{figure*}

Amongst the features that were removed after using LIME, the source IP and destination IP addresses were also removed. It is common for mitigation systems, when determining whether a network packet or flow is an attack or not, to analyze the source and destination IP addresses. By doing so, a mitigation system can determine the intent of the source by connecting their IP address to their behavior and traffic frequency. The LLC-VAE however, performed better after removing the IP addresses, improving overall model performance for the test dataset. We performed a training run on the settings used with preset 4 \ref{tab:es-p4} where we removed the IP address input features, and got the test results seen in figure \ref{fig:p4-noip}. Figure \ref{fig:p4-ip} shows the confusion matrix for preset 4 when performing binary classification, which used the IP addresses during training. The benign detection accuracy decreased by about 1\% when removing the IP addresses. The malicious detection accuracy increased by about 16\%, which is a significant leap in overall model performance. This points to the LLC-VAE overfitting on the IP address input features, learning specific IP addresses instead of the relationships between the IP addresses and the other input features. 

\subsection{LBD-VAE Results}
\label{analysis-lbd-results}

The second approach, LBD-VAE, will use similar tuning settings to the first approach. The presets will not be used directly, but variations will be tested, based on the findings of what worked well for the LLC-VAE, these new settings can be seen in table \ref{tab:lbd-settings}. 
\begin{table}[htbp]
    \begin{subtable}[b]{0.4\textwidth}
        \caption{LBD-preset 1}
        \label{tab:lbd-s1}
        \centering
        \arrayrulecolor{black}
        \begin{tabular}{l|r}
            \hline
            Parameter & Value                                             \\
            \hline
            LR        & $1 \cdot 10^{-4}$                                 \\
            KLM       & 1                                                 \\
            Steps 1   & 20,000                                            \\
            Steps 2   & 30,000                                            \\
            ST        & Log                                               \\
            LT        & Conv                                              \\
            RLT       & Batch                                             \\
            Features  & 76                                                \\
            KS        & 5                                                 \\
            Stride    & 2-1-1                                             \\
        \end{tabular}
    \end{subtable}
    \hfill
    \begin{subtable}[b]{0.4\textwidth}
        \caption{LBD-preset 2}
        \label{tab:lbd-s2}
        \centering
        \arrayrulecolor{black}
        \begin{tabular}{l|r}
            \hline
            Parameter & Value                                             \\
            \hline
            LR        & $1 \cdot 10^{-4}$                                 \\
            \rowcolor{Gray}
            KLM       & 4                                                 \\
            Steps 1   & 20,000                                            \\
            Steps 2   & 30,000                                            \\
            ST        & Log                                               \\
            LT        & Conv                                              \\
            RLT       & Batch                                             \\
            \rowcolor{Gray}
            Features  & 40                                                \\
            KS        & 5                                                 \\
            \rowcolor{Gray}
            Stride    & 1-1-1                                             \\
        \end{tabular}
    \end{subtable}
    \hfill
    \begin{subtable}[b]{0.4\textwidth}
        \caption{LBD-preset 3}
        \label{tab:lbd-s3}
        \centering
        \arrayrulecolor{black}
        \begin{tabular}{l|r}
            \hline
            Parameter & Value                                             \\
            \hline
            LR        & $1 \cdot 10^{-4}$                                 \\
            \rowcolor{Gray}
            KLM       & $1 \cdot 10^{-6}$                                 \\
            \rowcolor{Gray}
            Steps 1   & 1,500                                             \\
            Steps 2   & 30,000                                            \\
            ST        & Log                                               \\
            LT        & Conv                                              \\
            RLT       & Batch                                             \\
            Features  & 40                                                \\
            KS        & 5                                                 \\
            Stride    & 1-1-1                                             \\
        \end{tabular}
    \end{subtable}
    \hfill
    \begin{subtable}[b]{0.4\textwidth}
        \caption{LBD-preset 4}
        \label{tab:lbd-s4}
        \centering
        \arrayrulecolor{black}
        \begin{tabular}{l|r}
            \hline
            Parameter & Value                                             \\
            \hline
            LR        & $1 \cdot 10^{-4}$                                 \\
            KLM       & $1 \cdot 10^{-6}$                                 \\
            \rowcolor{Gray}
            Steps 1   & 100,000                                           \\
            Steps 2   & 30,000                                            \\
            ST        & Log                                               \\
            LT        & Conv                                              \\
            RLT       & Batch                                             \\
            Features  & 40                                                \\
            \rowcolor{Gray}
            KS        & 7                                                 \\
            \rowcolor{Gray}
            Stride    & 2-2-1                                             \\
        \end{tabular}
    \end{subtable}
    \caption[Settings for the LBD-VAE Model]{Settings for the LBD-VAE Model. Steps 1 is the number of iterations for "model 1", steps 2 is the number of iterations for "model 2".}
    \label{tab:lbd-settings}
\end{table}
\FloatBarrier

The LBD-VAE consists of two different models that need to be trained in order, where the first model is trained in isolation from the second one, as seen in figure \ref{fig:methodology-lbd}. As was done previously, all training runs will use the CSECICIDS2018 dataset with a 60-40 split between the training set and the validation set respectively. The training set has been modified to contain an equal amount of benign and malicious flows, to prevent learning bias.

\subsection{Comparison of Results}
\label{analysis-comparison}

\begin{figure*}[htbp]
    \centering
    \begin{subfigure}[b]{0.4\textwidth}
        \includegraphics[scale=0.4]{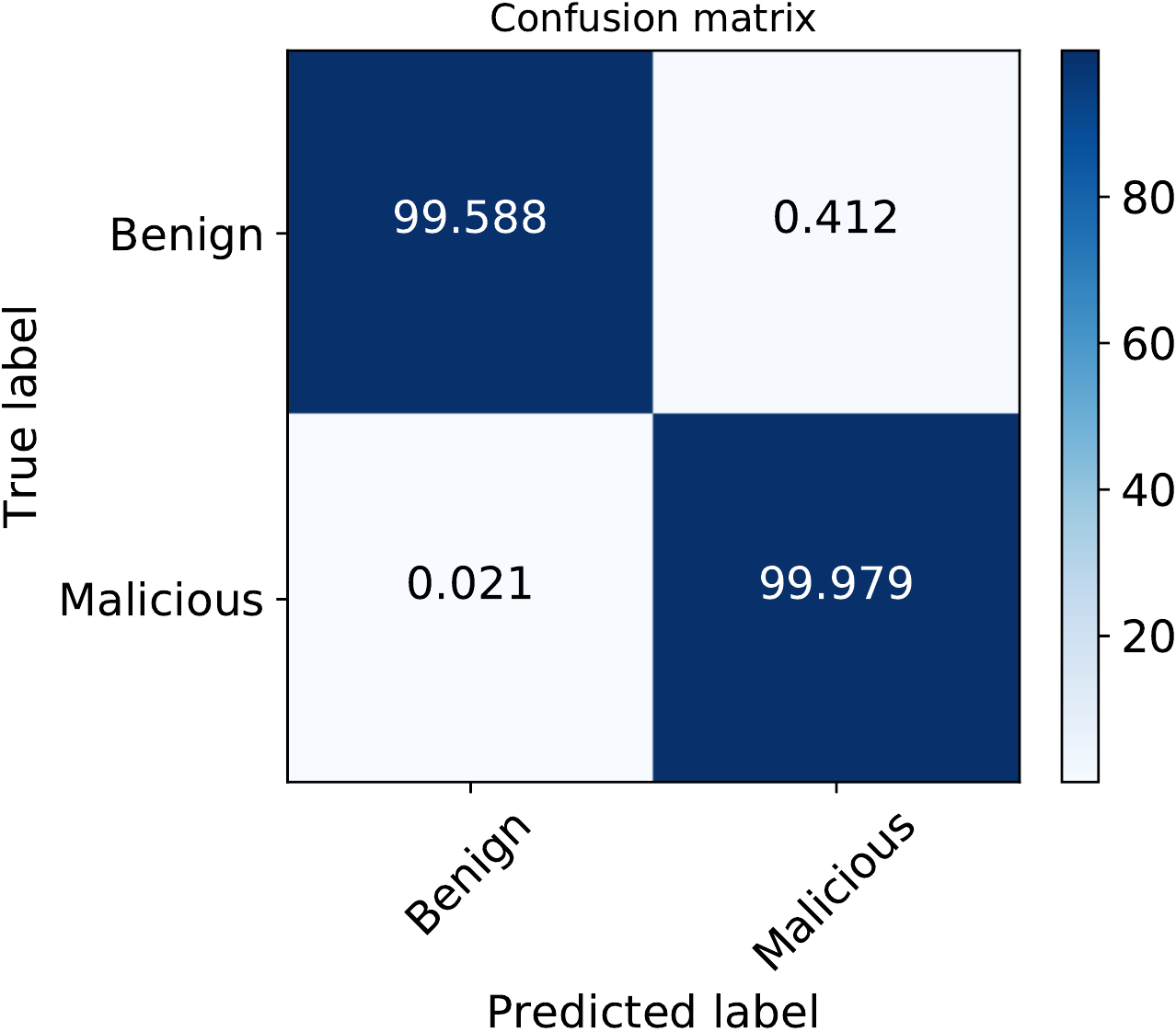}
        \caption{LLC-VAE validation accuracy.}
        \label{fig:llc-validation}
    \end{subfigure}
    \hfill
    \begin{subfigure}[b]{0.4\textwidth}
        \includegraphics[scale=0.4]{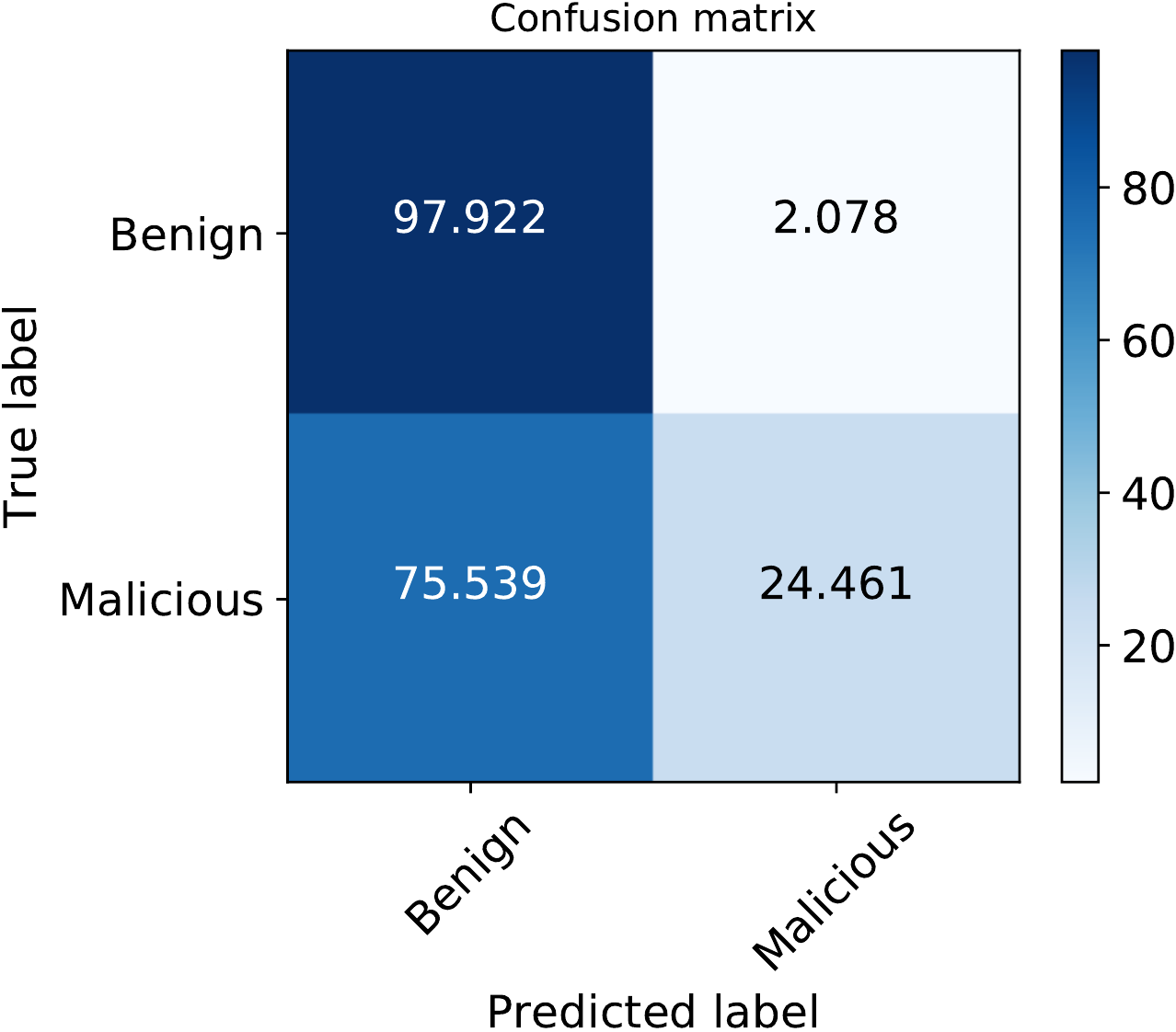}
        \caption{LBD-VAE validation accuracy.}
        \label{fig:lbd-validation}
    \end{subfigure}
    \caption[LLC-VAE and LBD-VAE Validation Accuracy Comparison]{Validation accuracy comparison.}
    \label{fig:analysis-llc-lbd-validation}
\end{figure*}

At the currently best settings, using preset 4b \ref{tab:p4b}, the LLC-VAE achieved a benign accuracy of 99.59\% and a malicious accuracy of 99.98\% on the validation set\footnote{We use validation set here for comparison between the LLC-VAE and the LBD-VAE, since we did not do any test runs using the LBD-VAE because of poor training performance.}, as seen in figure \ref{fig:llc-validation}. This means that based on the validation set, 4 out of every 1,000 legitimate flows, and the corresponding source IP addresses, would be blocked. At the same time, only 2 out of every 10,000 malicious flows would be let through the LLC-VAE. The LBD-VAE prioritized benign accuracy, which is better than prioritizing malicious accuracy.  However, the overall accuracy, as seen in \ref{fig:lbd-validation}, was much lower compared to the accuracy of the LLC-VAE.

\begin{figure*}[htbp]
    \centering
    \begin{subfigure}[b]{0.4\textwidth}
        \includegraphics[scale=0.4]{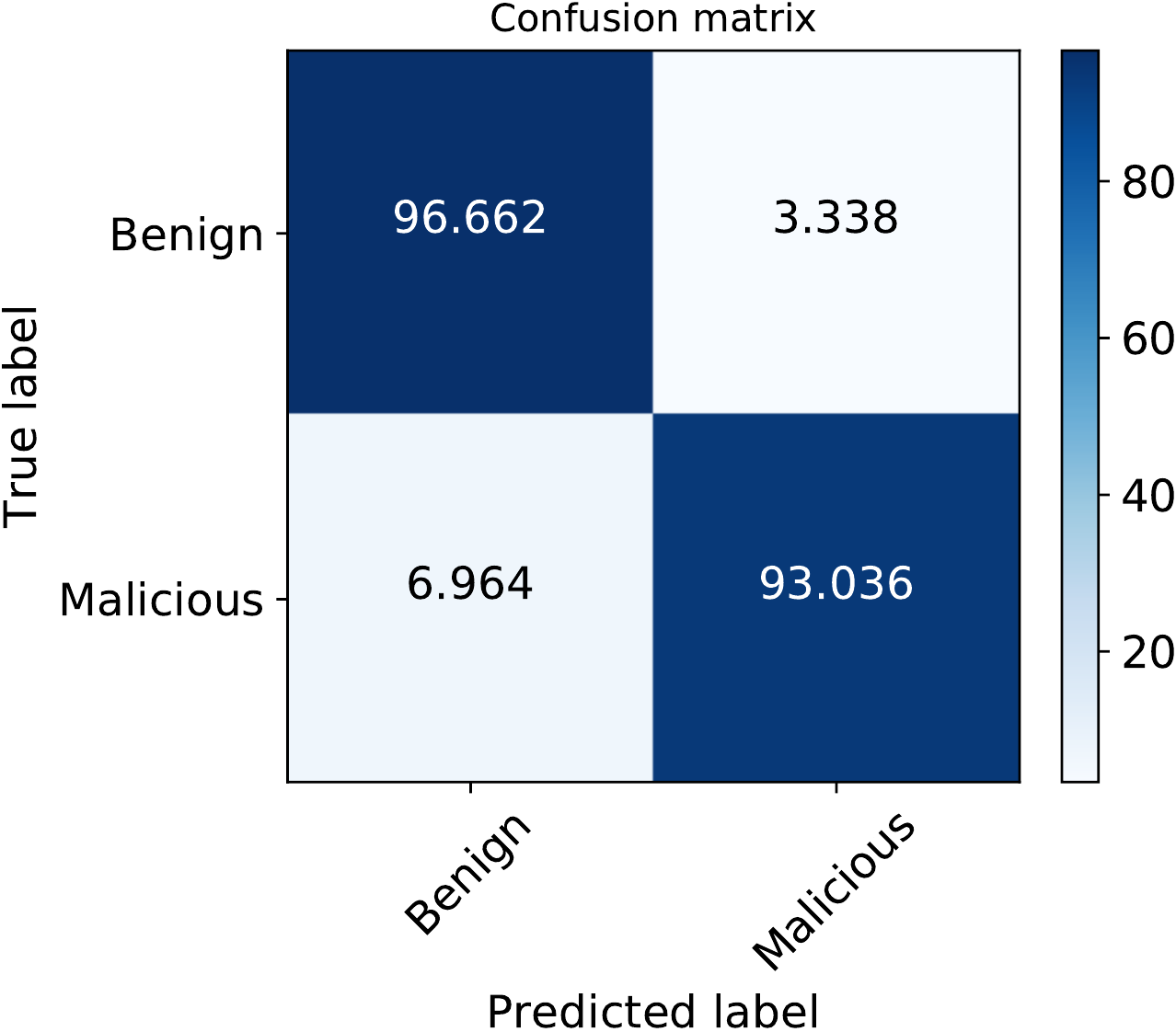}
        \caption{LLC-VAE test accuracy.}
        \label{fig:llc-test}
    \end{subfigure}
    \hfill
    \begin{subfigure}[b]{0.4\textwidth}
        \includegraphics[scale=0.4]{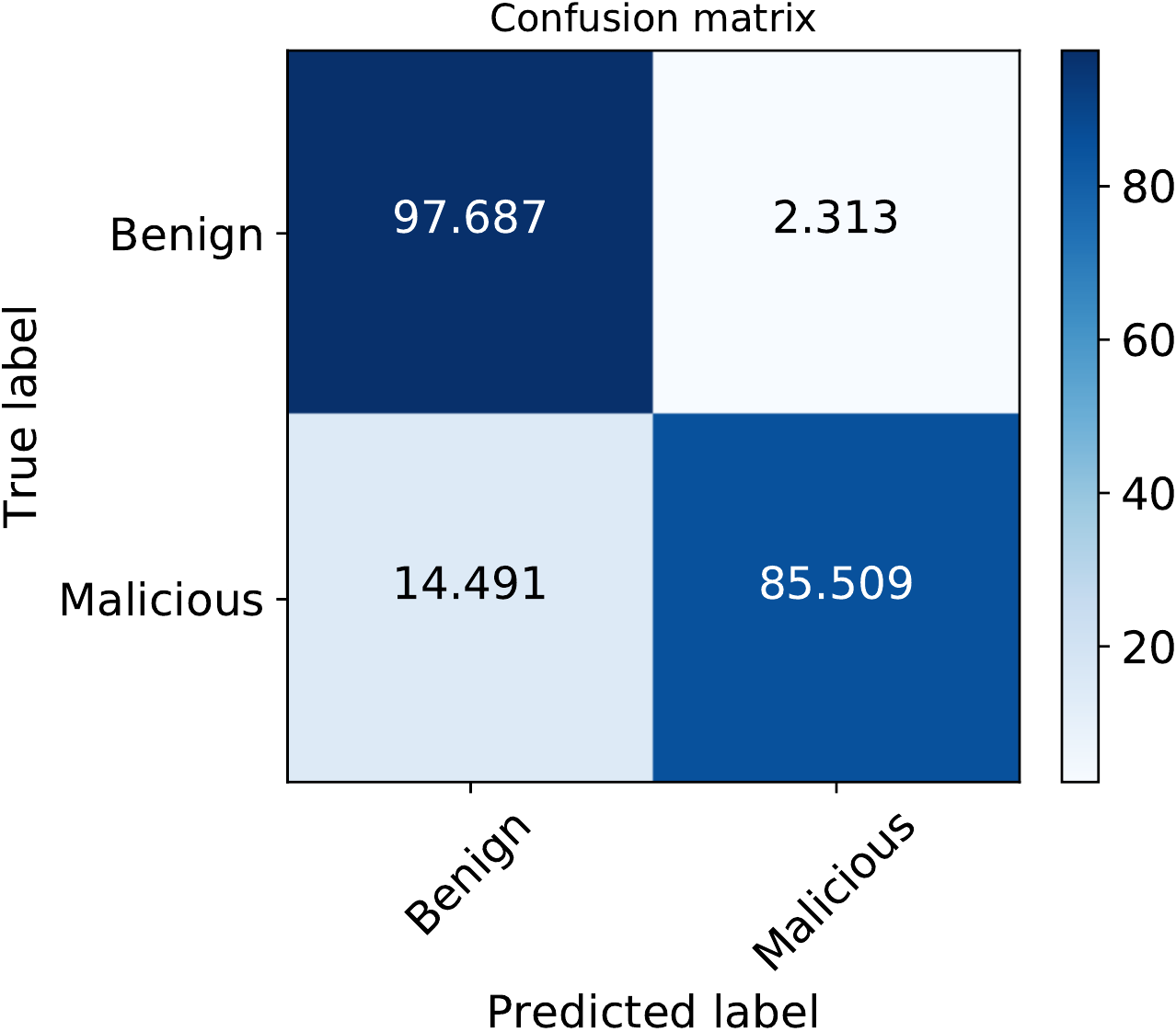}
        \caption{CNN test accuracy.}
        \label{fig:cnn-test}
    \end{subfigure}
    \caption[LLC-VAE and Simple CNN Test Accuracy Comparison]{Test accuracy comparison.}
    \label{fig:analysis-llc-cnn-test}
\end{figure*}

Compared with the simple convolutional neural network, the LLC-VAE performed better overall on the test set. The best test results for the LLC-VAE was on preset 4b \ref{tab:p4b}, as can be seen in figure \ref{fig:llc-test}. The best test results for the simple convolutional network was on preset 6a \ref{tab:p6a}, as can be seen in figure \ref{fig:cnn-test}. The LLC-VAE had less than 1\% worse benign accuracy, but substantially increased malicious accuracy, at about 7.5\% higher. For both models the validation accuracy was significantly higher, but ultimately, the test accuracy gives a better understanding of how the models would perform in general, since the test set comes from a different computer network. Hence, there are fewer similarities that could cause overfitting, as the datasets are internally correlated to a large degree.

Using the LLC-VAE as a part of a mitigation system is within reason, even though 3 in every 100 benign flows would be blocked. Although, relying solely on the detection capabilities of the LLC-VAE would not be a good solution. Instead, it should be incorporated as a part of a larger mitigation system.

\subsubsection{Flow Processing Time}

After fully training the LLC-VAE on the best preset, preset 4b \ref{tab:p4b}, we measured the speed of the model. The LLC-VAE was estimated to be able to process one batch in about 10.8 ms $\pm$ 0.3 ms, where one batch contains 1,024 flows. This is between 90 - 95 batches per second, which translates to about 92,000 - 97,000 traffic flows per second. These measurements are only from the LLC-VAE, and does not include time taken for data transformations. Furthermore, the model has not been optimized for speed either. If this approach were to be added as a part of a full mitigation system, the overall flow processing time would be expected to increase, so these measurements only give an indication of what to expect from the LLC-VAE.

\section{Conclusion}
\label{conclusion}

This article has presented two different approaches for DoS and DDoS mitigation using deep learning. Both approaches build on the framework of a Variational Autoencoder, using pre-generated datasets to classify different types of computer network traffic. These datasets provide the two approaches with input features from network traffic flows, allowing them to learn to filter normal and malicious traffic.

The first approach, LLC-VAE, is a latent layer classification network that utilizes the latent layer encodings of a Variational Autoencoder. The LLC-VAE showed clear signs of overfitting towards the training dataset in the beginning, but the generalization capabilities has been greatly improved through various tunings. The improved generalization has in large part been achieved by adjusting the KL-Loss weight, manual feature selection, and tuning of the convolutional layers.

The performance of the LLC-VAE has been tested on two different datasets, split into a training set, a validation set, and a test set. The training and validation sets have data that are internally correlated, hence we use the test set to record model results. When we compared the LLC-VAE performance with a simple convolutional network of a similar structure, the LLC-VAE was overall better at generalizing, achieving better results on the test set.  At the core of the VAE is the KL-Loss value, which regulates the latent layers. Lowering the KL-Loss weight improved the overall model performance. When the weight was lowered past a certain point however, the overall performance of the LLC-VAE declined. Acknowledging that using a Variational Autoencoder over a standard autoencoder had a positive impact on the ability of the model to classify normal and malicious traffic flows.

The second approach, LBD-VAE, relies on the VAE to separate normal and malicious traffic flows into two different probability distributions. A Loss Based Detector is applied to the reconstruction loss, classifying traffic flows using a linear classification layer. The VAE is trained exclusively on normal traffic, and the LBD is trained on a combination of normal and malicious traffic. Since the VAE is only trained using normal traffic, the LBD-VAE is theoretically capable of classifying DoS and DDoS attack types not seen during training. This is because malicious flows would not fit into the probability distribution of the normal flows. The results from the training runs using the LBD-VAE showed that it is currently unable to be used as a part of a mitigation system. The VAE had difficulty with separating the two probability distributions, hence the classifier achieved unsatisfactory results. Still, if the LBD-VAE were to be tuned further, it has the potential of becoming a viable method for mitigation.

This article has proven that deep learning based techniques can be effective at countering DoS and DDoS attacks. The second approach, LBD-VAE, currently does not perform well enough to be used as a mitigation system, but is theoretically promising. More research should be performed to explore other possibilities. The overall best test results was achieved by the LLC-VAE, being able to classify benign and malicious traffic at upwards of 97\% and 93\% accuracy respectively, on simulated data in the generalized case. The LLC-VAE has proven itself capable of competing with traditional mitigation methods, but will need further tuning to ensure even better performance.

\bibliographystyle{frontiersinSCNS_ENG_HUMS}
\bibliography{references.bib}

\begin{thebibliography}{47}
\providecommand{\natexlab}[1]{#1}
\expandafter\ifx\csname urlstyle\endcsname\relax
  \providecommand{\doi}[1]{doi:\discretionary{}{}{}#1}\else
  \providecommand{\doi}{doi:\discretionary{}{}{}\begingroup
  \urlstyle{rm}\Url}\fi
\providecommand{\selectlanguage}[1]{\relax}
\providecommand{\bibAnnoteFile}[1]{%
  \IfFileExists{#1}{\begin{quotation}\noindent\textsc{Key:} #1\\
  \textsc{Annotation:}\ \input{#1}\end{quotation}}{}}
\providecommand{\bibAnnote}[2]{%
  \begin{quotation}\noindent\textsc{Key:} #1\\
  \textsc{Annotation:}\ #2\end{quotation}}

\bibitem[{B{\aa}rli(2019)}]{baarli2019ddos}
B{\aa}rli, E.~M. (2019).
\newblock \emph{DDoS and DoS Mitigation Using a Variational Autoencoder}.
\newblock Master's thesis
\bibAnnoteFile{baarli2019ddos}

\bibitem[{Behal and Kumar(2017)}]{Behal-and-Kumar-2017}
Behal, S. and Kumar, K. (2017).
\newblock Detection of ddos attacks and flash events using information theory
  metrics--an empirical investigation.
\newblock \emph{Computer Communications} 103, 18--28
\bibAnnoteFile{Behal-and-Kumar-2017}

\bibitem[{Blei et~al.(2017)Blei, Kucukelbir, and McAuliffe}]{blei-et-al-2017}
Blei, D.~M., Kucukelbir, A., and McAuliffe, J.~D. (2017).
\newblock Variational inference: A review for statisticians.
\newblock \emph{Journal of the American Statistical Association} 112, 859--877
\bibAnnoteFile{blei-et-al-2017}

\bibitem[{Bourlard and Kamp(1988)}]{Bourlard-and-Kamp-1988}
Bourlard, H. and Kamp, Y. (1988).
\newblock Auto-association by multilayer perceptrons and singular value
  decomposition.
\newblock \emph{Biological Cybernetics} 59, 291--294.
\newblock \doi{10.1007/BF00332918}
\bibAnnoteFile{Bourlard-and-Kamp-1988}

\bibitem[{Cisco~Systems(2019)}]{Cisco-forecast-2019}
Cisco~Systems, I. (2019).
\newblock Cisco visual networking index: Forecast and trends, 2017–2022
  \url{https://www.cisco.com/c/en/us/solutions/collateral/service-provider/visual-networking-index-vni/white-paper-c11-741490.html}
\bibAnnoteFile{Cisco-forecast-2019}

\bibitem[{{Doersch}(2016)}]{Doersch-2016}
{Doersch}, C. (2016).
\newblock {Tutorial on Variational Autoencoders}.
\newblock \emph{arXiv e-prints} , arXiv:1606.05908
\bibAnnoteFile{Doersch-2016}

\bibitem[{Establishment and the Canadian Institute~for
  Cybersecurity(2018)}]{CSECICIDS2018}
[Dataset] Establishment, T. C.~S. and the Canadian Institute~for Cybersecurity
  (2018).
\newblock Csecicids2018 dataset download and information.
\newblock Dataset information:
  \url{https://www.unb.ca/cic/datasets/ids-2018.html}, and download:
  \url{https://registry.opendata.aws/cse-cic-ids2018/}
\bibAnnoteFile{CSECICIDS2018}

\bibitem[{for Cybersecurity(2017)}]{CICIDS2017}
[Dataset] for Cybersecurity, C.~I. (2017).
\newblock Cicids2017 dataset download and information.
\newblock \url{https://www.unb.ca/cic/datasets/ids-2017.html}
\bibAnnoteFile{CICIDS2017}

\bibitem[{Goldstein et~al.(2008)Goldstein, Lampert, Reif, Stahl, and
  Breuel}]{Goldstein-et-al-2008}
Goldstein, M., Lampert, C., Reif, M., Stahl, A., and Breuel, T.~M. (2008).
\newblock Bayes optimal ddos mitigation by adaptive history-based ip filtering.
\newblock In \emph{Seventh International Conference on Networking (ICN 2008)}
  (IEEE Computer Society Press), 174--179.
\newblock \doi{10.1109/ICN.2008.64}
\bibAnnoteFile{Goldstein-et-al-2008}

\bibitem[{Goodfellow et~al.(2016)Goodfellow, Bengio, and
  Courville}]{Goodfellow-et-al-2016}
Goodfellow, I., Bengio, Y., and Courville, A. (2016).
\newblock \emph{Deep Learning} (MIT Press).
\newblock \url{http://www.deeplearningbook.org}
\bibAnnoteFile{Goodfellow-et-al-2016}

\bibitem[{He et~al.(2016)He, Zhang, Ren, and Sun}]{He-et-al-2016}
He, K., Zhang, X., Ren, S., and Sun, J. (2016).
\newblock Deep residual learning for image recognition.
\newblock In \emph{The IEEE Conference on Computer Vision and Pattern
  Recognition (CVPR)}
\bibAnnoteFile{He-et-al-2016}

\bibitem[{{Ioffe} and {Szegedy}(2015)}]{Ioffe-et-al-2015}
{Ioffe}, S. and {Szegedy}, C. (2015).
\newblock {Batch Normalization: Accelerating Deep Network Training by Reducing
  Internal Covariate Shift}.
\newblock \emph{arXiv e-prints} , arXiv:1502.03167
\bibAnnoteFile{Ioffe-et-al-2015}

\bibitem[{{Irie} and {Miyake}(1988)}]{Irie-and-Miyake-1988}
{Irie} and {Miyake} (1988).
\newblock Capabilities of three-layered perceptrons.
\newblock In \emph{IEEE 1988 International Conference on Neural Networks}.
  641--648 vol.1.
\newblock \doi{10.1109/ICNN.1988.23901}
\bibAnnoteFile{Irie-and-Miyake-1988}

\bibitem[{Keskar et~al.(2016)Keskar, Mudigere, Nocedal, Smelyanskiy, and
  Tang}]{Keskar-et-al-2016}
Keskar, N.~S., Mudigere, D., Nocedal, J., Smelyanskiy, M., and Tang, P. T.~P.
  (2016).
\newblock On large-batch training for deep learning: Generalization gap and
  sharp minima.
\newblock \emph{CoRR} abs/1609.04836
\bibAnnoteFile{Keskar-et-al-2016}

\bibitem[{{Kingma} and {Ba}(2014)}]{Kingma-and-Ba-2014}
{Kingma}, D.~P. and {Ba}, J. (2014).
\newblock {Adam: A Method for Stochastic Optimization}.
\newblock \emph{arXiv e-prints} , arXiv:1412.6980
\bibAnnoteFile{Kingma-and-Ba-2014}

\bibitem[{{Kingma} and {Welling}(2013)}]{Kingma-and-Welling-2013}
{Kingma}, D.~P. and {Welling}, M. (2013).
\newblock {Auto-Encoding Variational Bayes}.
\newblock \emph{arXiv e-prints} , arXiv:1312.6114
\bibAnnoteFile{Kingma-and-Welling-2013}

\bibitem[{Kongshavn et~al.(2020)Kongshavn, Haugerud, Yazidi, Maseng, and
  Hammer}]{kongshavn2020mitigating}
Kongshavn, M., Haugerud, H., Yazidi, A., Maseng, T., and Hammer, H. (2020).
\newblock Mitigating ddos using weight-based geographical clustering.
\newblock \emph{Concurrency and Computation: Practice and Experience} 32, e5679
\bibAnnoteFile{kongshavn2020mitigating}

\bibitem[{Lab(2015{\natexlab{a}})}]{Kaspersky-risk-report-2015}
Lab, K. (2015{\natexlab{a}}).
\newblock Denial of service: How businesses evaluate the threat of ddos attacks
  \url{https://media.kasperskycontenthub.com/wp-content/uploads/sites/45/2018/03/08234158/IT_Risks_Survey_Report_Threat_of_DDoS_Attacks.pdf}
\bibAnnoteFile{Kaspersky-risk-report-2015}

\bibitem[{Lab(2015{\natexlab{b}})}]{Kaspersky-risk-report-2015-original}
Lab, K. (2015{\natexlab{b}}).
\newblock Global it security risks survey
  \url{https://media.kaspersky.com/en/business-security/it-security-risks-survey-2015.pdf}
\bibAnnoteFile{Kaspersky-risk-report-2015-original}

\bibitem[{Lashkari et~al.(2017)Lashkari, Draper-Gil, Mamun, and
  Ghorbani}]{Lashkari-et-al-2017}
Lashkari, A.~H., Draper-Gil, G., Mamun, M. S.~I., and Ghorbani, A.~A. (2017).
\newblock Characterization of tor traffic using time based features.
\newblock In \emph{ICISSP}. 253--262
\bibAnnoteFile{Lashkari-et-al-2017}

\bibitem[{LeCun et~al.(1998)LeCun, Cortes, and J.C~Burges}]{MNIST}
[Dataset] LeCun, Y., Cortes, C., and J.C~Burges, C. (1998).
\newblock Mnist dataset download and information.
\newblock \url{http://yann.lecun.com/exdb/mnist/}
\bibAnnoteFile{MNIST}

\bibitem[{{Ling} and {Okada}(2007)}]{Ling-and-Okada-r13-2}
{Ling}, H. and {Okada}, K. (2007).
\newblock An efficient earth mover's distance algorithm for robust histogram
  comparison.
\newblock \emph{IEEE Transactions on Pattern Analysis and Machine Intelligence}
  29, 840--853.
\newblock \doi{10.1109/TPAMI.2007.1058}
\bibAnnoteFile{Ling-and-Okada-r13-2}

\bibitem[{Moura et~al.(2020)Moura, Hesselman, Schaapman, Boerman, and
  de~Weerdt}]{moura2020into}
Moura, G.~C., Hesselman, C., Schaapman, G., Boerman, N., and de~Weerdt, O.
  (2020).
\newblock Into the ddos maelstrom: a longitudinal study of a scrubbing service.
\newblock In \emph{2020 IEEE European Symposium on Security and Privacy
  Workshops (EuroS\&PW)} (IEEE), 550--558
\bibAnnoteFile{moura2020into}

\bibitem[{Ng et~al.(2011)}]{ng-et-al-2011}
Ng, A. et~al. (2011).
\newblock Sparse autoencoder.
\newblock \emph{CS294A Lecture notes} 72, 1--19
\bibAnnoteFile{ng-et-al-2011}

\bibitem[{Niyaz et~al.(2016)Niyaz, Sun, and Javaid}]{Niyaz-et-al-2016}
Niyaz, Q., Sun, W., and Javaid, A.~Y. (2016).
\newblock A deep learning based ddos detection system in software-defined
  networking (sdn).
\newblock \emph{arXiv preprint arXiv:1611.07400}
\bibAnnoteFile{Niyaz-et-al-2016}

\bibitem[{Nychis et~al.(2008)Nychis, Sekar, Andersen, Kim, and
  Zhang}]{Nychis-et-al-2008}
Nychis, G., Sekar, V., Andersen, D.~G., Kim, H., and Zhang, H. (2008).
\newblock An empirical evaluation of entropy-based traffic anomaly detection.
\newblock In \emph{Proceedings of the 8th ACM SIGCOMM Conference on Internet
  Measurement} (New York, NY, USA: ACM), IMC '08, 151--156.
\newblock \doi{10.1145/1452520.1452539}
\bibAnnoteFile{Nychis-et-al-2008}

\bibitem[{Pekta{\c{s}} and Acarman(2018)}]{Pektacs-and-Acarman-2018}
Pekta{\c{s}}, A. and Acarman, T. (2018).
\newblock A deep learning method to detect network intrusion through flow-based
  features.
\newblock \emph{International Journal of Network Management} , e2050
\bibAnnoteFile{Pektacs-and-Acarman-2018}

\bibitem[{Perera and Patel(2018)}]{perera-et-al-2018}
Perera, P. and Patel, V.~M. (2018).
\newblock Learning deep features for one-class classification.
\newblock \emph{arXiv preprint arXiv:1801.05365}
\bibAnnoteFile{perera-et-al-2018}

\bibitem[{Pu et~al.(2016)Pu, Gan, Henao, Yuan, Li, Stevens
  et~al.}]{Pu-et-al-2016}
Pu, Y., Gan, Z., Henao, R., Yuan, X., Li, C., Stevens, A., et~al. (2016).
\newblock Variational autoencoder for deep learning of images, labels and
  captions.
\newblock In \emph{Advances in Neural Information Processing Systems 29}, eds.
  D.~D. Lee, M.~Sugiyama, U.~V. Luxburg, I.~Guyon, and R.~Garnett (Curran
  Associates, Inc.). 2352--2360
\bibAnnoteFile{Pu-et-al-2016}

\bibitem[{{Radford} et~al.(2018{\natexlab{a}}){Radford}, {Apolonio}, {Trias},
  and {Simpson}}]{Radford-et-al-2018-r7-1}
{Radford}, B.~J., {Apolonio}, L.~M., {Trias}, A.~J., and {Simpson}, J.~A.
  (2018{\natexlab{a}}).
\newblock {Network Traffic Anomaly Detection Using Recurrent Neural Networks}.
\newblock \emph{arXiv e-prints} , arXiv:1803.10769
\bibAnnoteFile{Radford-et-al-2018-r7-1}

\bibitem[{{Radford} et~al.(2018{\natexlab{b}}){Radford}, {Richardson}, and
  {Davis}}]{Radford-et-al-2018}
{Radford}, B.~J., {Richardson}, B.~D., and {Davis}, S.~E. (2018{\natexlab{b}}).
\newblock {Sequence Aggregation Rules for Anomaly Detection in Computer Network
  Traffic}.
\newblock \emph{arXiv e-prints} , arXiv:1805.03735
\bibAnnoteFile{Radford-et-al-2018}

\bibitem[{Scott and Spaniel(2016)}]{Scott-2016}
Scott, J. and Spaniel, D. (2016).
\newblock Rise of the machines: The dyn attack was just a practice run
\bibAnnoteFile{Scott-2016}

\bibitem[{{Sermanet} et~al.(2013){Sermanet}, {Eigen}, {Zhang}, {Mathieu},
  {Fergus}, and {LeCun}}]{Sermanet-et-al-2013}
{Sermanet}, P., {Eigen}, D., {Zhang}, X., {Mathieu}, M., {Fergus}, R., and
  {LeCun}, Y. (2013).
\newblock {OverFeat: Integrated Recognition, Localization and Detection using
  Convolutional Networks}.
\newblock \emph{arXiv e-prints} , arXiv:1312.6229
\bibAnnoteFile{Sermanet-et-al-2013}

\bibitem[{Shannon(2001)}]{Shannon-2001}
Shannon, C.~E. (2001).
\newblock A mathematical theory of communication.
\newblock \emph{SIGMOBILE Mob. Comput. Commun. Rev.} 5, 3--55.
\newblock \doi{10.1145/584091.584093}
\bibAnnoteFile{Shannon-2001}

\bibitem[{Sharafaldin et~al.(2018)Sharafaldin, Habibi~Lashkari, and
  Ghorbani}]{Sharafaldin-et-al-2018}
Sharafaldin, I., Habibi~Lashkari, A., and Ghorbani, A. (2018).
\newblock Toward generating a new intrusion detection dataset and intrusion
  traffic characterization.
\newblock 108--116.
\newblock \doi{10.5220/0006639801080116}
\bibAnnoteFile{Sharafaldin-et-al-2018}

\bibitem[{Shiravi et~al.(2012)Shiravi, Shiravi, Tavallaee, and
  Ghorbani}]{Shiravi-et-al-2012}
Shiravi, A., Shiravi, H., Tavallaee, M., and Ghorbani, A.~A. (2012).
\newblock Toward developing a systematic approach to generate benchmark
  datasets for intrusion detection.
\newblock \emph{Computers \& Security} 31, 357 -- 374.
\newblock \doi{https://doi.org/10.1016/j.cose.2011.12.012}.
\newblock Dataset download link: \url{https://www.unb.ca/cic/datasets/ids.html}
\bibAnnoteFile{Shiravi-et-al-2012}

\bibitem[{Smith et~al.(2017)Smith, Kindermans, and Le}]{Smith-et-al-2017}
Smith, S.~L., Kindermans, P., and Le, Q.~V. (2017).
\newblock Don't decay the learning rate, increase the batch size.
\newblock \emph{CoRR} abs/1711.00489
\bibAnnoteFile{Smith-et-al-2017}

\bibitem[{Smith(1997)}]{Smith-1997}
Smith, S.~W. (1997).
\newblock \emph{The Scientist and Engineer's Guide to Digital Signal
  Processing} (San Diego, CA, USA: California Technical Publishing).
\newblock \url{http://www.dspguide.com/ch13/2.html}
\bibAnnoteFile{Smith-1997}

\bibitem[{Srivastava et~al.(2014)Srivastava, Hinton, Krizhevsky, Sutskever, and
  Salakhutdinov}]{Srivastava-et-al-2014}
Srivastava, N., Hinton, G., Krizhevsky, A., Sutskever, I., and Salakhutdinov,
  R. (2014).
\newblock Dropout: a simple way to prevent neural networks from overfitting.
\newblock \emph{The Journal of Machine Learning Research} 15, 1929--1958
\bibAnnoteFile{Srivastava-et-al-2014}

\bibitem[{{Tan} et~al.(2014){Tan}, {Jamdagni}, {He}, {Nanda}, and
  {Liu}}]{Tan-et-al-2014-r13-1}
{Tan}, Z., {Jamdagni}, A., {He}, X., {Nanda}, P., and {Liu}, R.~P. (2014).
\newblock A system for denial-of-service attack detection based on multivariate
  correlation analysis.
\newblock \emph{IEEE Transactions on Parallel and Distributed Systems} 25,
  447--456.
\newblock \doi{10.1109/TPDS.2013.146}
\bibAnnoteFile{Tan-et-al-2014-r13-1}

\bibitem[{{Tan} et~al.(2015){Tan}, {Jamdagni}, {He}, {Nanda}, {Liu}, and
  {Hu}}]{Tan-et-al-2015}
{Tan}, Z., {Jamdagni}, A., {He}, X., {Nanda}, P., {Liu}, R.~P., and {Hu}, J.
  (2015).
\newblock Detection of denial-of-service attacks based on computer vision
  techniques.
\newblock \emph{IEEE Transactions on Computers} 64, 2519--2533.
\newblock \doi{10.1109/TC.2014.2375218}
\bibAnnoteFile{Tan-et-al-2015}

\bibitem[{{Tulio Ribeiro} et~al.(2016){Tulio Ribeiro}, {Singh}, and
  {Guestrin}}]{Ribeiro-et-al-2016}
{Tulio Ribeiro}, M., {Singh}, S., and {Guestrin}, C. (2016).
\newblock {``Why Should I Trust You?'': Explaining the Predictions of Any
  Classifier}.
\newblock \emph{arXiv e-prints} , arXiv:1602.04938Github Project Page:
  \url{https://github.com/marcotcr/lime}
\bibAnnoteFile{Ribeiro-et-al-2016}

\bibitem[{University~of California(1999)}]{KDD99}
[Dataset] University~of California, I. (1999).
\newblock Kdd99 dataset download and information.
\newblock \url{http://kdd.ics.uci.edu/databases/kddcup99/kddcup99.html}
\bibAnnoteFile{KDD99}

\bibitem[{Vincent et~al.(2010)Vincent, Larochelle, Lajoie, Bengio, and
  Manzagol}]{vincent-et-al-2010}
Vincent, P., Larochelle, H., Lajoie, I., Bengio, Y., and Manzagol, P.-A.
  (2010).
\newblock Stacked denoising autoencoders: Learning useful representations in a
  deep network with a local denoising criterion.
\newblock \emph{Journal of machine learning research} 11, 3371--3408
\bibAnnoteFile{vincent-et-al-2010}

\bibitem[{{Xie} and {Yu}(2009)}]{Xie-and-Yu-2009}
{Xie}, Y. and {Yu}, S. (2009).
\newblock A large-scale hidden semi-markov model for anomaly detection on user
  browsing behaviors.
\newblock \emph{IEEE/ACM Transactions on Networking} 17, 54--65.
\newblock \doi{10.1109/TNET.2008.923716}
\bibAnnoteFile{Xie-and-Yu-2009}

\bibitem[{You et~al.(2020)You, Jiao, Li, and Zhou}]{you2020scheduling}
You, W., Jiao, L., Li, J., and Zhou, R. (2020).
\newblock Scheduling ddos cloud scrubbing in isp networks via randomized online
  auctions.
\newblock In \emph{IEEE International Conference on Computer Communications
  (INFOCOM)}
\bibAnnoteFile{you2020scheduling}

\bibitem[{{Zargar} et~al.(2013){Zargar}, {Joshi}, and
  {Tipper}}]{Zargar-et-al-2013}
{Zargar}, S.~T., {Joshi}, J., and {Tipper}, D. (2013).
\newblock A survey of defense mechanisms against distributed denial of service
  (ddos) flooding attacks.
\newblock \emph{IEEE Communications Surveys Tutorials} 15, 2046--2069.
\newblock \doi{10.1109/SURV.2013.031413.00127}
\bibAnnoteFile{Zargar-et-al-2013}

\end{thebibliography}

\end{document}